\def\paperID{0004}
\def\confName{CVPR}
\def\confYear{2026}

\def\paperTitle{TerraSky3D: Multi-View Reconstructions of European Landmarks in 4K}

\def\authorBlock{
    Mattia D'Urso$^1$
    \quad
    Yuxi Hu$^1$
    \quad
    Christian Sormann$^2$
    \quad \\
    Mattia Rossi$^2$ 
    \quad 
    Friedrich Fraundorfer$^1$\\
    {\normalsize 
        $^1$Graz University of Technology \{name.surname@tugraz.at\}
        \quad
        $^2$Sony  \{name.surname@sony.com\}
    }
}

\newif\ifreview 
\newif\ifarxiv 
\newif\ifcamera \newcommand{\cameraready}{\cameratrue}
\newif\ifrebuttal 
\cameraready

\pdfoutput=1
\documentclass[10pt,twocolumn,letterpaper]{article}
\ifreview \usepackage[review]{cvpr} \fi
\ifarxiv \usepackage[pagenumbers]{cvpr} \fi
\ifrebuttal \usepackage[rebuttal]{cvpr} \fi
\ifcamera \usepackage{cvpr} \fi
 
\usepackage{graphicx}	
\usepackage{amsmath}	
\usepackage{amssymb}	
\usepackage{booktabs}
\usepackage{times}
\usepackage{microtype}
\usepackage{epsfig}
\usepackage{caption}
\usepackage{float}
\usepackage{placeins}
\usepackage{color, colortbl}
\usepackage{stfloats}
\usepackage{enumitem}
\usepackage{tabularx}
\usepackage{xstring}
\usepackage{multirow}
\usepackage{xspace}
\usepackage{url}
\usepackage{subcaption}
\usepackage{xcolor}
\usepackage{pifont}
\usepackage[hang,flushmargin]{footmisc}

\ifcamera \usepackage[accsupp]{axessibility} \fi

% ...

\ifarxiv  \fi

\newcommand{\R}[1]{{%
    \textbf{%
        \ifstrequal{#1}{1}{\textcolor{red}{R#1}}{%
        \ifstrequal{#1}{2}{\textcolor{blue}{R#1}}{%
        \ifstrequal{#1}{3}{\textcolor{magenta}{R#1}}{%
        \ifstrequal{#1}{4}{\textcolor{teal}{R#1}}{%
                           \textcolor{cyan}{R#1}%
        }}}}%
    }%
}}

\newcommand{\cmark}{\textcolor{green!80!black}{\ding{51}}}
\newcommand{\xmark}{\textcolor{red}{\ding{55}}}
\usepackage{xr-hyper}

\makeatletter
\newcommand*{\addFileDependency}[1]{
  \typeout{(#1)}
  \@addtofilelist{#1}
  \IfFileExists{#1}{}{\typeout{No file #1.}}
}

\makeatother
\newcommand*{\myexternaldocument}[1]{
    \externaldocument{#1}
    \addFileDependency{#1.tex}
    \addFileDependency{#1.aux}
}

\definecolor{cvprblue}{rgb}{0.21,0.49,0.74}
\usepackage[pagebackref,breaklinks,colorlinks,allcolors=cvprblue]{hyperref}
\usepackage[capitalize]{cleveref}
\crefname{section}{Sec.}{Secs.}
\crefname{table}{Table}{Tables}
\crefname{figure}{Fig.}{Figs.}

\ifarxiv \crefname{appendix}{App.}{Apps.}
\else \crefname{appendix}{Suppl.}{Suppls.} \fi

\unless\ifarxiv \myexternaldocument{_supplementary} \fi

\begin{document}
%% TITLE
\title{\paperTitle}
\author{\authorBlock}

\begin{figure*}[t]
    \renewcommand\twocolumn[1][]{#1}%
    \maketitle
    \centering

    % Left image (top aligned)
    \begin{minipage}[t]{0.45\textwidth}
        \vspace{0pt}
        \centering
        \includegraphics[width=\textwidth]{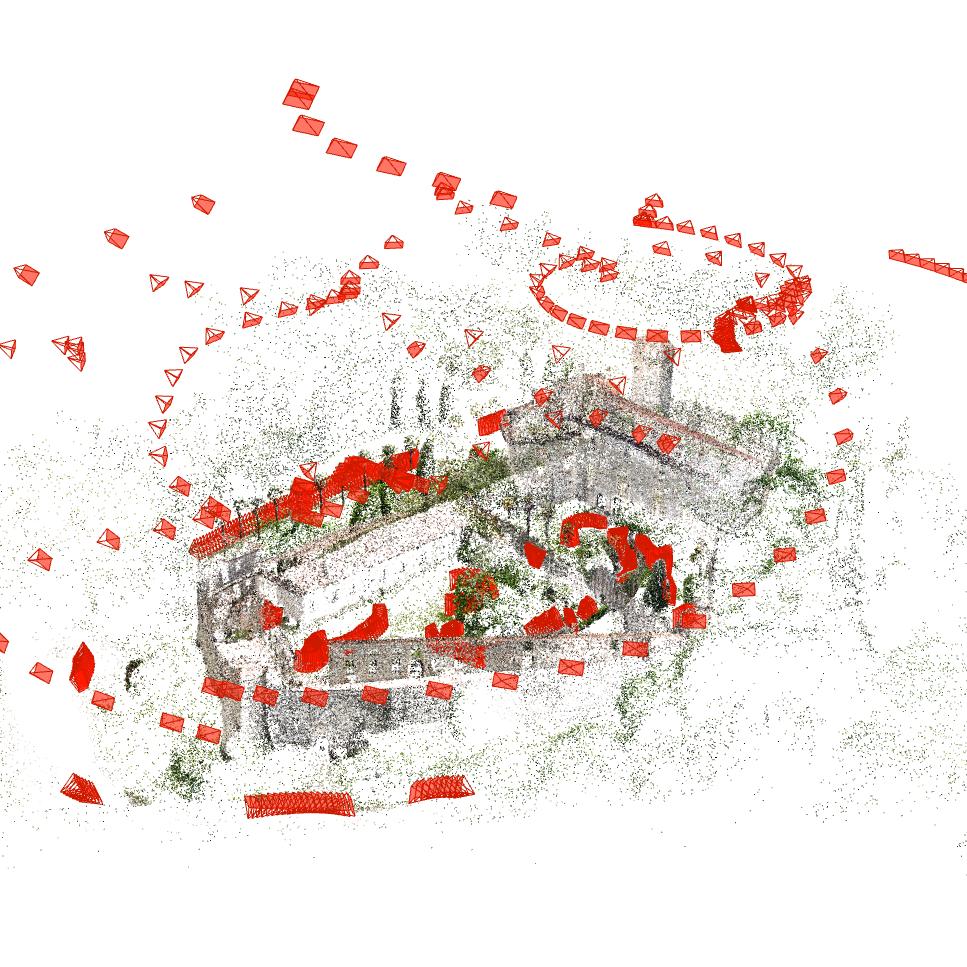}
    \end{minipage}
    \hfill
    % Right 2x4 grid (top aligned) - THIS IS THE MODIFIED PART
    \begin{minipage}[t]{0.53\textwidth}
        \vspace{0pt}
        \centering

    % --- Row 1 ---
    \begin{minipage}[t]{0.24\textwidth}
        \vspace{0pt} % Ensures minipage aligns at the very top
        \includegraphics[width=\textwidth]{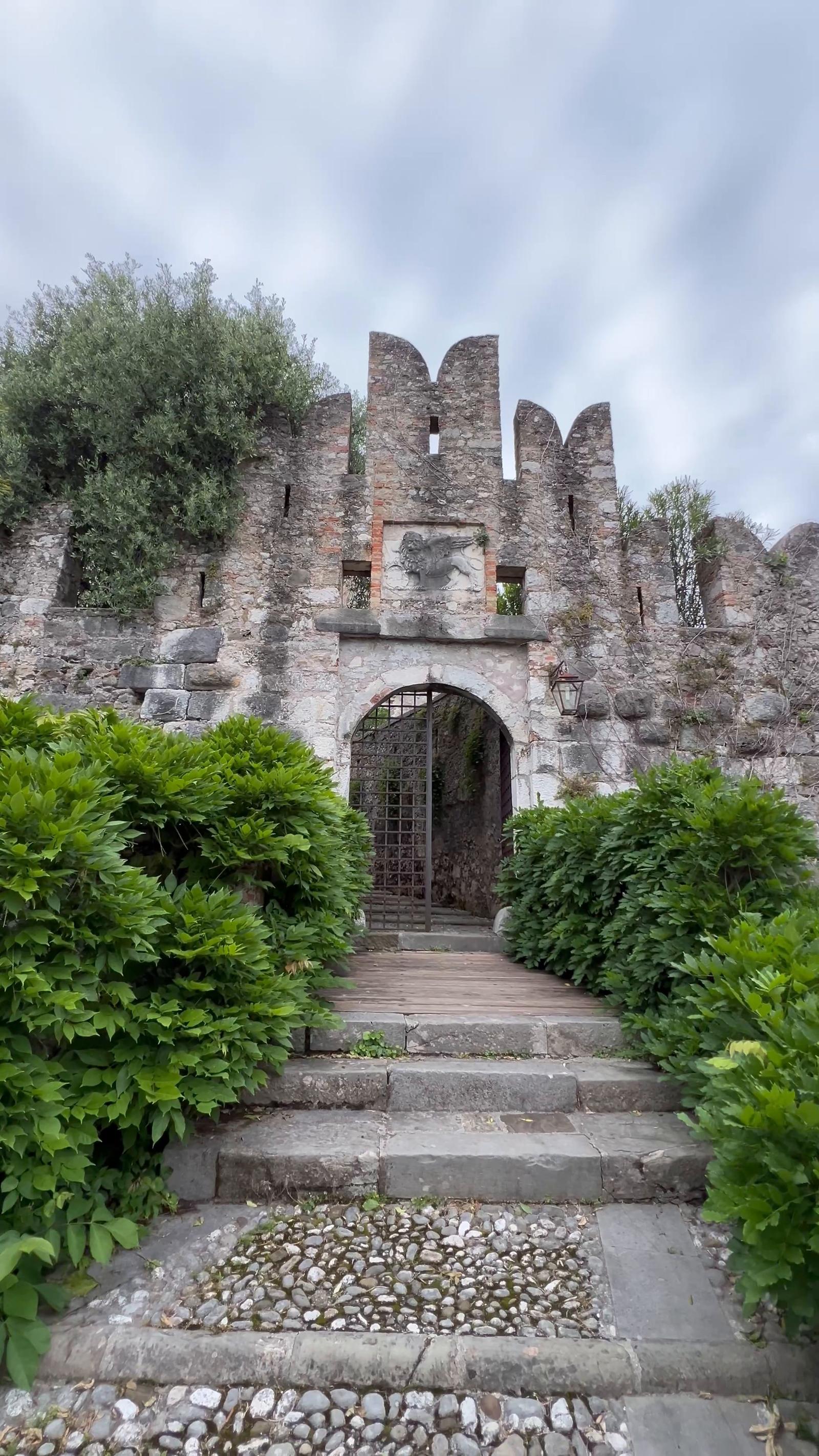}
    \end{minipage}
    \hspace{2pt} % <-- KEY CHANGE: Adds equal horizontal space
    \begin{minipage}[t]{0.24\textwidth}
        \vspace{0pt}
        \includegraphics[width=\textwidth]{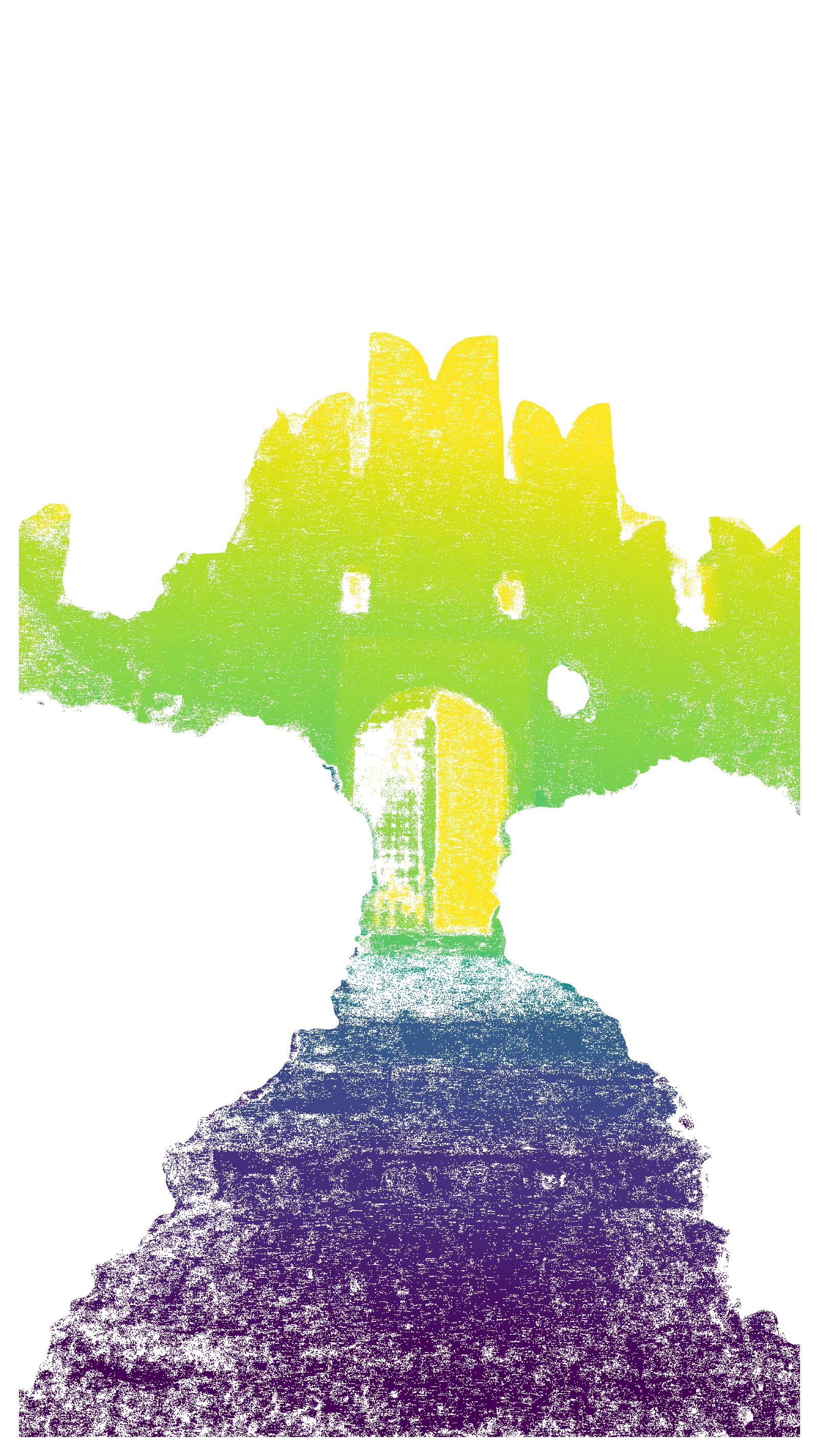}
    \end{minipage}
    \hspace{2pt} % <-- KEY CHANGE: Adds equal horizontal space
    \begin{minipage}[t]{0.375\textwidth}
        \vspace{0pt}
        \includegraphics[width=\textwidth]{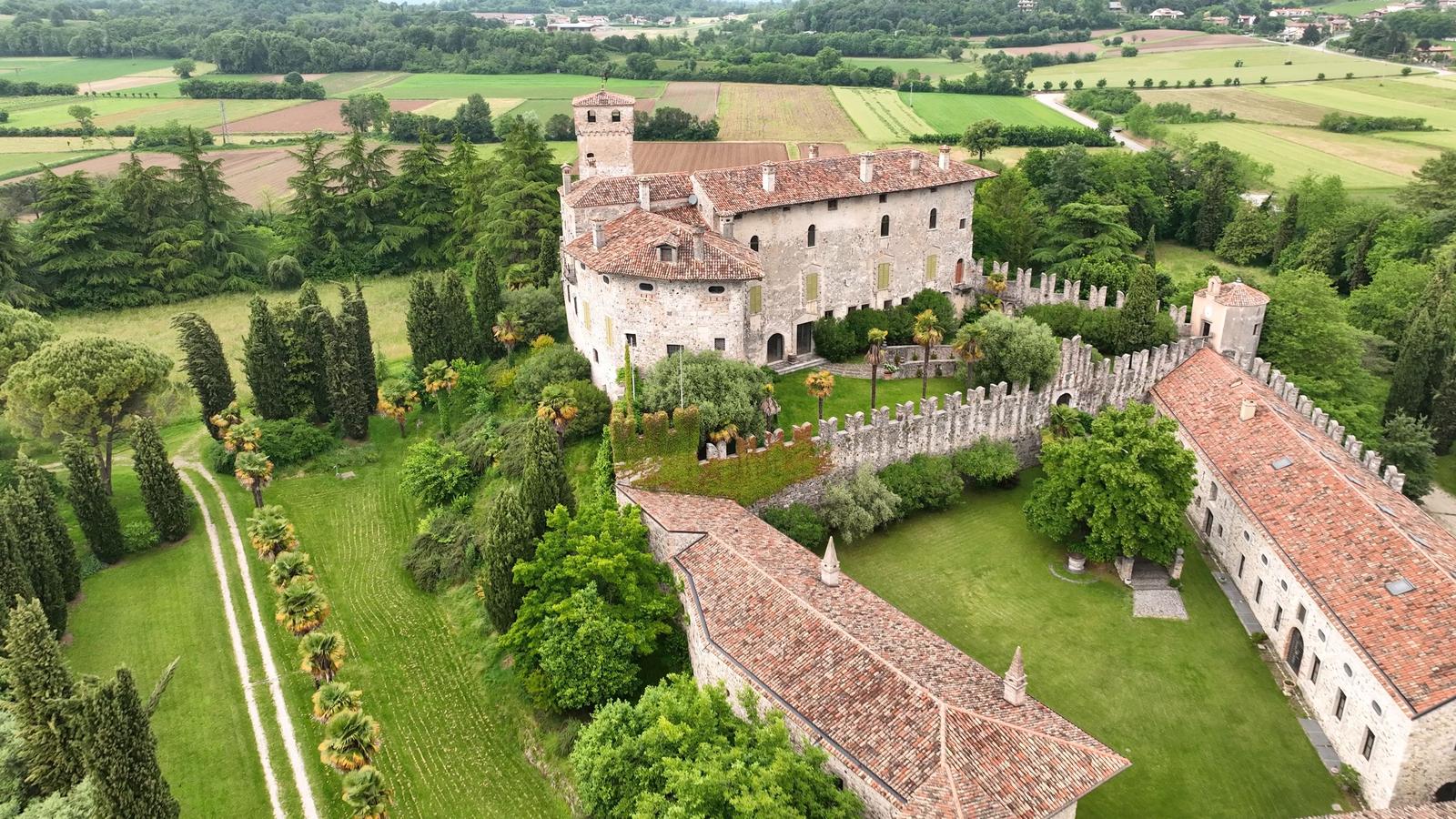}
        \vspace{2pt} % Adds space between the stacked images in this column
        \includegraphics[width=\textwidth]{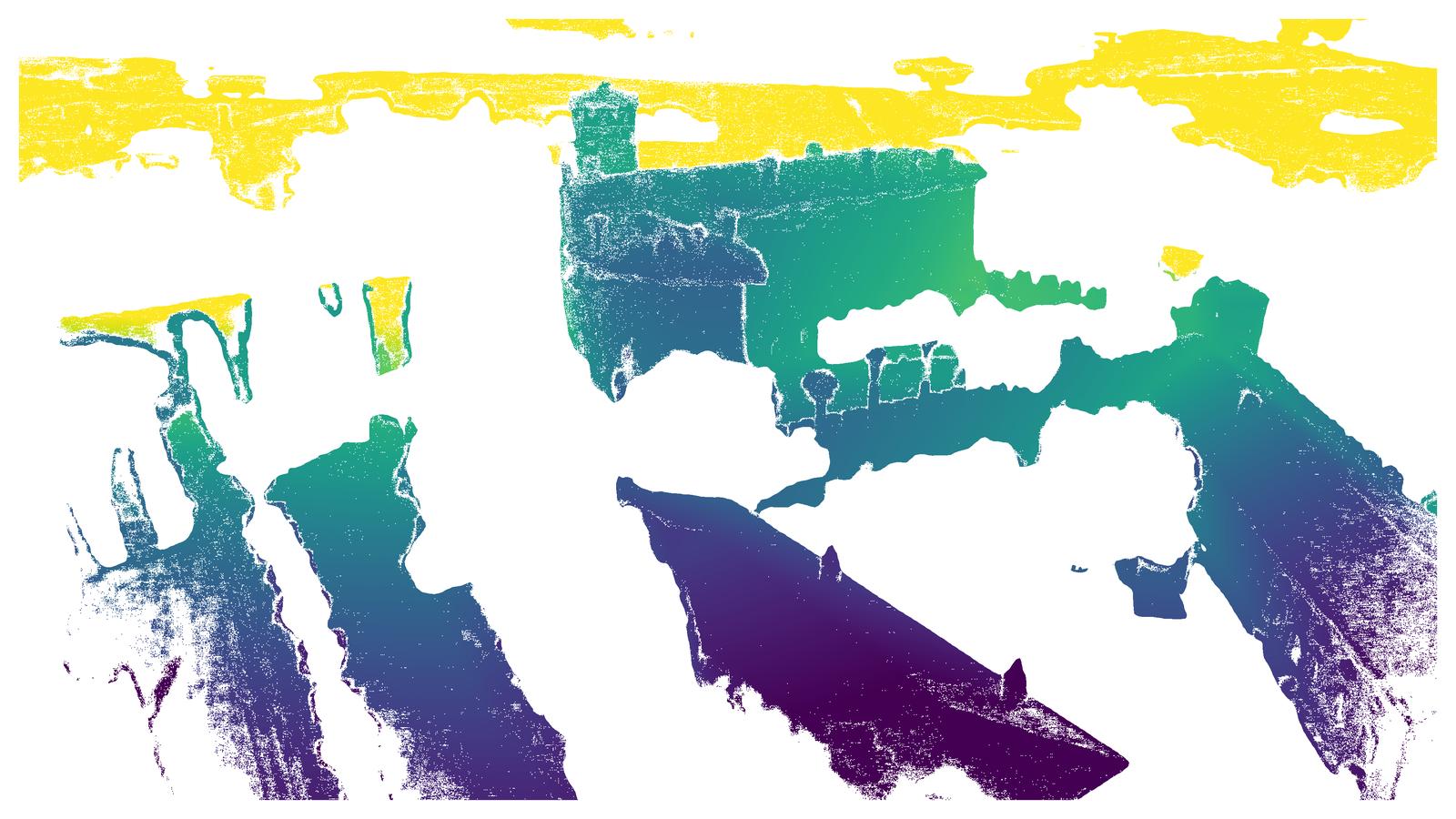}
    \end{minipage}
    
    \vspace{8pt} % <-- KEY CHANGE: This single command creates all the vertical space between Row 1 and Row 2
    
    % --- Row 2 ---
    \begin{minipage}[t]{0.24\textwidth}
        \vspace{0pt}
        \includegraphics[width=\textwidth]{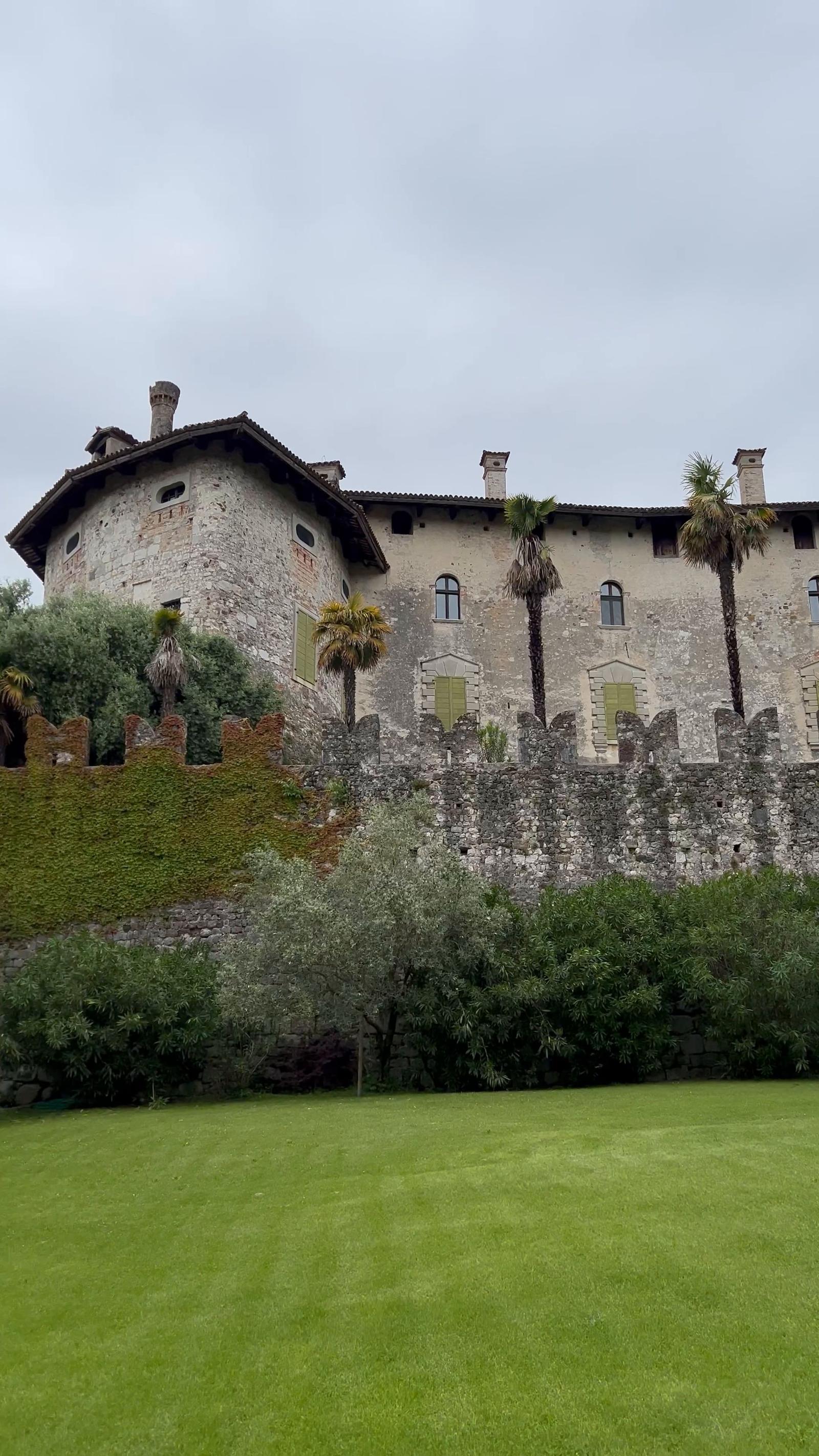}
    \end{minipage}
    \hspace{2pt} % <-- KEY CHANGE: Adds equal horizontal space
    \begin{minipage}[t]{0.24\textwidth}
        \vspace{0pt}
        \includegraphics[width=\textwidth]{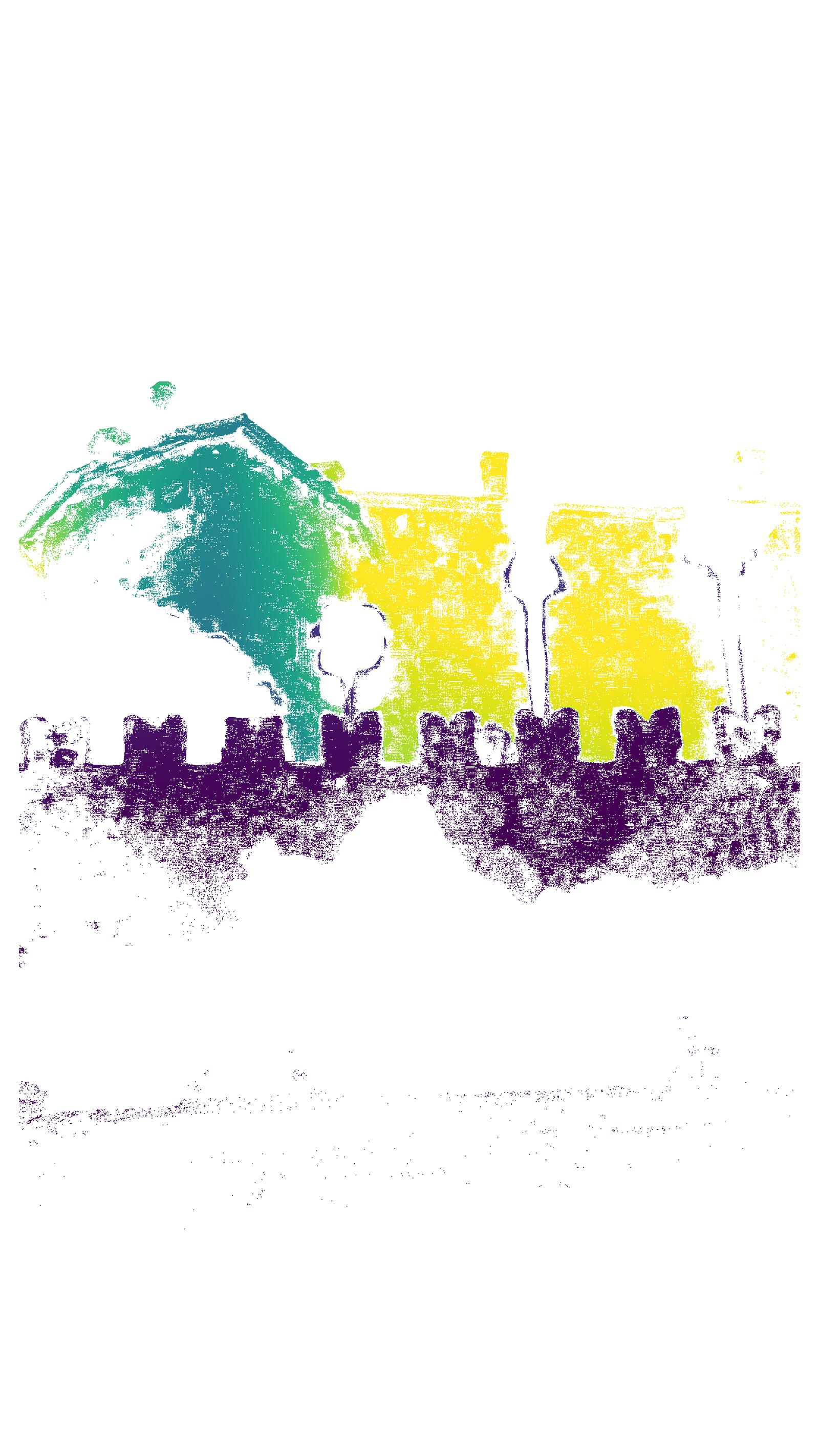}
    \end{minipage}
    \hspace{2pt} % <-- KEY CHANGE: Adds equal horizontal space
    \begin{minipage}[t]{0.375\textwidth}
        \vspace{0pt}
        \includegraphics[width=\textwidth]{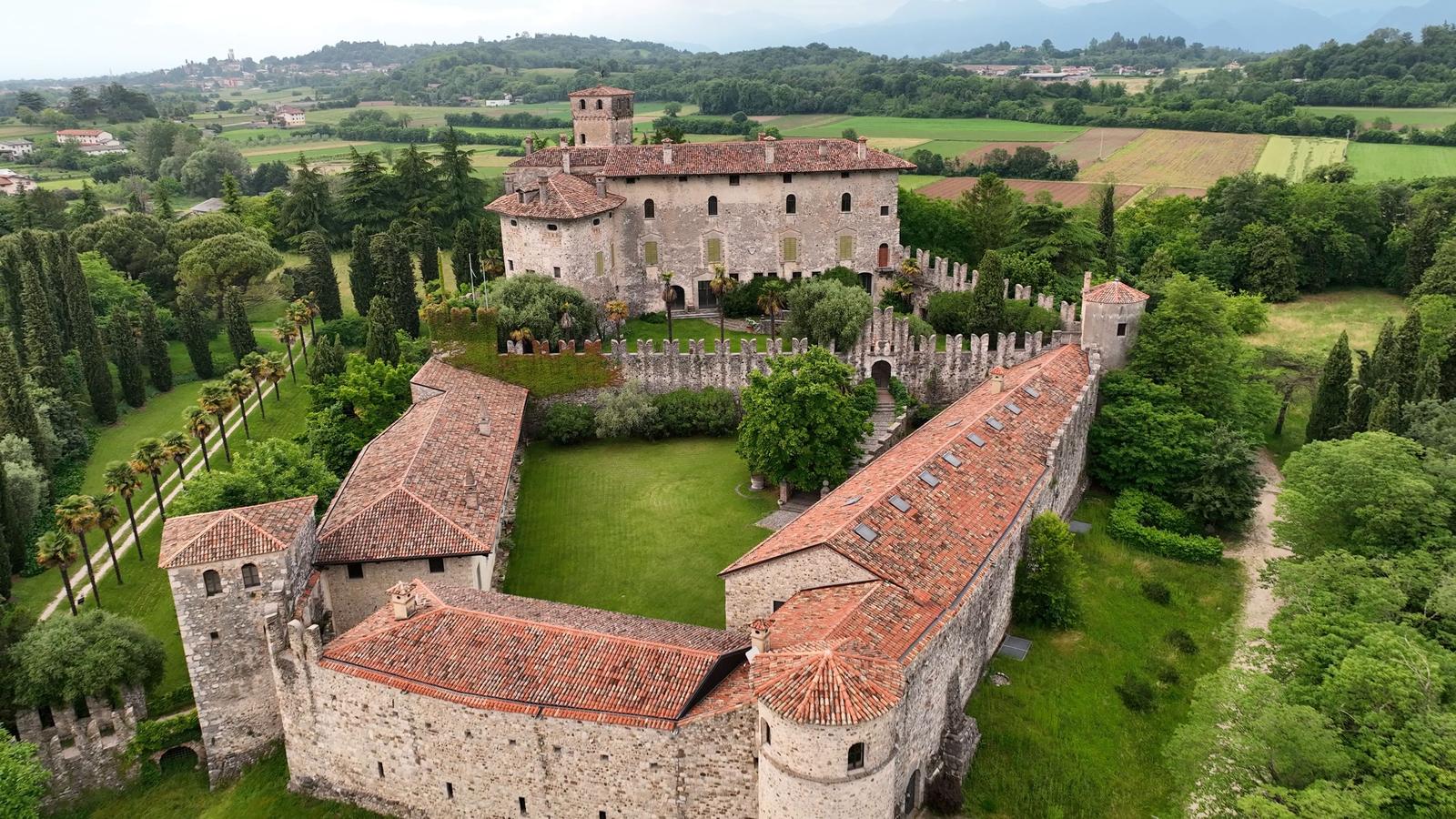}
        \vspace{2pt} % Adds space between the stacked images in this column
        \includegraphics[width=\textwidth]{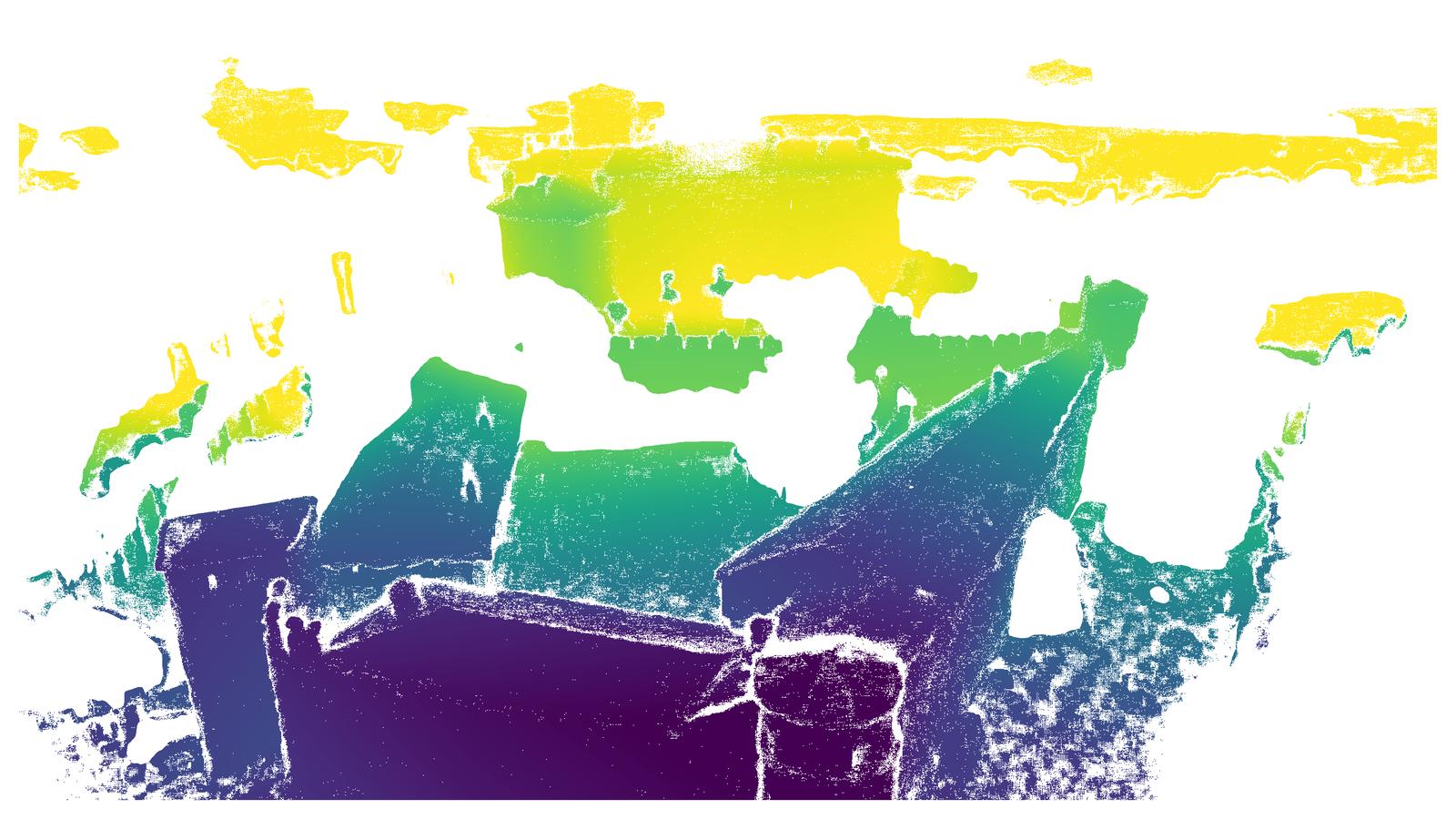}
    \end{minipage}
   
    \end{minipage}
    \caption{\textbf{Example scene from TerraSky3D.} Left: Sparse reconstruction of the Villalta Castle, Italy. Right: Representative images collected from aerial and ground perspectives, shown with their corresponding semantically filtered depth maps.} 
    \label{fig:villalta}
\end{figure*}

\begin{abstract}

Despite the growing need for data of more and more sophisticated 3D reconstruction pipelines, we can still observe a scarcity of suitable public datasets. Existing 3D datasets are either low resolution, limited to a small amount of scenes, based on images of varying quality because retrieved from the internet, or limited to specific scenarios.

Motivated by this lack of suitable 3D datasets, we captured TerraSky3D, a high-resolution large-scale 3D reconstruction dataset divided into 155 ground, aerial, and mixed scenes. The dataset focuses on European landmarks and comes with curated calibration data, camera poses, and depth maps. TerraSky3D tries to answer the need for challenging dataset that can be used to train and evaluate 3D reconstruction-related pipelines. 

\end{abstract}
\section{Introduction}
\label{sec:intro}

The robust understanding and reconstruction of 3D environments from 2D imagery remains a fundamental challenge in computer vision. Advances in downstream applications such as robotic navigation, augmented and virtual reality, and cultural heritage preservation increasingly rely on high-quality, large-scale 3D datasets to train and benchmark modern deep learning models. Specifically, methods in Structure-from-Motion (SfM), Multi-View Stereo (MVS), and neural scene representation are continuously pushing the boundaries of photorealism and geometric accuracy, driving the demand for datasets that accurately reflect real-world geometry.

Existing large-scale datasets often suffer from several limitations. Datasets built from internet photo collections \cite{li2018megadepth, Tung2024MegaScenes} frequently contain inconsistent image quality, unreliable camera intrinsics, and noisy geometry. Furthermore, the majority of datasets focus exclusively on either ground-level (street view) \cite{li2018megadepth, Tung2024MegaScenes, schops2017multi, jin2021image} or aerial \cite{zheng2020university1652, vuong2025aerialmegadepth, zheng2025culture3d} perspectives. This creates a significant modality gap that hinders the development of robust cross-view algorithms. The rapid adoption of unmanned aerial vehicles (UAVs) in surveying and infrastructure inspection has made the aerial perspective increasingly relevant in practical applications, yet robust models are hindered by the lack of aerial and ground training data. This limits the development of cross-view algorithms capable of understanding drastically different viewing angles and scales, a necessary capability for real-world scenarios, such as aerial and ground relocalization. Additionally, recent state-of-the-art methods such as RoMa v2 \cite{edstedt2025roma} demonstrate that incorporating aerial and ground data significantly improves model robustness against large viewpoint changes.

In this paper, we introduce \textbf{TerraSky3D}, a new large-scale dataset designed to address the challenges outlined above. TerraSky3D provides a comprehensive collection of diverse outdoor scene categories, primarily focused on complex European landmarks across Central and Southern Europe, as shown in \cref{fig:map}. By leveraging a high-resolution 4K acquisition pipeline and adopting a rigorously revisited processing methodology for generating dense geometric products, we ensure geometrically consistent data for both ground-level and aerial viewpoints. The dataset includes scenes composed of only ground views, only aerial views, and, mixed aerial and ground scenes, offering extensive versatility for various research domains. Figs. \ref{fig:villalta},  \ref{fig:vienna}, \ref{fig:caporetto}, \ref{fig:erto}, \ref{fig:barcis} show some examples of the proposed scenes.
Our contributions are summarized as follows.
\begin{enumerate}
    \item \textbf{A New Large-Scale Mixed Dataset:} A novel large-scale dataset of high-resolution ground-level and aerial imagery featuring mixed scenes.
    
    \item \textbf{A Simplified Processing Pipeline:} A modern and simplified processing methodology to convert raw video acquisitions into dense, geometrically reliable 3D reconstructions.
    
    \item \textbf{A Comprehensive Evaluation:} A set of experiments validating our data and evaluating current state-of-the-art methods on modern acquisition data.
\end{enumerate}

\noindent Find the dataset at \href{https://github.com/mattiadurso/TerraSky3D}{github.com/mattiadurso/TerraSky3D}. 
\section{Related Work}
\label{sec:related}

\begin{figure}[t]
    \centering
    \includegraphics[width=\linewidth]{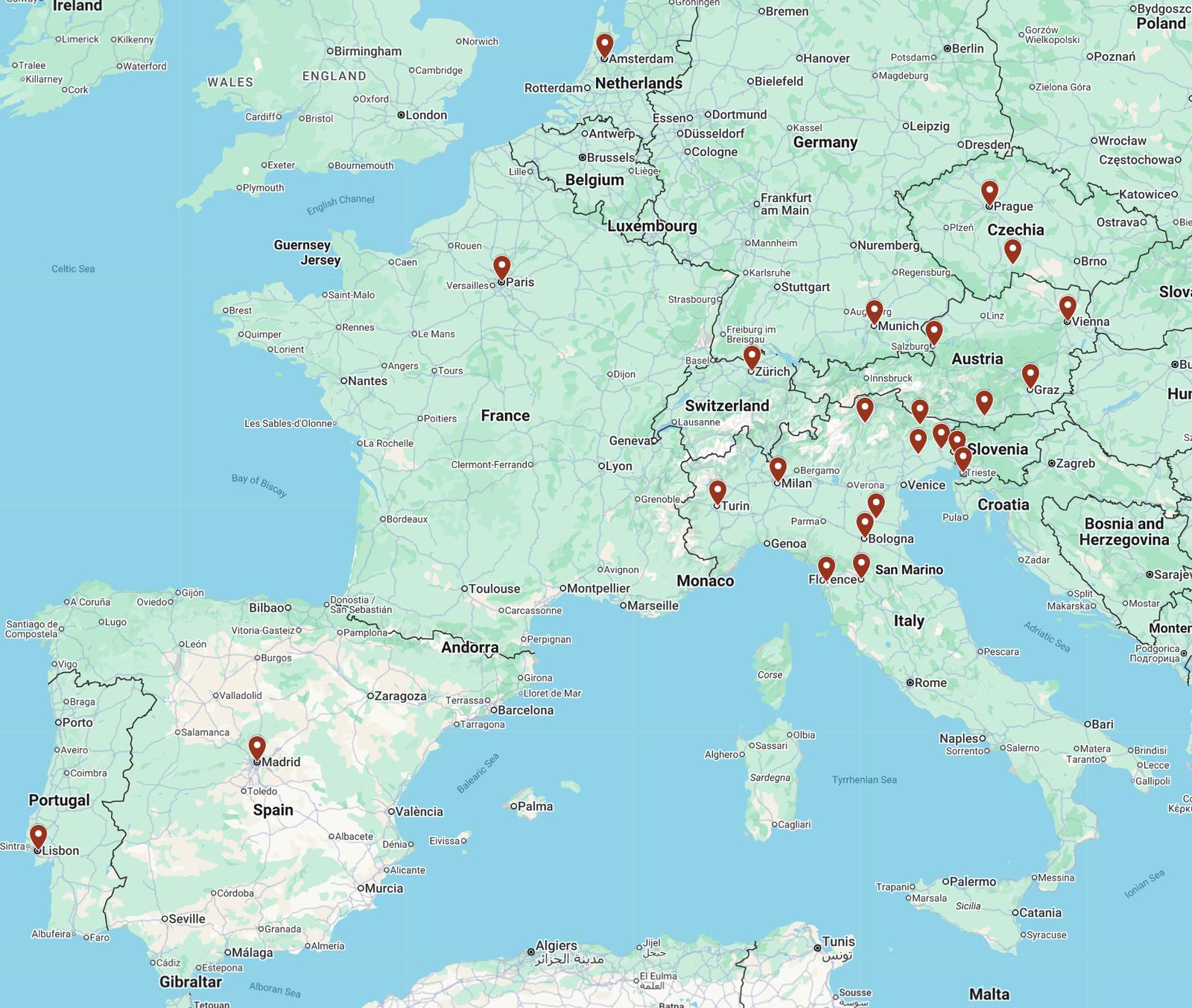}
    \caption{\textbf{Geographical Distribution of Data Collection Sites in \mbox{TerraSky3D}.} The dataset includes sites across 11 European countries. Close locations might share the same red pin.} 
    \label{fig:map}
\end{figure}

\paragraph{Large-Scale Outdoor Photo Collections.}
\label{sec:large_scale_photo_collections}
Large-scale datasets derived from unconstrained internet photos, such as MegaDepth~\cite{li2018megadepth} and MegaScenes~\cite{Tung2024MegaScenes}, have significantly advanced tasks like pose estimation, visual place recognition, depth estimation, and novel view synthesis. These datasets are typically curated by collecting vast quantities of images from public internet sources, which then require extensive processing and cleaning pipelines. Both MegaDepth and MegaScenes utilize COLMAP \cite{schonberger2016structure} to estimate the 3D geometry. MegaDepth extends this process to dense depth estimation using the COLMAP MVS pipeline, followed by post-processing steps such as depth filtering, masking, and refinement. In contrast, MegaScenes primarily provides sparse point clouds. A common challenge in both datasets is detecting and removing artifacts, such as watermarks or timestamps.

A significant limitation of these internet-sourced collections is the lack of known intrinsic camera parameters or shared camera parameters across images. Consequently, intrinsics must be estimated by COLMAP for each image, a process that is prone to inaccuracies, especially when each camera corresponds to only one image. Furthermore, these datasets often contain images with varying aspect ratios within the same scene. %While these datasets are a great demonstration of large-scale reconstruction capabilities, their collection and curation process does not reflect modern and cleaner data collection strategies.

TerraSky3D tackles these problems. We do not collect data from internet sources; instead, we capture each scene specifically for 3D reconstruction. This approach provides sufficient angular views for COLMAP to accurately recover both the 3D structure and the camera poses. By carefully collecting the data, we ensured all images were captured at 4K resolution with a constant aspect ratio. Furthermore, we leverage a recent method for depth estimation, employing APD-MVS~\cite{Wang2023apd} for depth generation. Finally, we employ Mask2Former for semantic filtering within a streamlined yet effective pipeline. %It is also worth noting that, due to our controlled collection process, our images are free of artifacts.

\begin{table*}[t]
    \centering
    \caption{\textbf{Comparison of MVS and 3D Reconstruction Datasets.} An \xmark\phantom{i} under \textbf{Scenes} indicates that the number of scenes is not available. \textbf{Types} are G (Ground) or M (Mixed Air and Ground). \cmark under \textbf{Real} indicates collected imagery, whereas \xmark\phantom{i}denotes computer-generated images; both aerial and ground are indicated when available (e.g., \xmark/\cmark means aerial is computer-generated but ground is real). HR denotes High Resolution (Full HD or above). \textbf{Calib.} indicates camera pre-calibration (\cmark) or intrinsic estimation during the reconstruction process (\xmark). \cmark\phantom{i}\textbf{COLMAP} indicates geometry estimated with COLMAP, while \xmark\phantom{i}indicates no geometry is available. RC stands for Reality Capture.}
    \label{tab:dataset_comparison}
    \begin{tabularx}{\textwidth}{l c c c c c c c X}
        \toprule
        \textbf{Dataset} & 
        \textbf{Year} & 
        \textbf{\#Scenes} & 
        \textbf{Type} & 
        \textbf{Real} & % New Column after Type
        \textbf{HR} & 
        \textbf{Calib.} & 
        \textbf{COLMAP} & 
        \textbf{Depth} \\
        \midrule
        
        % --- Internet Datasets ---
        MegaDepth \cite{li2018megadepth}           & 2018  & 135 & G & \cmark & \xmark & \xmark & \cmark & COLMAP-MVS \\
        MegaScenes \cite{Tung2024MegaScenes}        & 2024  & 100K & G & \cmark & \xmark & \xmark & \cmark & \xmark  \\
        % \addlinespace

        % --- Drone/Aerial Imagery ---
        University-1652 \cite{zheng2020university1652}     & 2020 & 1652 & M & \xmark/\cmark & \xmark & \xmark & \cmark & COLMAP-MVS \\
        CVD-SfM \cite{li2025cvd}                           & 2025 & 3 & M & \cmark/\cmark & \cmark & \xmark & \cmark+GPS & \xmark \\
        Culture3D \cite{zheng2025culture3d}                & 2025 & 41 & M & \cmark/\cmark & \cmark & \cmark & \cmark+RC\phantom{0} & \xmark \\
        AerialMegadepth \cite{vuong2025aerialmegadepth}   & 2025 & 137 & M & \xmark/\cmark & \xmark & \xmark & \cmark & COLMAP-MVS \\
        Megadepth Air-to-Ground \cite{chen2025rdd}        & 2025 & \xmark & M & \cmark/\cmark & \xmark & \xmark & \cmark & \xmark \\
        % \addlinespace
        
        % --- Pose / Benchmark Datasets ---
        Tokyo 24/7 \cite{torii2015tokyo}       & 2015 & 125 & G & \cmark & \xmark & \xmark & \xmark & \xmark \\
        HPatches \cite{balntas2017hpatches}    & 2017 & 116 & G & \cmark & \xmark & \xmark & \xmark & \xmark \\
        %ETH3D \cite{schops2017multi}           & 2017 & 25 & G & \cmark & \cmark & \cmark & \cmark & LiDAR \\
        AAchen \cite{sattler2018benchmarking}  & 2018 & 1 & G & \cmark & \xmark & \cmark & \cmark & \xmark \\
        MegaDepth-1500 \cite{sun2021loftr}     & 2018 & 2 & G & \cmark & \xmark & \xmark & \cmark & COLMAP-MVS \\
        Oxford and Paris \cite{radenovic2018revisiting}  & 2018 & 11 & G & \cmark & \xmark & \xmark & \xmark & \xmark \\
        IMC Phototourism \cite{jin2021image}   & 2021 & 9 & G & \cmark & \xmark & \xmark & \cmark & COLMAP-MVS \\
        Graz4K \cite{durso2026sandesc}         & 2026 & 6 & G & \cmark & \cmark & \cmark & \cmark & \xmark \\
        
        \midrule
        % --- YOUR DATASET ---
        \textbf{TerraSky3D} & 2026 & 155 & M & \cmark/\cmark & \cmark & \cmark & \cmark & APD-MVS \\
        
        \bottomrule
    \end{tabularx}
    
\end{table*}

\paragraph{Mixed Aerial and Ground data.}
Drone imagery has become pivotal for 3D mapping and vision research, as evidenced by recent datasets~\cite{vuong2025aerialmegadepth, zheng2025culture3d, chen2025rdd, zheng2020university1652, li2025cvd}.
However, existing datasets exhibit limitations that hinder the training of robust real-world models for this purpose.
For instance, the aerial imagery in AerialMegaDepth~\cite{vuong2025aerialmegadepth} relies on pseudo-synthetic images rendered from 3D city-wide meshes rather than real photographs, which might introduce a significant domain gap.
Similarly, University-1652~\cite{zheng2020university1652} provides correspondences between satellite and drone views, yet its drone imagery is synthetic and ground imagery are scraped from Google Street View.
In terms of real-world data, Culture3D~\cite{zheng2025culture3d} employs a controlled capture methodology comparable to ours; however, it prioritizes dense coverage of indoor and outdoor scenes for novel view synthesis rather than 3D reconstructions.
Furthermore, while the collection process for the Air-to-Ground dataset is described in~\cite{chen2025rdd}, \emph{only the test set} is publicly available.
Finally, CVD-SfM~\cite{li2025cvd} introduces mixed aerial and ground scenes, yet it is restricted to very few scenes.
Our dataset addresses these shortcomings by providing a diverse set of high-quality scenes. Specifically, we provide numerous real drone images from aerial perspectives, fully registered within SfM reconstructions alongside ground-level views to ensure cross-domain alignment.

\paragraph{Stereo Pose Benchmarks.} 
%ETH3D~\cite{schops2017multi} has long served as a gold standard for pose benchmarks, providing high-quality imagery for precise estimation. However, its test set includes scenes with a limited number of images and often little overlap. 
The IMC Phototourism~\cite{jin2021image} and MegaDepth-1500~\cite{sun2021loftr} benchmarks, both derived from the MegaDepth dataset, are also widely used for pose estimation and feature matching. However, as they originate from the same source, they inherit the systemic issues discussed in \cref{sec:large_scale_photo_collections}, such as often unreliable intrinsic and extrinsic parameters. Furthermore, the extreme variations in illumination and contrast inherent in these images pose significant challenges for downstream tasks that rely on photometric consistency or differentiable rendering losses.

A further step toward modern image collection and benchmarking was made with Graz4K \cite{durso2026sandesc}, an urban 4K dataset collected with pre-calibrated intrinsics and reconstructed with COLMAP. Images were collected in a reduced time span, ensuring consistent lighting conditions and complete viewpoint coverage. 
Except for the MegaDepth Air-to-Ground test set \cite{chen2025rdd} and CVD-SfM~\cite{li2025cvd}, none of these benchmarks include captured aerial imagery, which is fundamental for testing models in a practical scenario.

\begin{figure*}[t]
    \centering

    % Left image (top aligned)
    \begin{minipage}[t]{0.45\textwidth}
        \vspace{0pt}
        \centering
        \includegraphics[width=\textwidth]{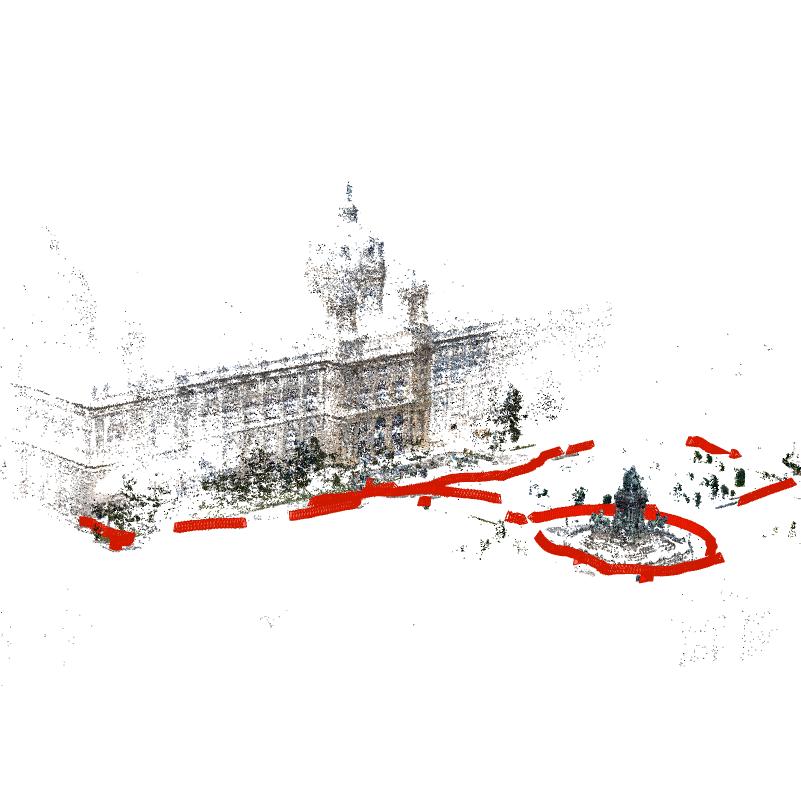}
    \end{minipage}
    \hfill
    % Right 2x4 grid (top aligned) - THIS IS THE MODIFIED PART
    \begin{minipage}[t]{0.525\textwidth}
        \vspace{0pt}
        \centering

        % --- Row 1 ---
        \begin{minipage}[t]{0.24\textwidth}
            \vspace{0pt}
            \includegraphics[width=\textwidth]{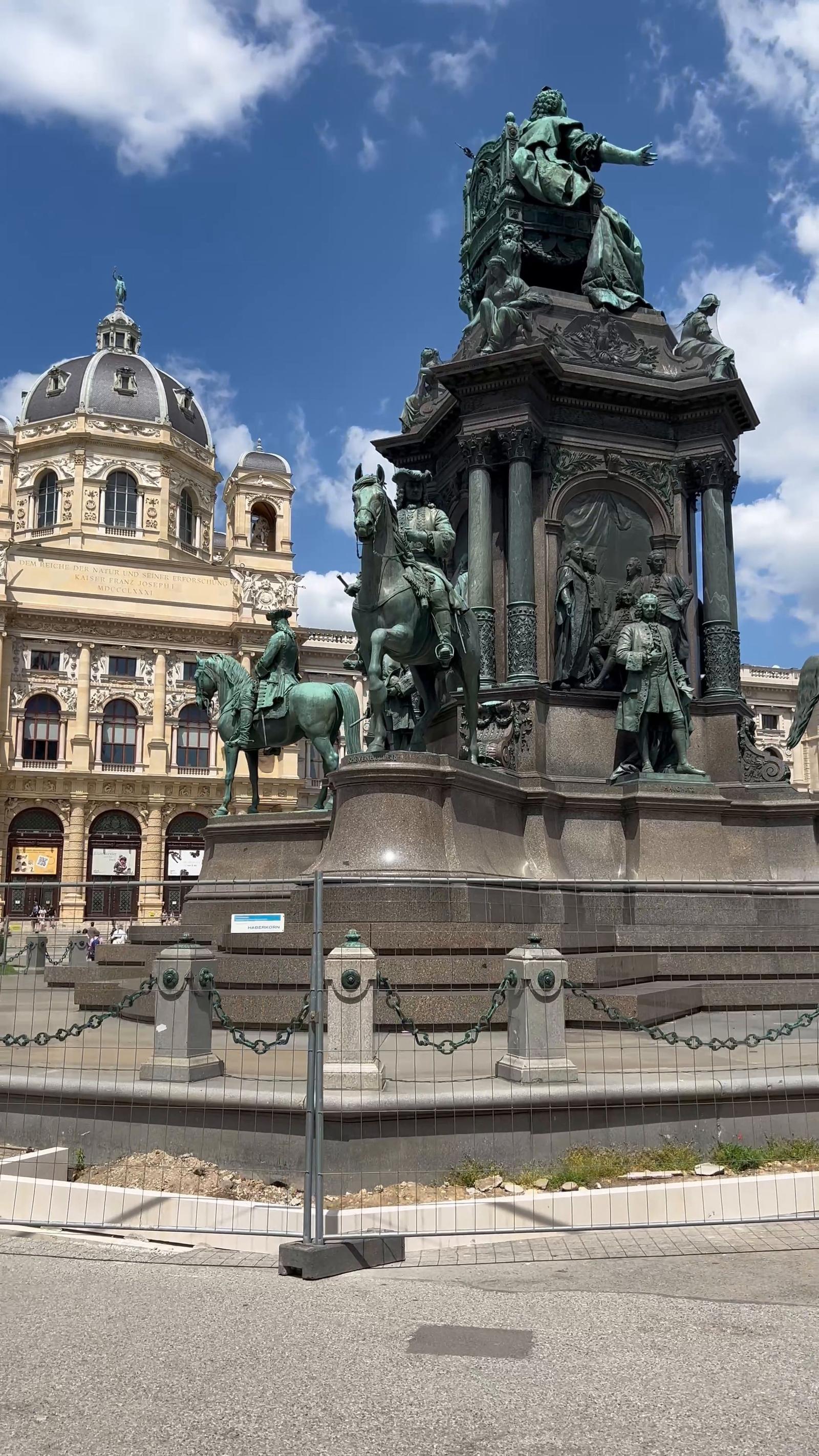}
        \end{minipage}
        \hfill
        \begin{minipage}[t]{0.24\textwidth}
            \vspace{0pt}
            \includegraphics[width=\textwidth]{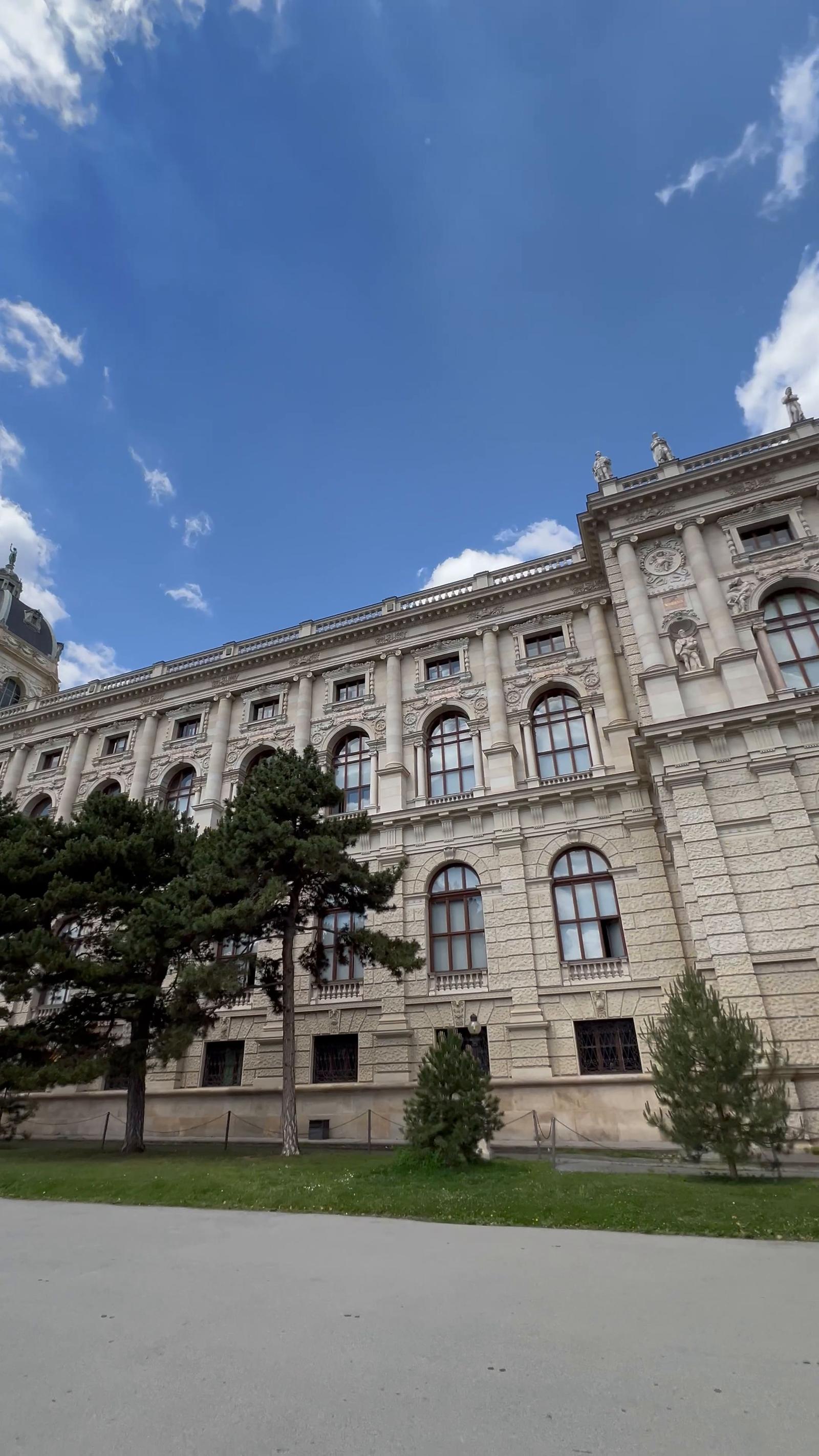}
        \end{minipage}
        \hfill
        \begin{minipage}[t]{0.24\textwidth}
            \vspace{0pt}
            % FIXME: Add your 3rd image path
            \includegraphics[width=\textwidth]{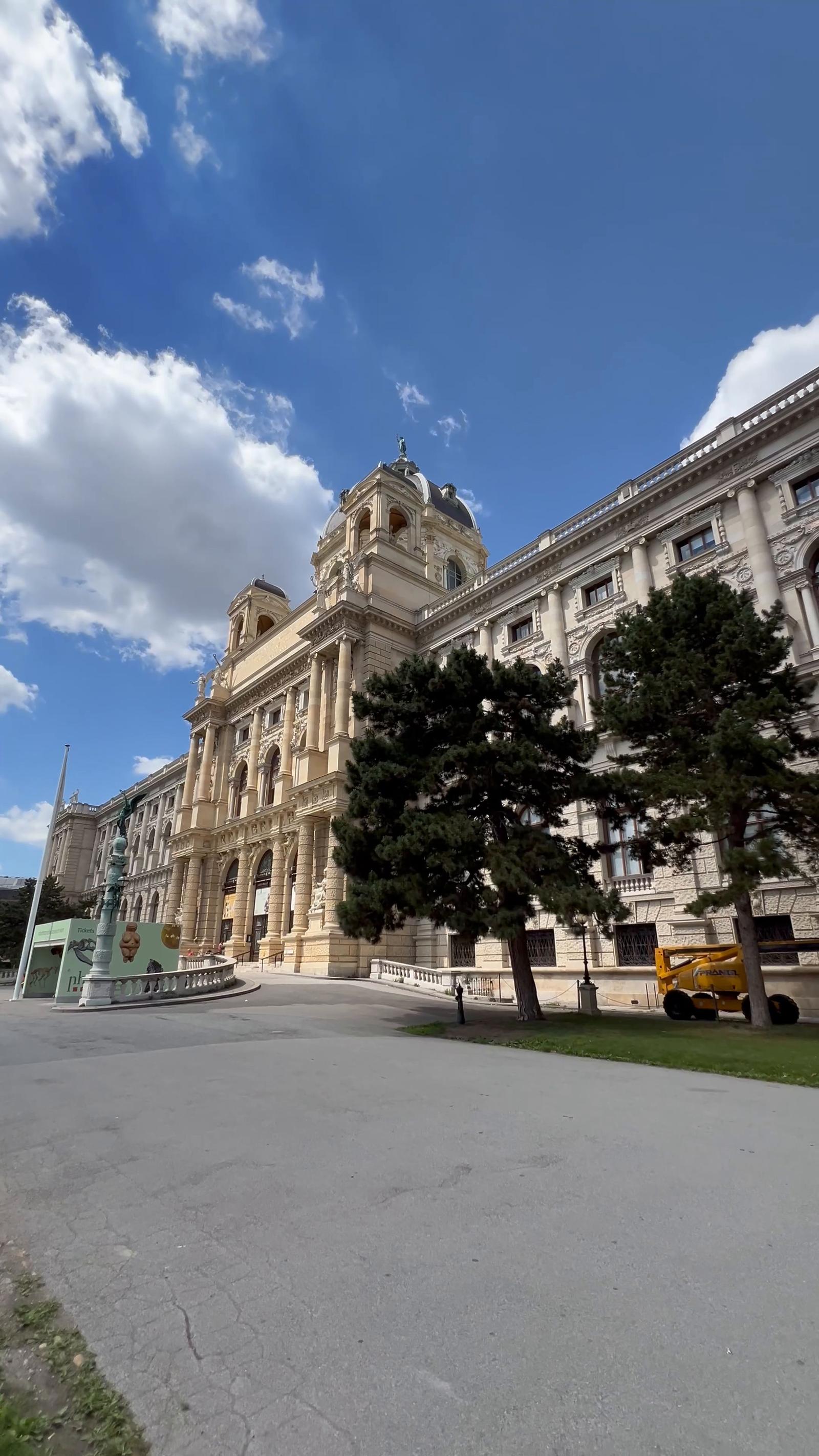}
        \end{minipage}
        \hfill
        \begin{minipage}[t]{0.24\textwidth}
            \vspace{0pt}
            % FIXME: Add your 4th image path
            \includegraphics[width=\textwidth]{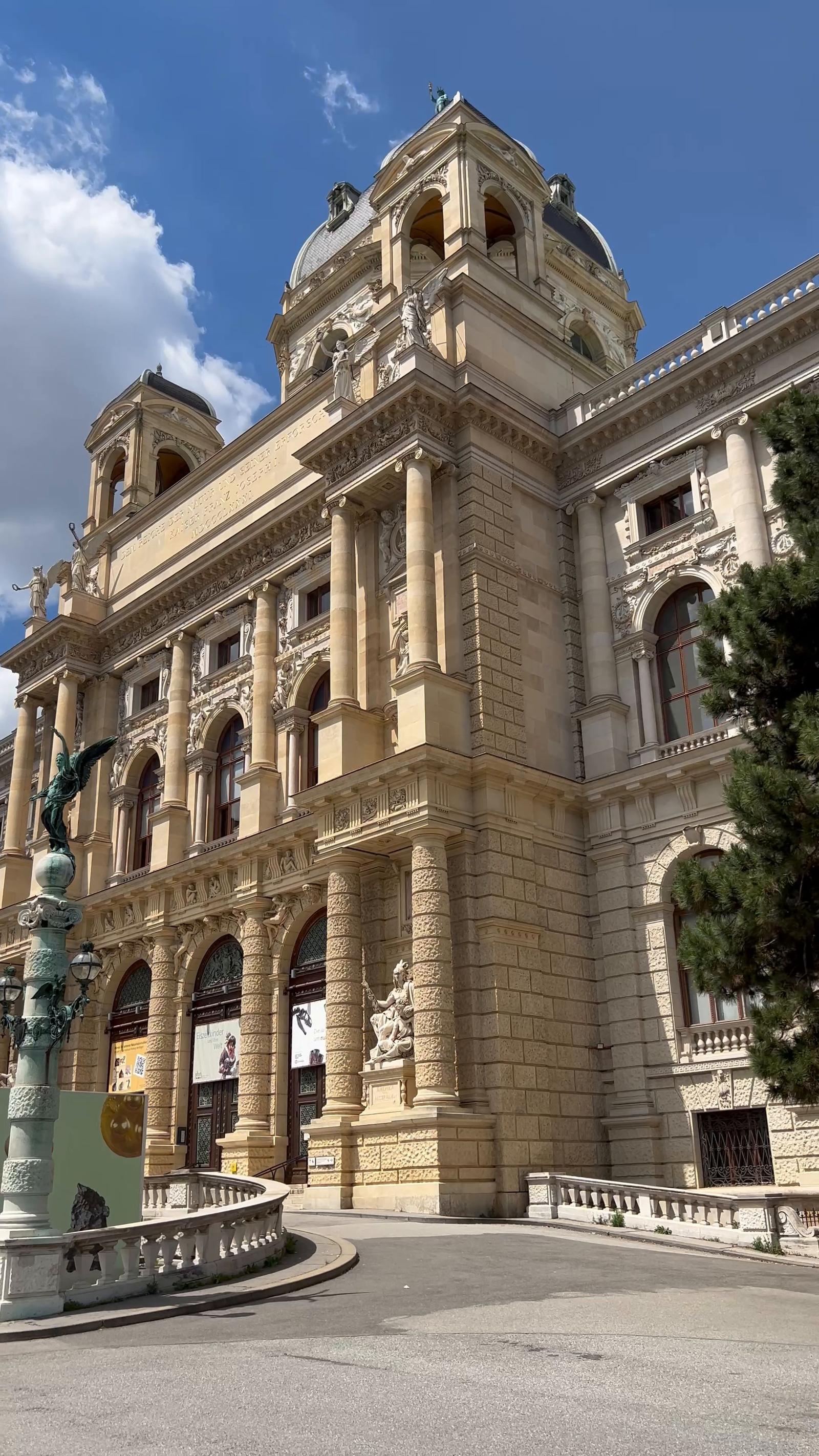}
        \end{minipage}

        \vspace{8pt} % Space between rows

        % --- Row 2 ---
        \begin{minipage}[t]{0.24\textwidth}
            \vspace{0pt}
            \includegraphics[width=\textwidth]{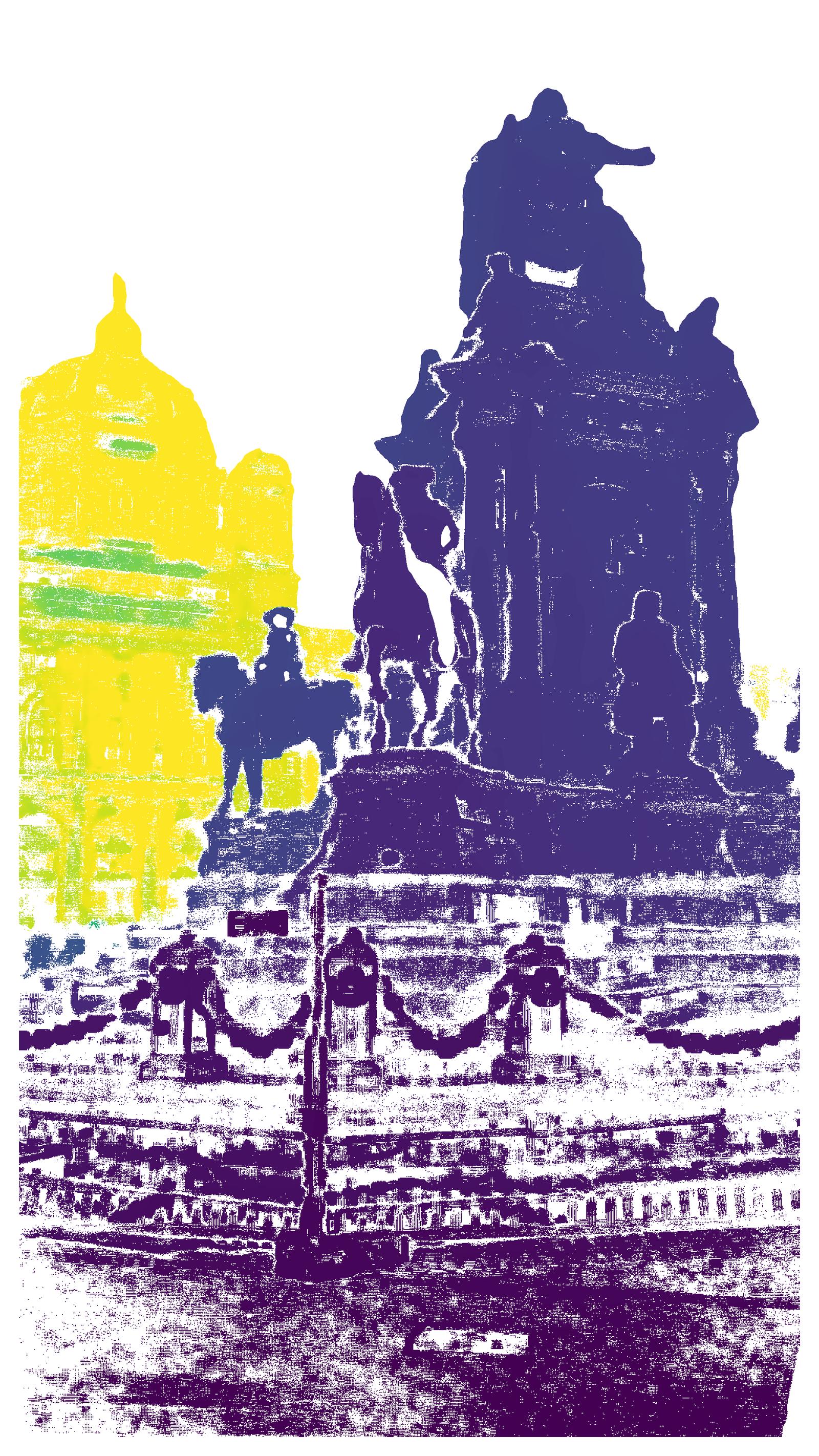}
        \hfill
        \end{minipage}
        \begin{minipage}[t]{0.24\textwidth}
            \vspace{0pt}
            \includegraphics[width=\textwidth]{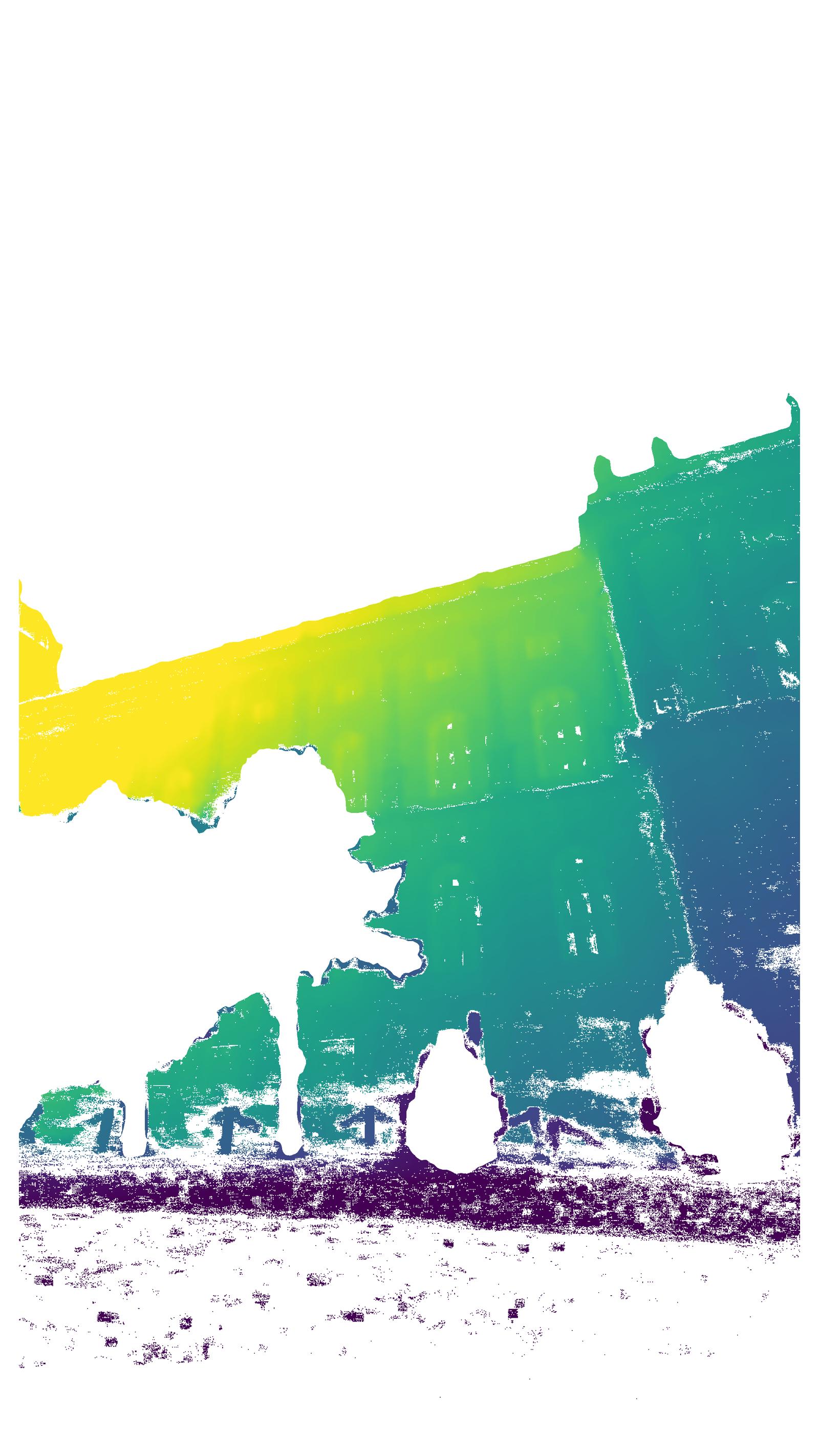}
        \end{minipage}
        \hfill
        \begin{minipage}[t]{0.24\textwidth}
            \vspace{0pt}
            % FIXME: Add your 7th image path
            \includegraphics[width=\textwidth]{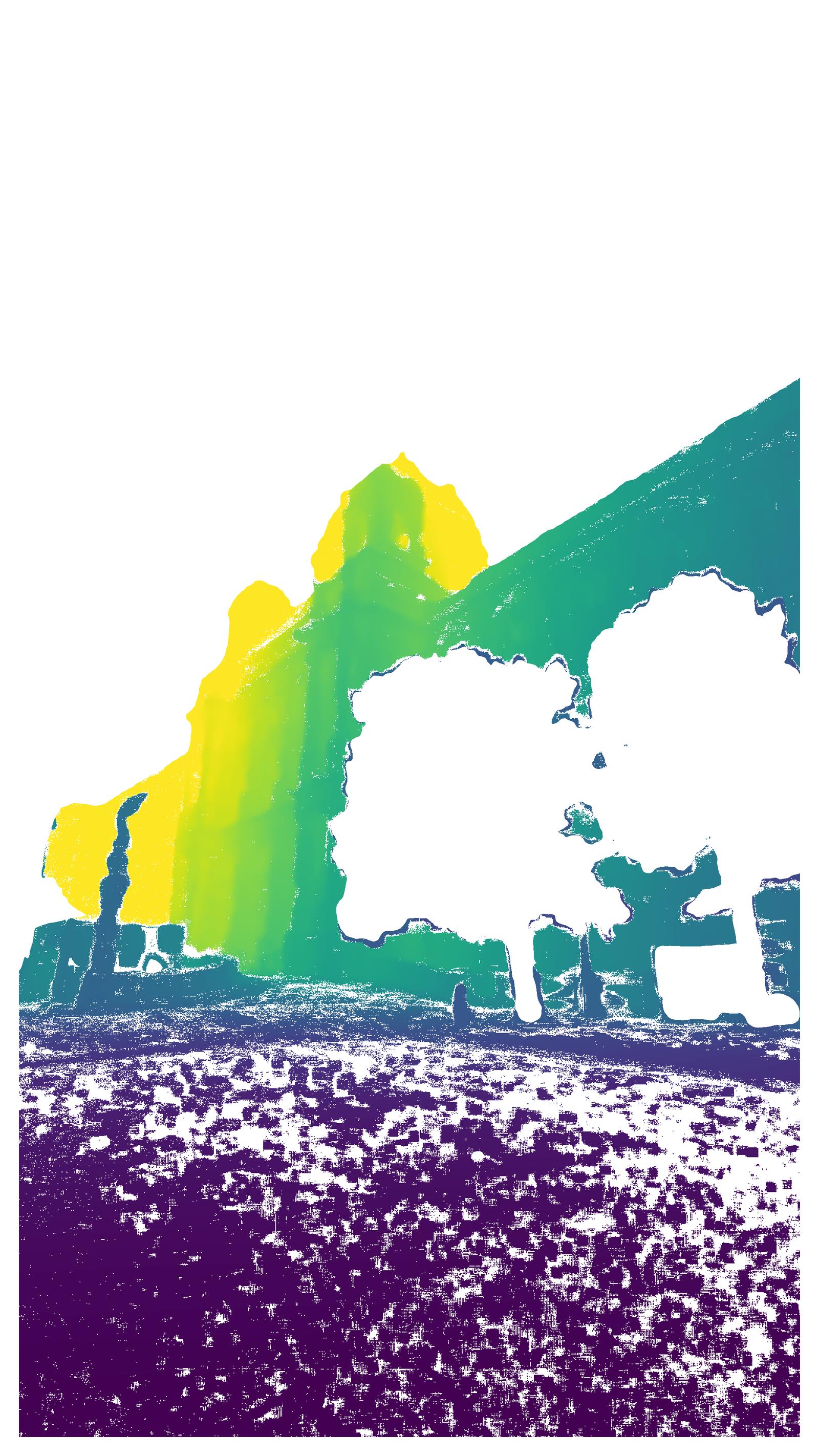}
        \end{minipage}
        \hfill
        \begin{minipage}[t]{0.24\textwidth}
            \vspace{0pt}
            % FIXME: Add your 8th image path
            \includegraphics[width=\textwidth]{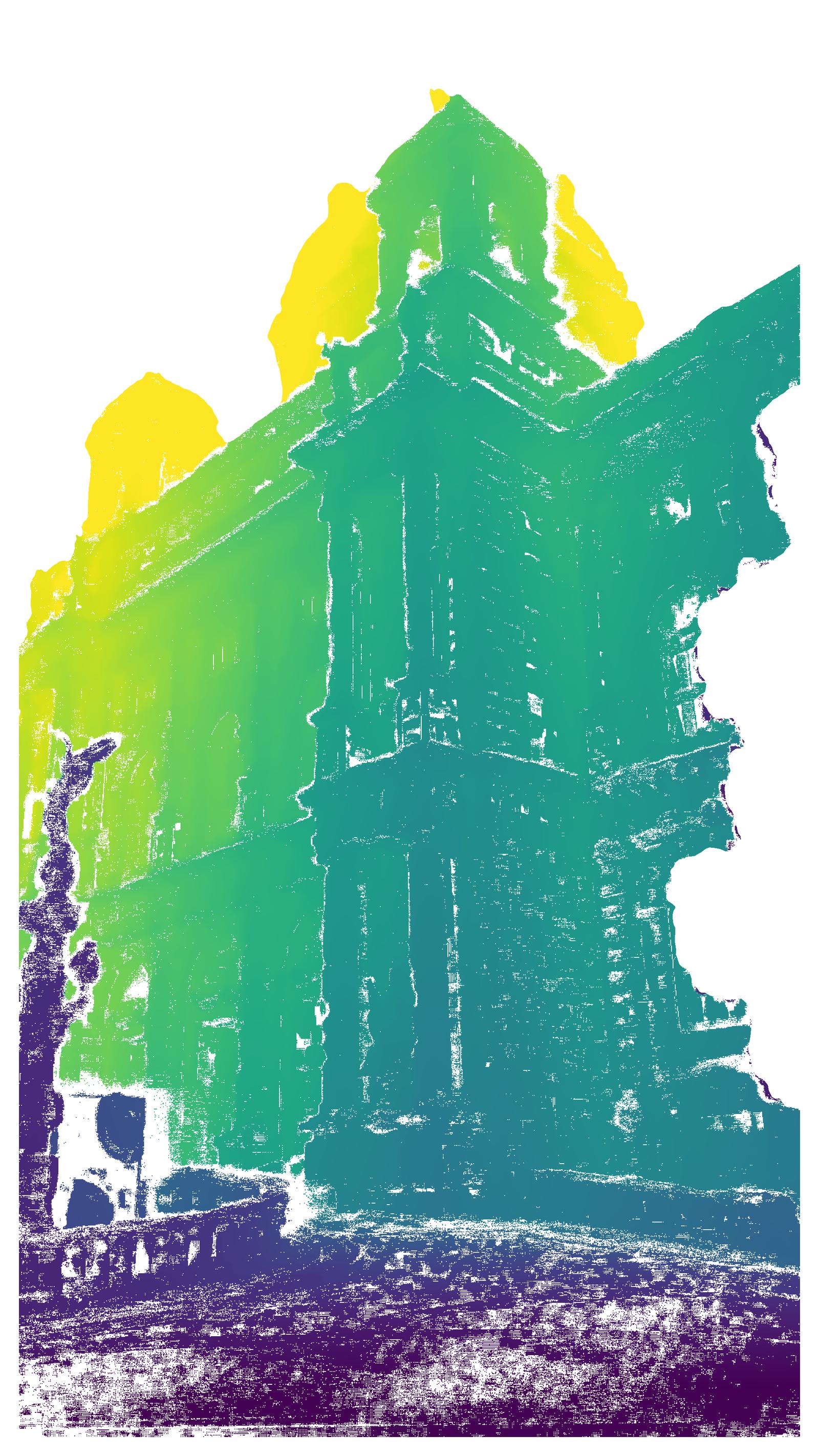}
        \end{minipage}
    \end{minipage}
    \caption{\textbf{Example Scene from TerraSky3D.} Left: Sparse reconstruction of the Natural History Museum, Vienna, Austria. Right: The first row shows example images, and the second row shows the corresponding semantically filtered depth maps.} 
    \label{fig:vienna}

\end{figure*}
\section{Pipeline}
\label{sec:method}

\subsection{Data Collection and Scene Diversity}
\label{sec:data_collection}

Data are collected primarily using various smartphones for ground capturing and a DJI Mavic 3 drone for aerial ones. For each scene, one to four cameras were used. This device redundancy significantly facilitates robust geometry estimation and allows for more precise intrinsic parameter refinement during bundle adjustment. 

The original captures consist of videos at 4K resolution and 30fps, collected across 26 cities in 10 countries, as illustrated in \cref{fig:map}. From the videos, approximately 50,000 images were extracted by uniform sampling.
The dataset includes 30 scenes with successfully registered images from both ground-level and aerial perspectives, a combination currently rare in public benchmarks. An additional 5 scenes feature aerial-only footage, while the remaining 115 scenes provide ground-level perspectives. The dataset encompasses a diverse array of European landmarks and urban environments, including medieval castles, historical buildings, arches, statues, fountains, bridges, dams, piers, shrines, as well as unique viewpoints such as lakeside villas captured from water perspectives. 
Overall, the dataset provides over 2.5 million image pairs with a global mean reprojection error of only 0.8 pixels. 

\subsection{Geometry Estimation}
\label{sec:geometry_estimation}
Despite the recent emergence of several alternatives, including traditional systems such as GLOMAP~\cite{pan2024global} and FastMap~\cite{li2025fastmap}, as well as learning-based approaches like VGGT~\cite{wang2025vggt}, we found COLMAP~\cite{schonberger2016structure} to remain the most robust and reliable solution for our setting. 

We pre-calibrate each camera using a ChArUco board and OpenCV~\cite{bradski2000opencv}, achieving sub-pixel reprojection errors. The resulting intrinsics are stored using the \texttt{SIMPLE\_RADIAL} model. Finally, after running COLMAP with these pre-calibrated intrinsics, we manually inspect each scene to ensure all cameras are correctly registered and positioned in locations consistent with the corresponding RGB images.

\subsection{Depth Computation}
\label{sec:depth}

We utilize Adaptive Patch Deformation MVS (APD-MVS) introduced by Wang et al. \cite{Wang2023apd} to address depth estimation failures in large, textureless regions. Unlike the standard PatchMatch algorithm used in COLMAP (and MegaDepth), which employs a fixed-size square matching window, APD-MVS dynamically adjusts its receptive field achieving state-of-the-art performance on benchmarks such as ETH3D~\cite{schops2017multi} and Tanks and Temples~\cite{Knapitsch2017}.

\begin{figure}[t]
    \centering
    \includegraphics[width=\linewidth]{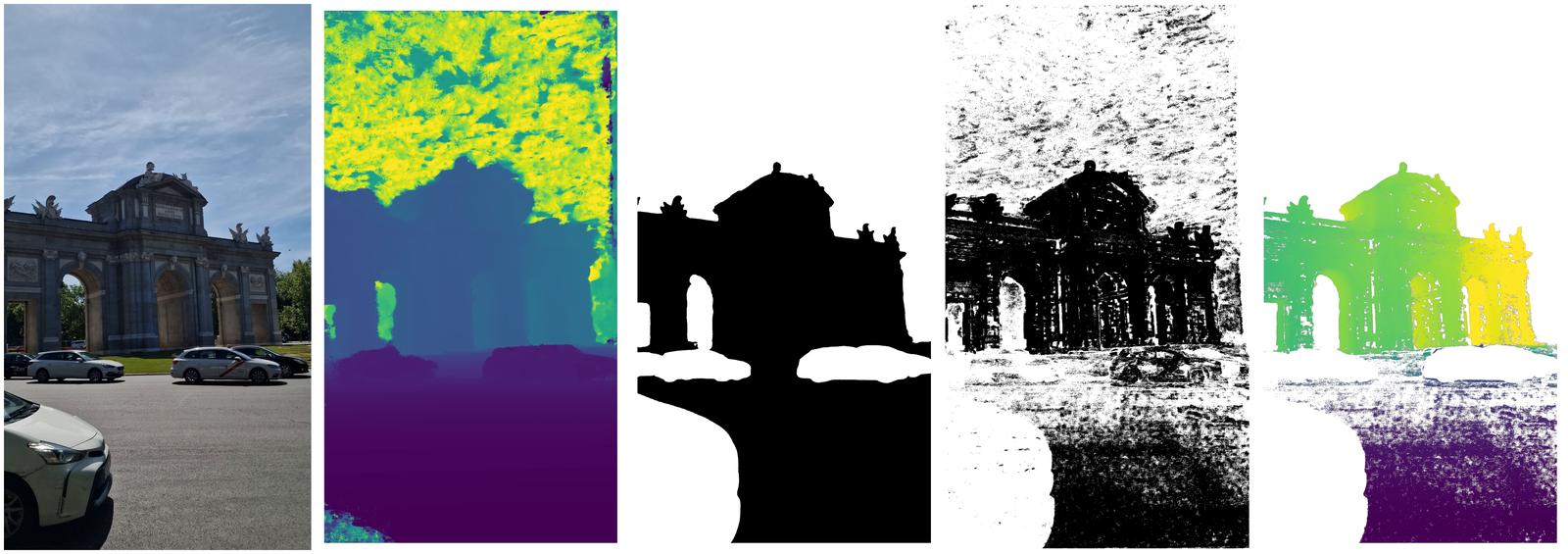}
        \caption{\textbf{Visualization of the Depth Filtering Process.} From left to right: original RGB image, raw depth map from APD-MVS, semantic mask, APD-MVS confidence mask, and the filtered depth map. The scene depicted is the \mbox{Arch of Victory}, Madrid, Spain.}
    \label{fig:depth}
\end{figure}

To refine the reconstruction in areas that violate MVS assumptions, such as the sky, vegetation, or transient objects, we implement a semantic filtering post-processing step. We employ Mask2Former~\cite{cheng2022mask2former} to segment these regions from the source RGB images, generating a unified binary mask used to prune the resulting depth maps (\cref{fig:depth}). Furthermore, we release the original APD-MVS confidence masks to facilitate additional downstream filtering.

\subsection{Format and Split} %this also might be removed
For each scene, the dataset provides RGB images organized in camera-specific folders, along with the corresponding COLMAP sparse reconstruction and filtered depth maps. While the entire dataset can be utilized for training large-scale models, we propose an official data split to ensure consistent evaluation. Specifically, the scenes listed in the lower section of \cref{tab:nvs} are designated as the official test set.
\section{Experiments}
\label{sec:experiments}

\begin{figure*}[t]
    \centering

    % Left image (top aligned)
    \begin{minipage}[t]{0.45\textwidth}
        \vspace{0pt}
        \centering
        \includegraphics[width=\textwidth]{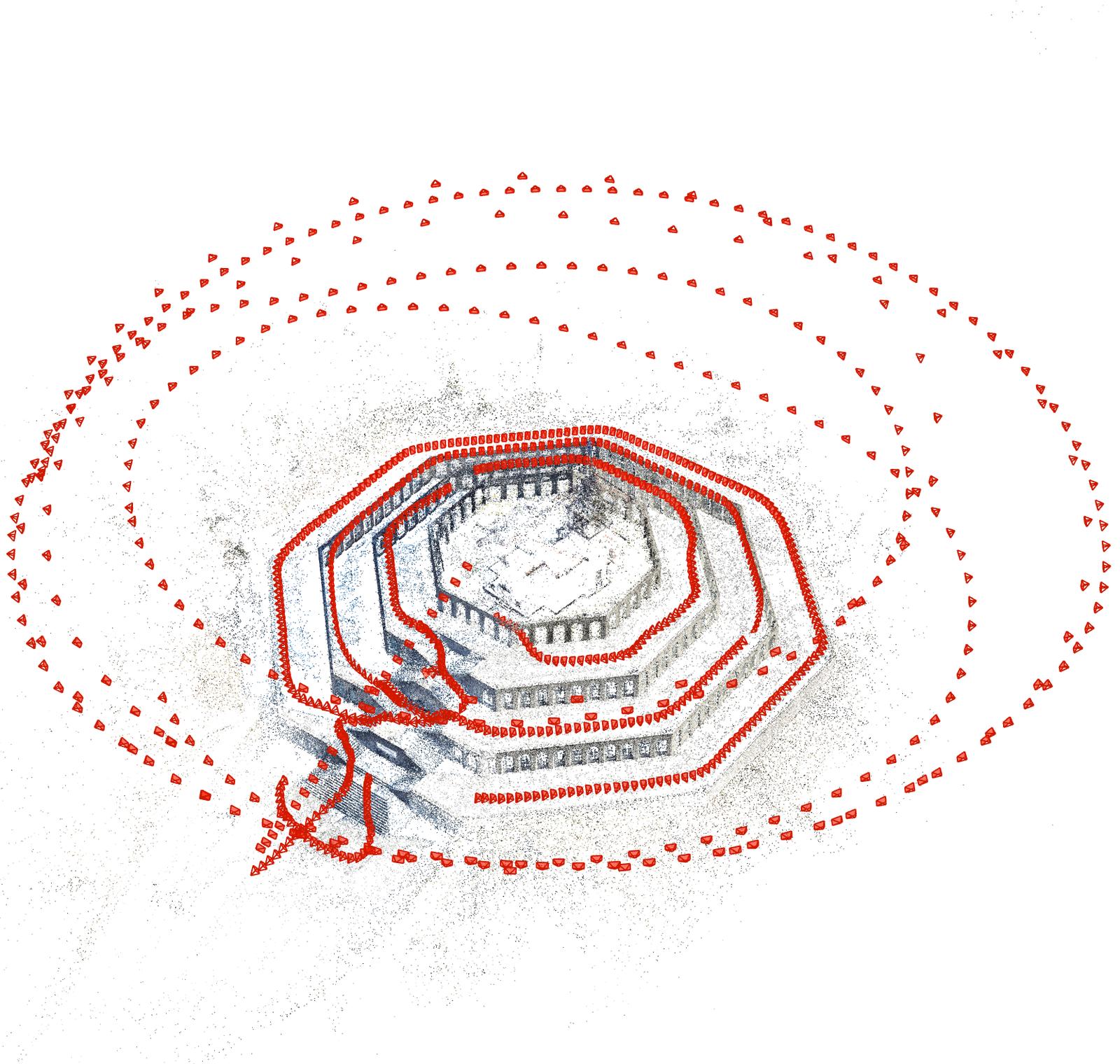}
    \end{minipage}
    \hfill
    % Right 2x4 grid (top aligned) - THIS IS THE MODIFIED PART
    \begin{minipage}[t]{0.53\textwidth}
        \vspace{0pt}
        \centering

    % --- Row 1 ---
    \begin{minipage}[t]{0.24\textwidth}
        \vspace{0pt} % Ensures minipage aligns at the very top
        \includegraphics[width=\textwidth]{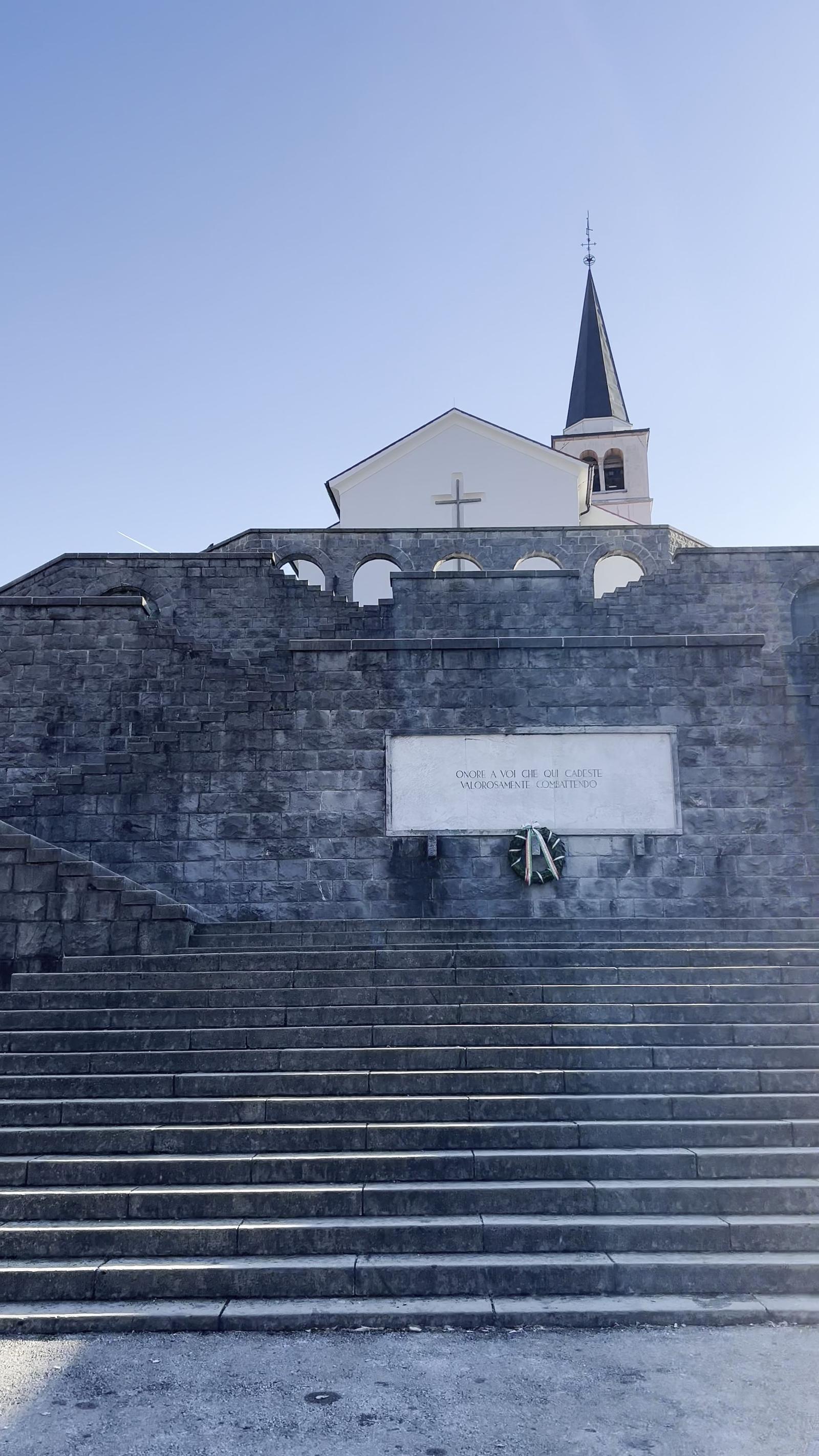}
    \end{minipage}
    \hspace{2pt} % <-- KEY CHANGE: Adds equal horizontal space
    \begin{minipage}[t]{0.24\textwidth}
        \vspace{0pt}
        \includegraphics[width=\textwidth]{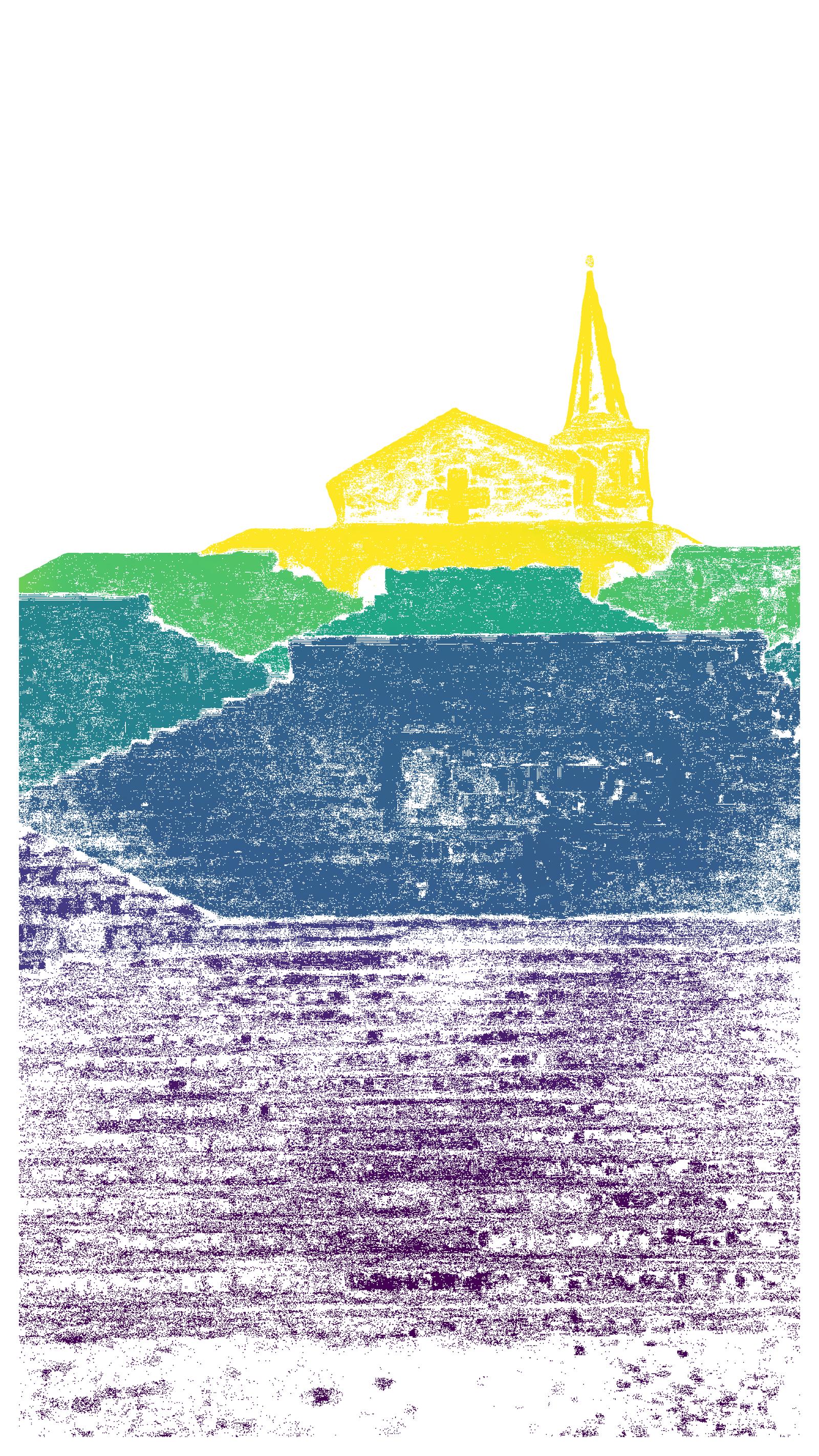}
    \end{minipage}
    \hspace{2pt} % <-- KEY CHANGE: Adds equal horizontal space
    \begin{minipage}[t]{0.375\textwidth}
        \vspace{0pt}
        \includegraphics[width=\textwidth]{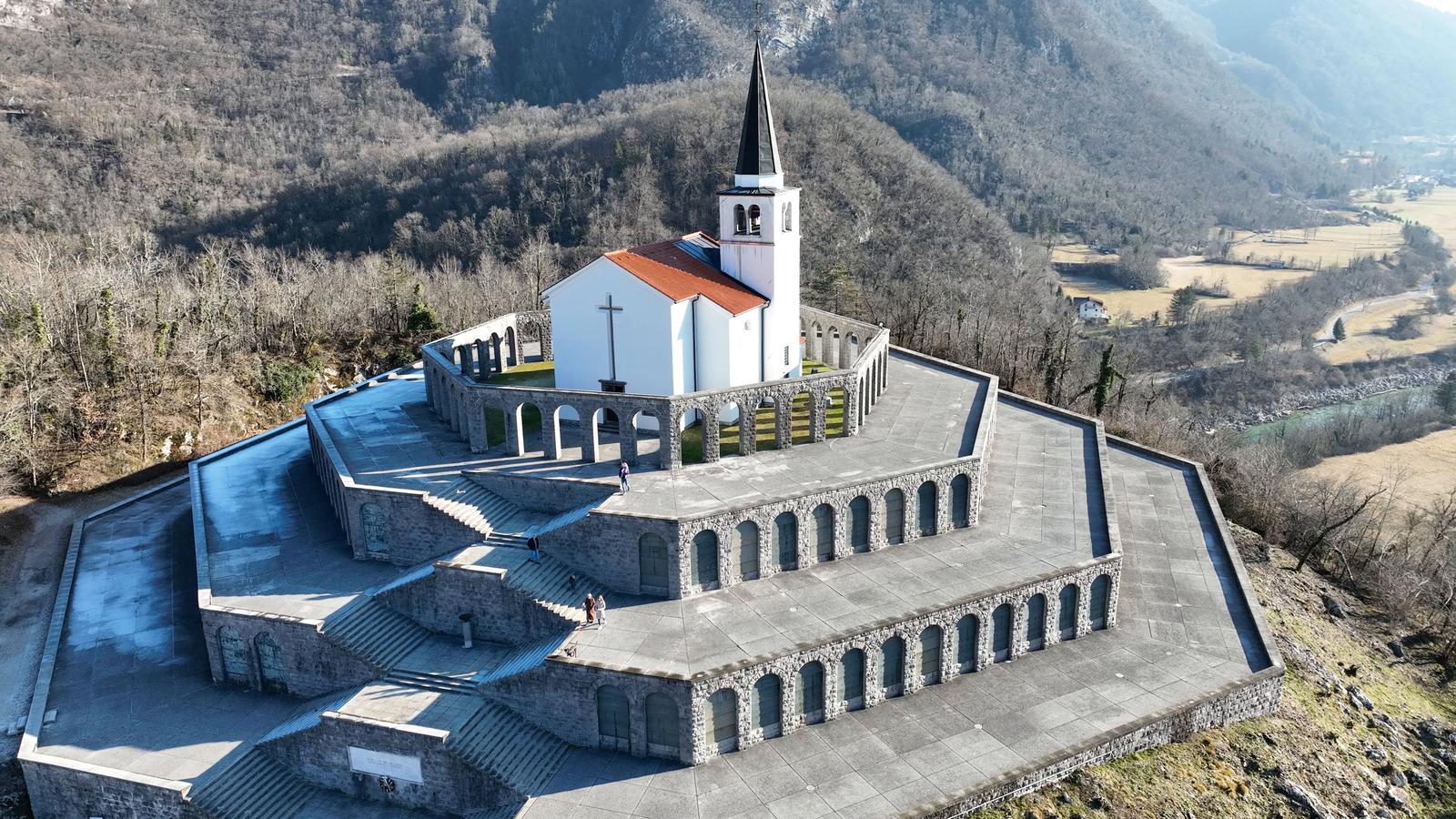}
        \vspace{2pt} % Adds space between the stacked images in this column
        \includegraphics[width=\textwidth]{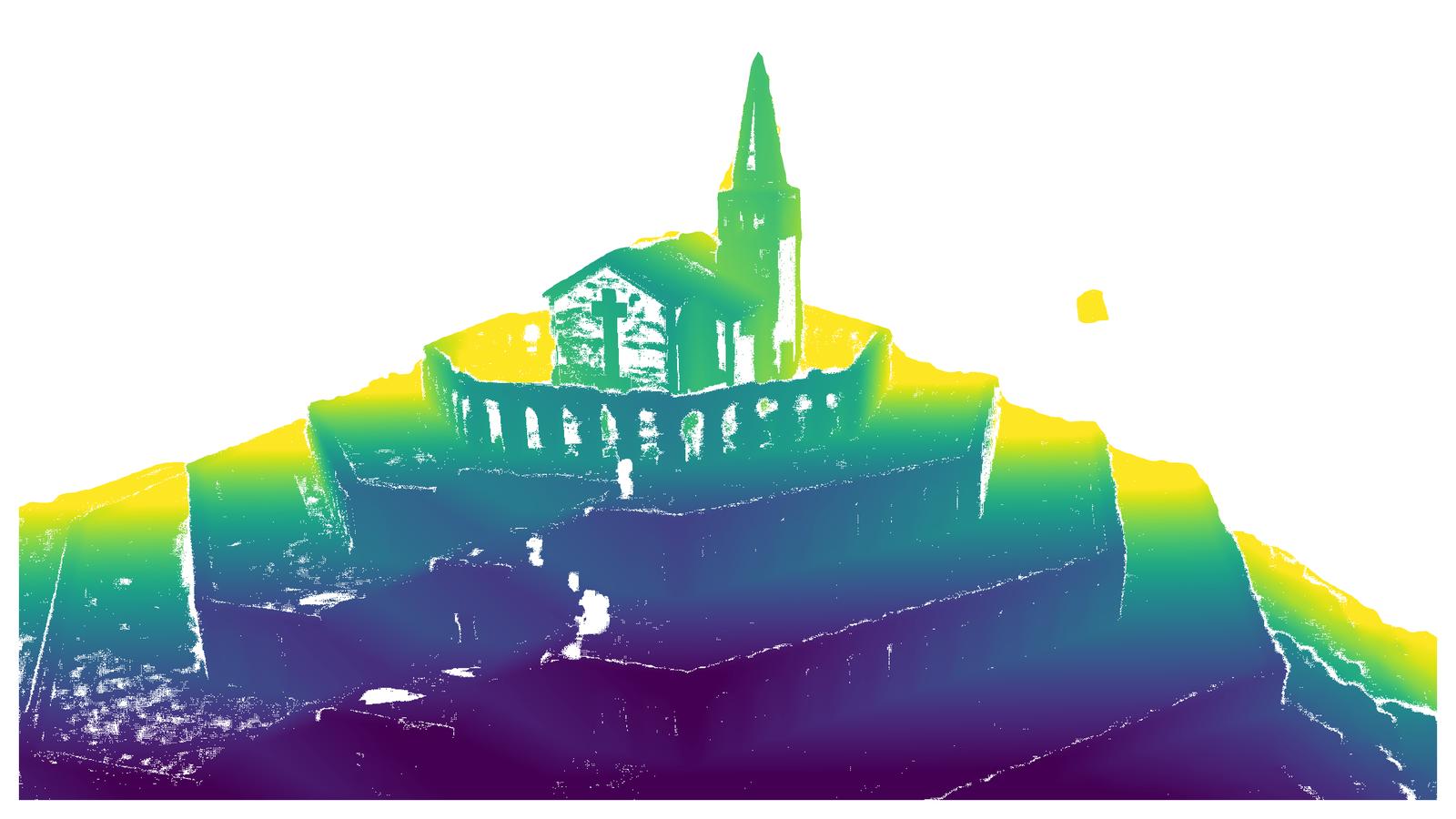}
    \end{minipage}
    
    \vspace{8pt} % <-- KEY CHANGE: This single command creates all the vertical space between Row 1 and Row 2
    
    % --- Row 2 ---
    \begin{minipage}[t]{0.24\textwidth}
        \vspace{0pt}
        \includegraphics[width=\textwidth]{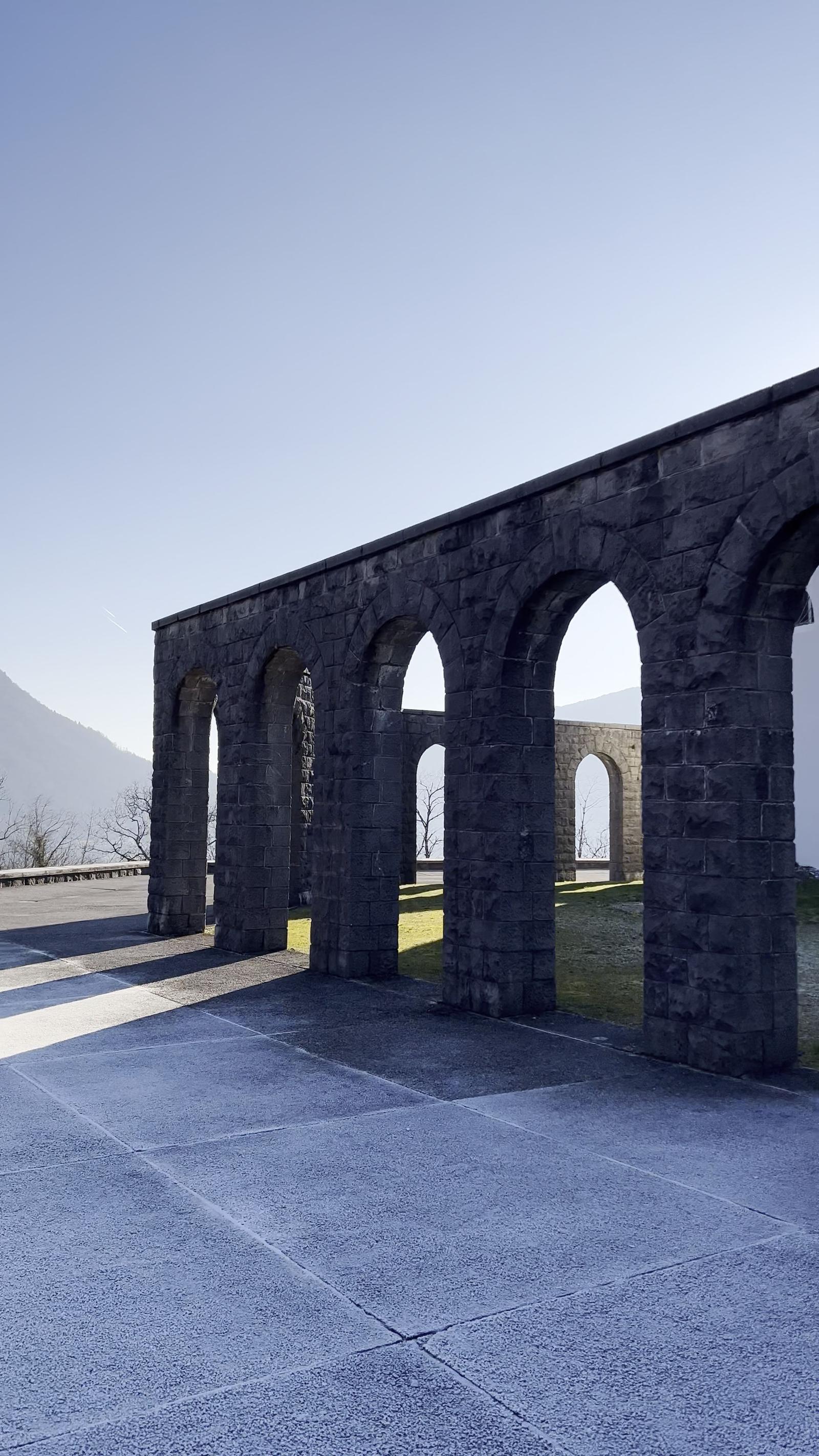}
    \end{minipage}
    \hspace{2pt} % <-- KEY CHANGE: Adds equal horizontal space
    \begin{minipage}[t]{0.24\textwidth}
        \vspace{0pt}
        \includegraphics[width=\textwidth]{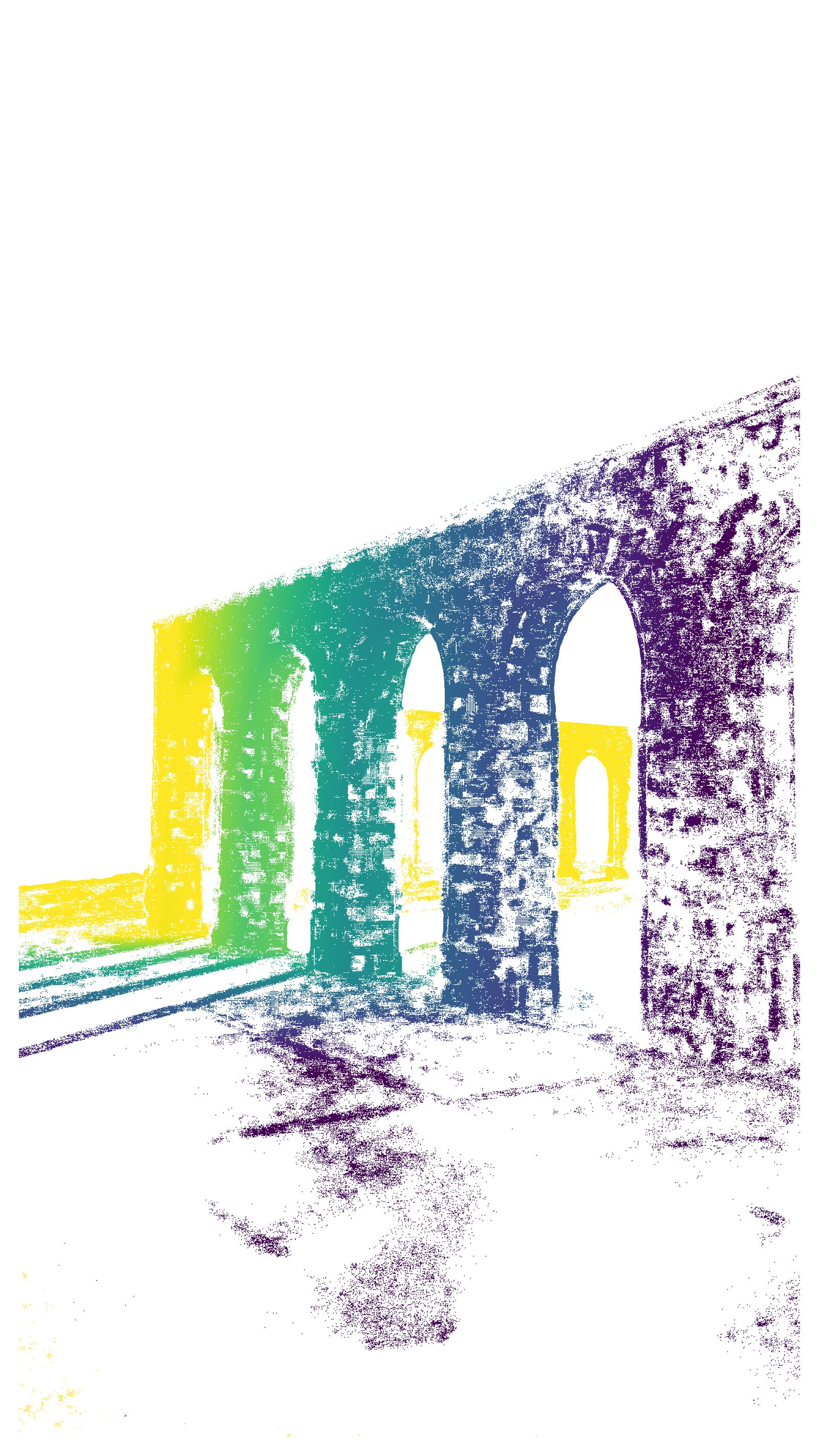}
    \end{minipage}
    \hspace{2pt} % <-- KEY CHANGE: Adds equal horizontal space
    \begin{minipage}[t]{0.375\textwidth}
        \vspace{0pt}
        \includegraphics[width=\textwidth]{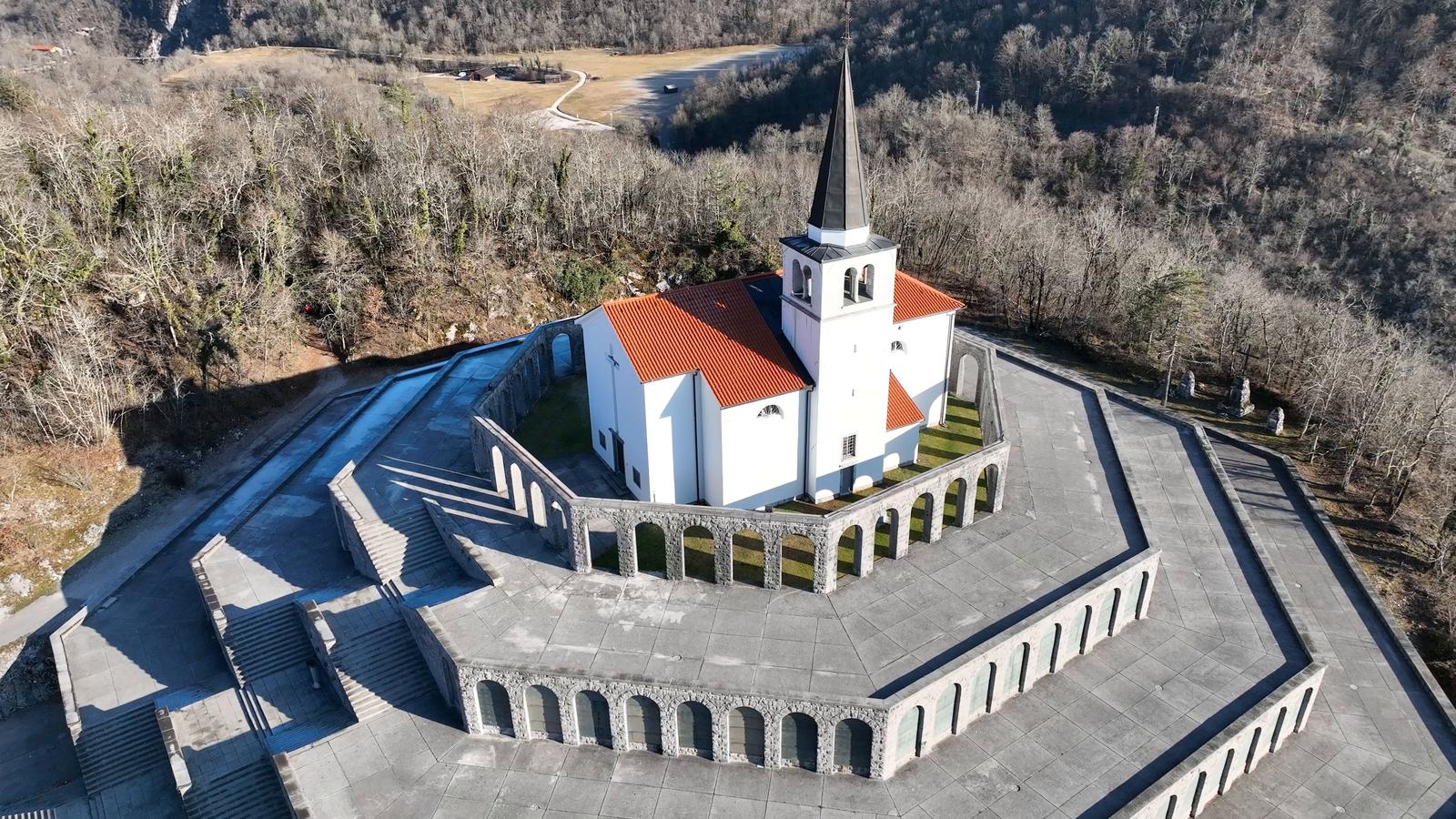}
        \vspace{2pt} % Adds space between the stacked images in this column
        \includegraphics[width=\textwidth]{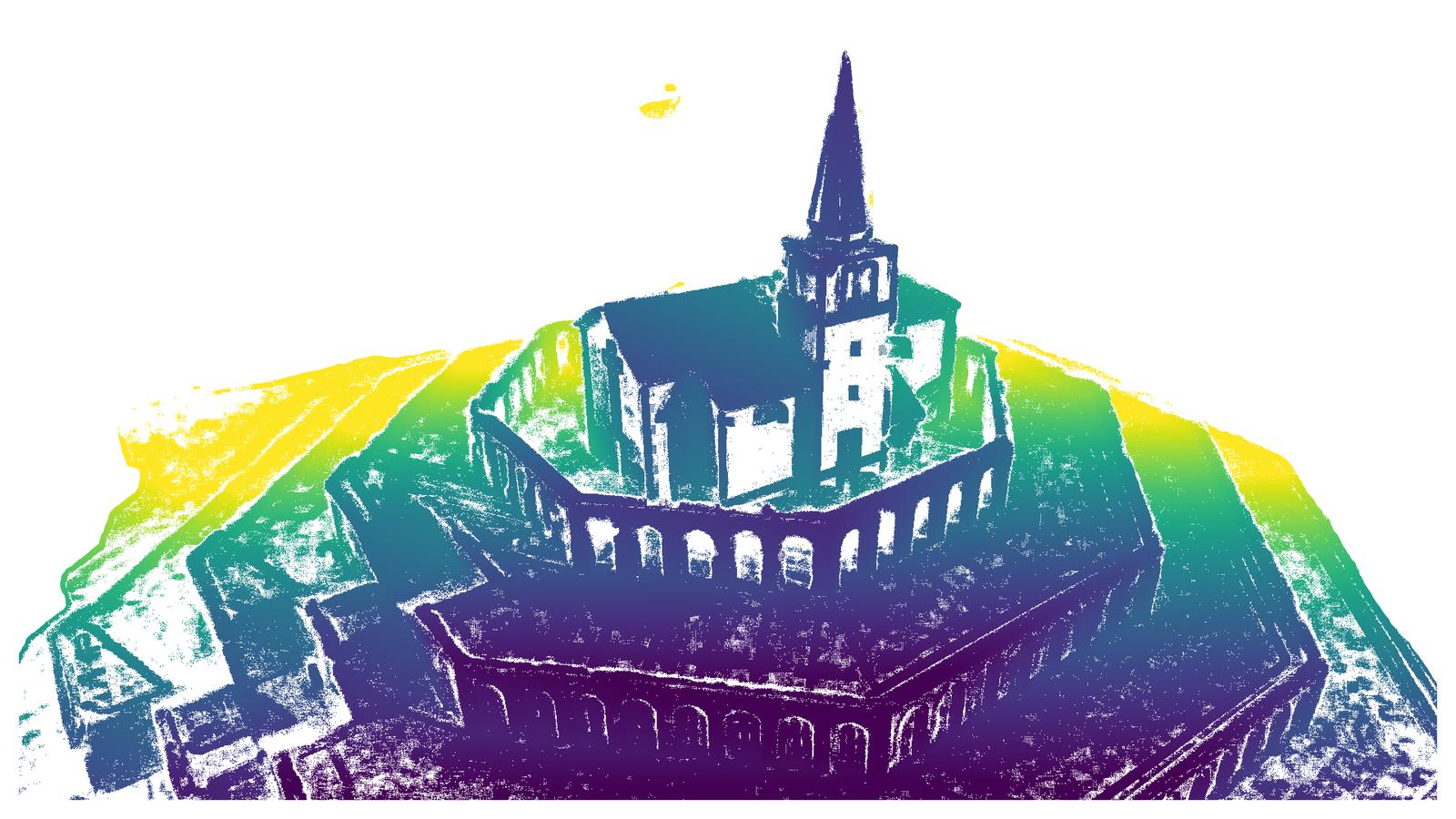}
    \end{minipage}

    \end{minipage} 
    \caption{\textbf{Example Scene from TerraSky3D.} Left: Sparse reconstruction of the Italian Charnel House, Kobarid, Slovenia. Right: Representative images collected from aerial and ground perspectives, shown with their corresponding semantically filtered depth maps.} 
    \label{fig:caporetto}
\end{figure*}

\subsection{Stereo Pose Estimation}
\label{sec:rel_pose}
We benchmark state-of-the-art models on our test set to evaluate their performance across modern ground, aerial, and aerial and ground (mixed) scenarios.
We consider two different categories of models: deep sparse feature extractors \cite{detone2018superpoint, tyszkiewicz2020disk, edstedt2023dedode, chen2025rdd, zhao2023aliked, durso2026sandesc} and deep matchers \cite{lindenberger2023lightglue, leroy2024grounding, edstedt2024roma}. 
Input images are downsampled to Full HD resolution for the first category and to the optimal resolution indicated by the respective authors for the second.
Beginning with the COLMAP matching graph, we first identify image pairs with at least 30 inliers. For pairs within the same category (ground-to-ground or aerial-to-aerial), we discard those with fewer than 100 or more than 500 matches to maintain a balanced graph. In contrast, we retain all valid pairs for the mixed aerial and ground case to ensure sufficient connectivity across viewpoints.
The test set comprises a total of over 43,000 pairs distributed as 74.9\% ground, 10.7\% aerial, and 14.4\% mixed. 
We report performance using the Area Under the Curve (AUC) of the pair-wise pose errors, where the error is defined as the maximum of the rotation error $\Delta R$ and the translation error $\Delta t$ as in \cite{lindenberger2023lightglue, durso2026sandesc, zhao2023aliked, edstedt2023dedode}. We evaluate this up to a threshold of 5$^\circ$. 

Quantitative results are summarized in \cref{tab:pose}, categorized by pair type (e.g., the ground category includes all and only pairs where both images are captured from a ground level). The mean score is computed by averaging the results of all three scenarios with equal weights.

We observe that recent sparse feature extractors exhibit highly similar performance in the \textit{Ground} category, suggesting a plateau in current methodologies or training data. Conversely, the \textit{Aerial} category reveals a broader spectrum of results, indicating that certain methods are better suited for aerial-to-aerial matching. The \textit{Mixed} scenario remains the most challenging due to extreme viewpoint variations, as evidenced by significantly lower scores across all methods. Notably, the combination of ALIKED~\cite{zhao2023aliked} and SANDesc~\cite{durso2026sandesc} emerges as the top-performing approach, with a substantial margin in both the \textit{Aerial} and \textit{Mixed} scenarios. 

Notably, retraining SANDesc on our TerraSky3D (TS3D) dataset using the original protocol allows the model to cover the traditional ground-level cases found in MegaDepth (MD) while expanding its capabilities to include both aerial-to-aerial and, most importantly, cross-view aerial and ground scenarios. While accuracy remains comparable to the baseline, namely SANDesc (MD), in ground-to-ground and aerial-to-aerial tasks, the model achieves a significant 1.8-point improvement in AUC@5$^\circ$ within the \textit{Mixed} scenario. This gain showcases the effectiveness of our specialized training data for cross-view tasks.

Although LightGlue~\cite{lindenberger2023lightglue} achieves performance levels comparable to ALIKED+SANDesc (TS3D), in both \textit{Aerial} and \textit{Mixed} scenarios, deep matchers generally outperform sparse methods.
MASt3R~\cite{leroy2024grounding} shows improvements over the LightGlue baselines, while RoMa~\cite{edstedt2024roma} emerges as the top-performing model across all categories. Its robustness is derived from a dense-warping paradigm rather than the traditional point-to-point matching, yielding up to a $10\times$ increase in inlier counts which significantly stabilizes geometry estimation. However, this superior accuracy entails a notable computational trade-off, as RoMa remains the slowest method among the evaluated suite, both for its size and quadratic cost.

\begin{table}[t]
    \centering  
    \caption{\textbf{Relative Pose Estimation Performance on the TerraSky3D Test Set.} We report performance in terms of AUC@$5^\circ$. The keypoints budget is set to 4096. MD and TS3D denote models trained on MegaDepth and TerraSky3D.} 
    \label{tab:pose} 
    \resizebox{\columnwidth}{!}{% Adjust table size to column width
    \begin{tabular}{llccc|c} 
        \toprule
        \multicolumn{2}{c}{} & \multicolumn{4}{c}{\textbf{AUC@5}$\uparrow$} \\
        \cmidrule(lr){3-6} % Horizontal line only over data columns
        & \textbf{Method} & \textbf{Ground} & \textbf{Aerial} & \textbf{Mixed} & \textbf{Mean} \\
        \midrule
        
        \multirow{8}{*}{\rotatebox{90}{\textbf{Sparse FE}}} & 
          SuperPoint \cite{detone2018superpoint} & 53.2 & 57.9 & 17.7 & 42.9 \\
        & DISK \cite{tyszkiewicz2020disk} & 56.1 & 72.1 & 29.5 & 52.6 \\
        & S-TREK  \cite{santellani2023strek} & 62.0 & 81.8 & 36.9 & 60.2 \\
        & DeDoDe-G  \cite{edstedt2023dedode} & 62.7 & 78.8 & 35.5 & 59.0 \\
        & RDD \cite{chen2025rdd}& 62.3 & 74.5 & 28.7 & 55.2 \\
        & ALIKED \cite{zhao2023aliked} & 62.4 & 70.3 & 27.5 & 53.4 \\
        & \rotatebox[origin=c]{180}{$\Lsh$} w/ SANDesc \cite{durso2026sandesc} (MD) & 64.2 & 85.6 & 42.9 & 64.2 \\
        & \rotatebox[origin=c]{180}{$\Lsh$} w/ SANDesc \cite{durso2026sandesc} (TS3D) & \textbf{64.7} & \textbf{85.7} & \textbf{44.7} & \textbf{65.0} \\
        % maybe add another version expoliting given high-res

        \midrule
        
        \multirow{4}{*}{\rotatebox{90}{\textbf{Matchers}}} & 
        % LoFTR \cite{sun2021loftr} & 62.3 & 31.9 & 29.9 & 41.4 \\
        DISK+LightGlue \cite{lindenberger2023lightglue} & 68.3 & 85.7 & 44.2 & 66.1 \\
        & ALIKED+LightGlue \cite{lindenberger2023lightglue}& 68.4 & 87.6 & 46.4 & 67.5 \\
        & MASt3R \cite{leroy2024grounding} & 73.5 & 88.0 & 55.0 & 72.2 \\
        & RoMa \cite{edstedt2024roma} & \textbf{77.4} & \textbf{93.3} & \textbf{75.6} & \textbf{82.1} \\
     
        \bottomrule
    \end{tabular}
    }

\end{table}

\subsection{End-to-End 3D Reconstruction}
We evaluate three end-to-end learning-based 3D reconstruction pipelines on TerraSky3D, namely VGGT \cite{wang2025vggt}, $\pi^3$ \cite{wang2025pi}, and MapAnything \cite{keetha2025mapanything}. To evaluate the accuracy of a given method, we measure the relative pose distances within the same scenes. In particular, we compare the ground truth relative transformation between two cameras against the relative transformation predicted by the evaluated model. We define the relative distance error $E_{rel}$ for every pair of images $(i, j)$ as follows:
\begin{equation}
    E_{rel}(i, j) = \rho \left( (P_i^{gt})^{-1} P_j^{gt}, \; \hat{P}_i^{-1} \hat{P}_j \right),
    \label{eq:rel_pose_dist}
\end{equation}
where $P^{gt}$ represents the ground truth absolute pose, $\hat{P}$ represents the estimated absolute pose, and the function $\rho(\cdot, \cdot)$ computes the maximum of the angular errors in rotation and translation between the two resulting relative transformations. Finally, we report the AUC@$5^\circ$ in \cref{tab:abs_pose}.

Our evaluation reveals distinct trade-offs between architectural robustness, and peak precision among the different state-of-the-art models. VGGT \cite{wang2025vggt} emerges as the most consistent performer, achieving the highest mean AUC at 5$^\circ$ of 49.7. Its ability to maintain high scores across diverse geographic locations suggests superior generalization capabilities. In contrast, while \textbf{$\pi^3$} \cite{wang2025pi} achieves the highest individual scores on several scenes, it suffers from catastrophic failures. Specifically, its performance significantly drops in aerial and ground scenarios, indicating geometric instability in these specific wide-baseline scenarios. 

\begin{table}[t]
    \centering    
    \caption{\textbf{Absolute Pose Estimation Performance on the TerraSky3D Test Set.} We report performance in terms of AUC@$5^\circ$. MapAny. stands for MapAnything.}
    \label{tab:abs_pose} 
    \resizebox{\columnwidth}{!}{%
    \begin{tabular}{l c cccc} 
        \toprule
        \multicolumn{1}{c}{} & \multicolumn{1}{c}{} & \multicolumn{3}{c}{\textbf{AUC@5}$\uparrow$} \\
        \cmidrule(lr){3-5}
        & \multicolumn{1}{c}{Imgs} 
        & \textbf{VGGT} \cite{wang2025vggt}
        & \textbf{MapAny.} \cite{keetha2025mapanything} 
        & $\boldsymbol{\pi^3}$ \cite{wang2025pi} \\
        \midrule
        Graz Church              & 239 & \textbf{71.4} & 53.7 & \phantom{0}0.0 \\
        Graz Townhall            & 335 & 17.9 & \textbf{51.7} & 26.9 \\
        Graz University          & 382 & \textbf{33.1} & 26.4 & 12.0 \\
        Munich Frauenkirche      & 123 & 68.2 & 65.2 & \textbf{77.3} \\
        Munich Marienplatz       & 246 & 56.4 & 60.0 & \textbf{68.5} \\
        Munich Theatine Church   & 120 & 67.0 & 49.5 & \textbf{70.3} \\
        Salzburg Andrakirche     & 114 & 70.3 & 44.7 & \textbf{82.4} \\
        Salzburg Rechte Altstadt & \phantom{0}58 & 68.9 & 65.2 & \textbf{78.5} \\
        Udine Devils Bridge      & 339 & \textbf{37.2} & 23.9 & \phantom{0}0.3 \\
        Udine Fagagna Church     & 274 & \textbf{49.8} & 40.8 & \phantom{0}0.9 \\
        Udine Villalta Castle    & 688 & \phantom{0}0.5 & \phantom{0}\textbf{2.2} & \phantom{0}0.0 \\
        Vienna State Opera       & 106 & 55.2 & \textbf{57.5} & 24.2 \\
        \midrule
        Mean                     &  & \textbf{49.7} & 45.0 & 36.8 \\
        \bottomrule
    \end{tabular}
    }
    
\end{table}

\subsection{Novel View Synthesis} 
To evaluate Novel View Synthesis (NVS), we establish a benchmark by uniformly sampling 1 every 8 images to create the test set, then we train 3D Gaussian Splatting (3DGS)~\cite{kerbl3Dgaussians} on the remains images for 30,000 iterations.

\cref{tab:nvs} reports the evaluation of the synthesized novel views in terms of Learned Perceptual Image Patch Similarity (LPIPS) \cite{zhang2018unreasonable}, Peak Signal-to-Noise Ratio (PSNR), and Structural Similarity Index Measure (SSIM). LPIPS assesses perceptual similarity, where lower scores correspond to better visual alignment with human perception. Conversely, for PSNR and SSIM, higher scores indicate superior reconstruction quality and structural fidelity.

\begin{table}[t]
    \centering
    \caption{\textbf{NVS Benchmark Results on TerraSky3D.} Scene types are G (Ground), A (Aerial), and M (Mixed Aerial and Ground).}
    \label{tab:nvs}
    \resizebox{\columnwidth}{!}{% Adjust table size to column width
    \begin{tabular}{llc ccc}
        \toprule
        \multicolumn{2}{c}{\textbf{Scene}} & \textbf{Type} & \textbf{PSNR}$\uparrow$ & \textbf{SSIM}$\uparrow$ & \textbf{LPIPS}$\downarrow$\\
        \cmidrule(lr){1-2} \cmidrule(lr){3-3} \cmidrule(lr){4-6}
        % % Reference set
        % TT?
        
        % Training scene
        \multirow{5}{*}{\rotatebox{90}{\textbf{Training set}}} & 
        
        Trieste Unity Square & G & 22.21 & 0.759 & 0.318 \\
        & Vienna Natural History Museum & G & 21.92 & 0.754 & 0.342 \\
        & Udine Abandoned Bridge & A & 21.73 & 0.625 & 0.326 \\
        & Gorizia Military Shrine & M & 21.53 & 0.656 & 0.398 \\
        & Udine Villalta Church   & M & 23.76 & 0.746 & 0.302 \\
        \cmidrule(lr){1-2} \cmidrule(lr){3-3} \cmidrule(lr){4-6}
        
        % Test set
        \multirow{12}{*}{\rotatebox{90}{\textbf{Test set}}} & 
        Graz Church & G & 22.39 & 0.801 & 0.197 \\
        & Graz Townhall & G & 21.54 & 0.733 & 0.266 \\
        & Graz University & G & 24.40 & 0.821 & 0.233 \\
        & Munich Frauenkirche & G & 21.12 & 0.728 & 0.250 \\
        & Munich Marienplatz & G & 19.38 & 0.708 & 0.323 \\
        & Munich Theatine Church & G & 21.53 & 0.799 & 0.225 \\
        & Salzburg Andrakirche & G & 21.32 & 0.811 & 0.222 \\
        & Salzburg Rechte Altstadt & G & 21.56 & 0.856 & 0.186 \\
        & Udine Devils Bridge & M & 23.27 & 0.748 & 0.324 \\
        & Udine Fagagna Church & M & 21.02 & 0.639 & 0.361 \\
        & Udine Villalta Castle & M & 19.95 & 0.529 & 0.488 \\
        & Vienna State Opera & G & 22.33 & 0.826 & 0.230 \\
         
        \bottomrule
    \end{tabular}
    }
    
\end{table}

\begin{figure*}[t]
    \centering

    % Left image (top aligned)
    \begin{minipage}[t]{0.45\textwidth}
        \vspace{0pt}
        \centering
        \includegraphics[width=\textwidth]{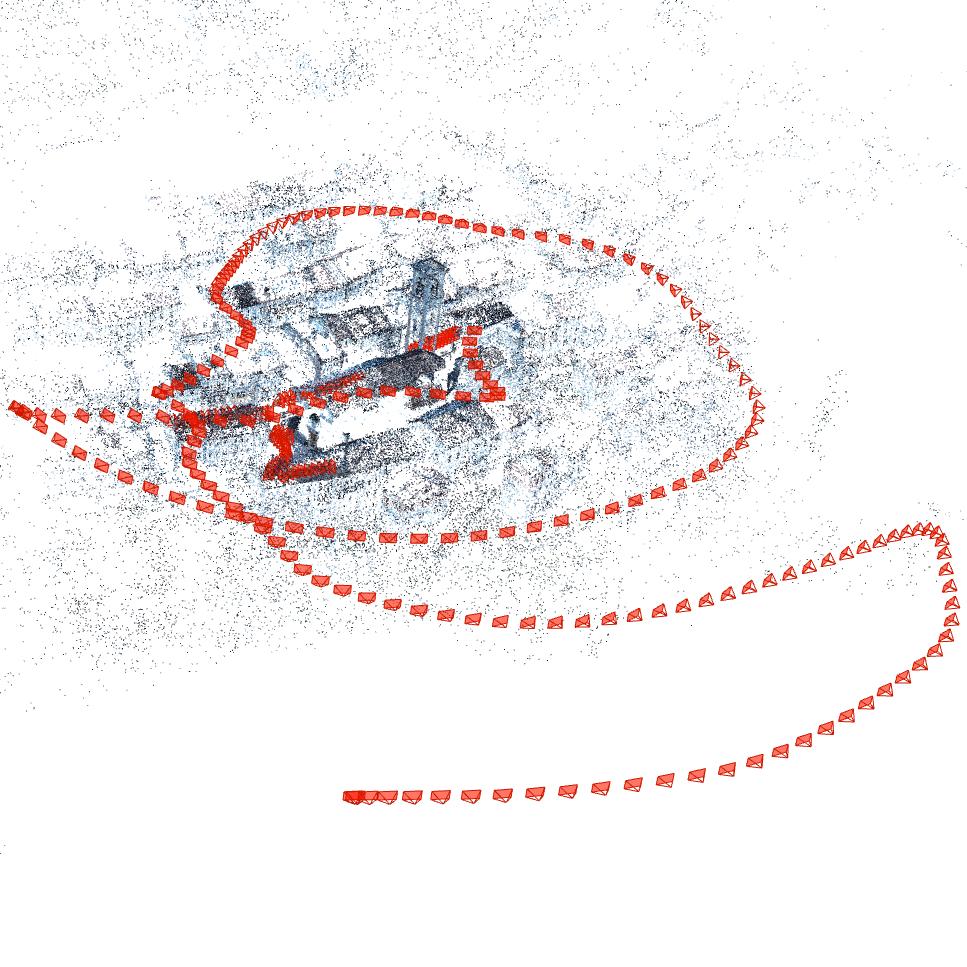}
    \end{minipage}
    \hfill
    % Right 2x4 grid (top aligned) - THIS IS THE MODIFIED PART
    \begin{minipage}[t]{0.53\textwidth}
        \vspace{0pt}
        \centering

    % --- Row 1 ---
    \begin{minipage}[t]{0.24\textwidth}
        \vspace{0pt} % Ensures minipage aligns at the very top
        \includegraphics[width=\textwidth]{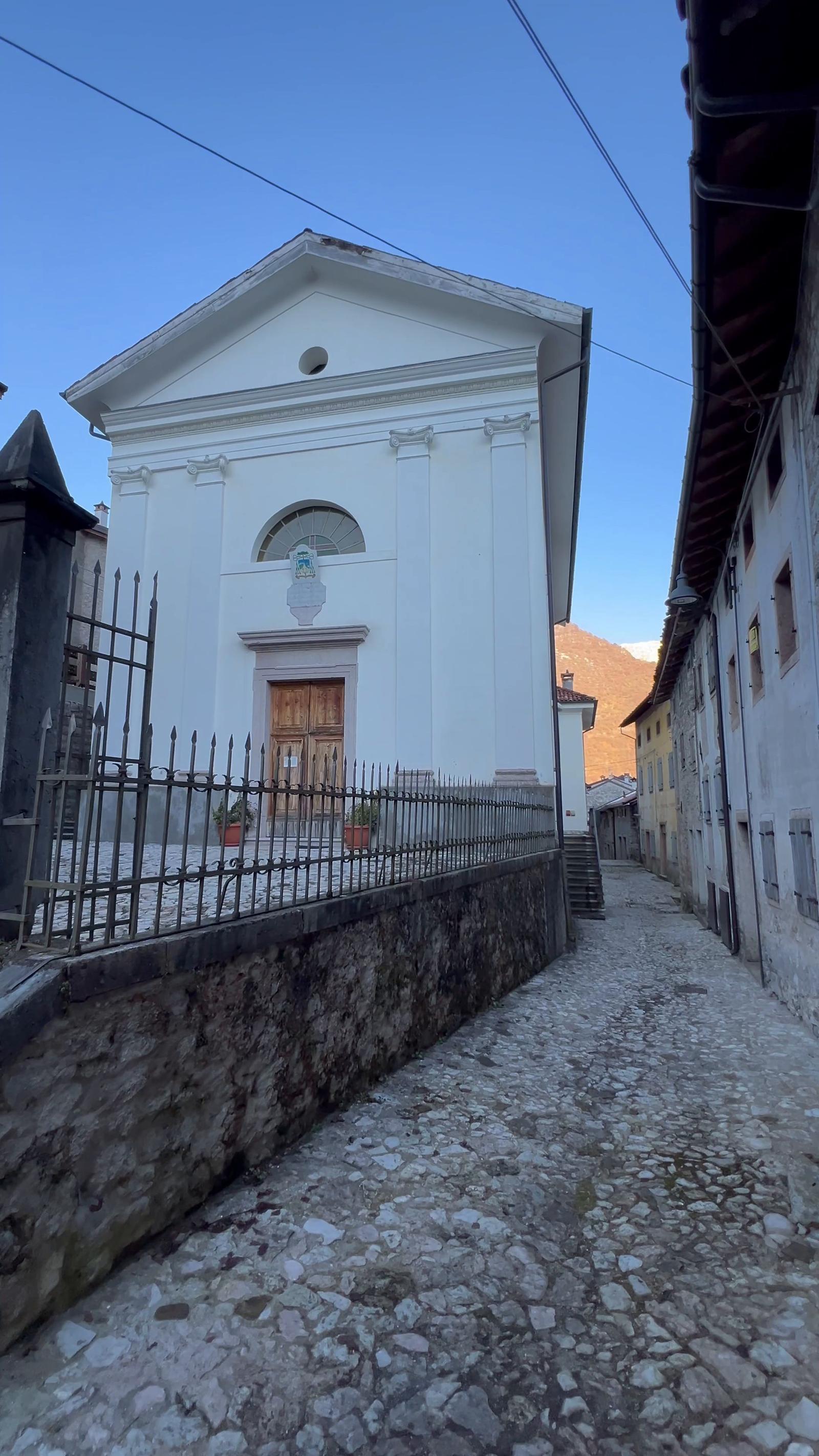}
    \end{minipage}
    \hspace{2pt} % <-- KEY CHANGE: Adds equal horizontal space
    \begin{minipage}[t]{0.24\textwidth}
        \vspace{0pt}
        \includegraphics[width=\textwidth]{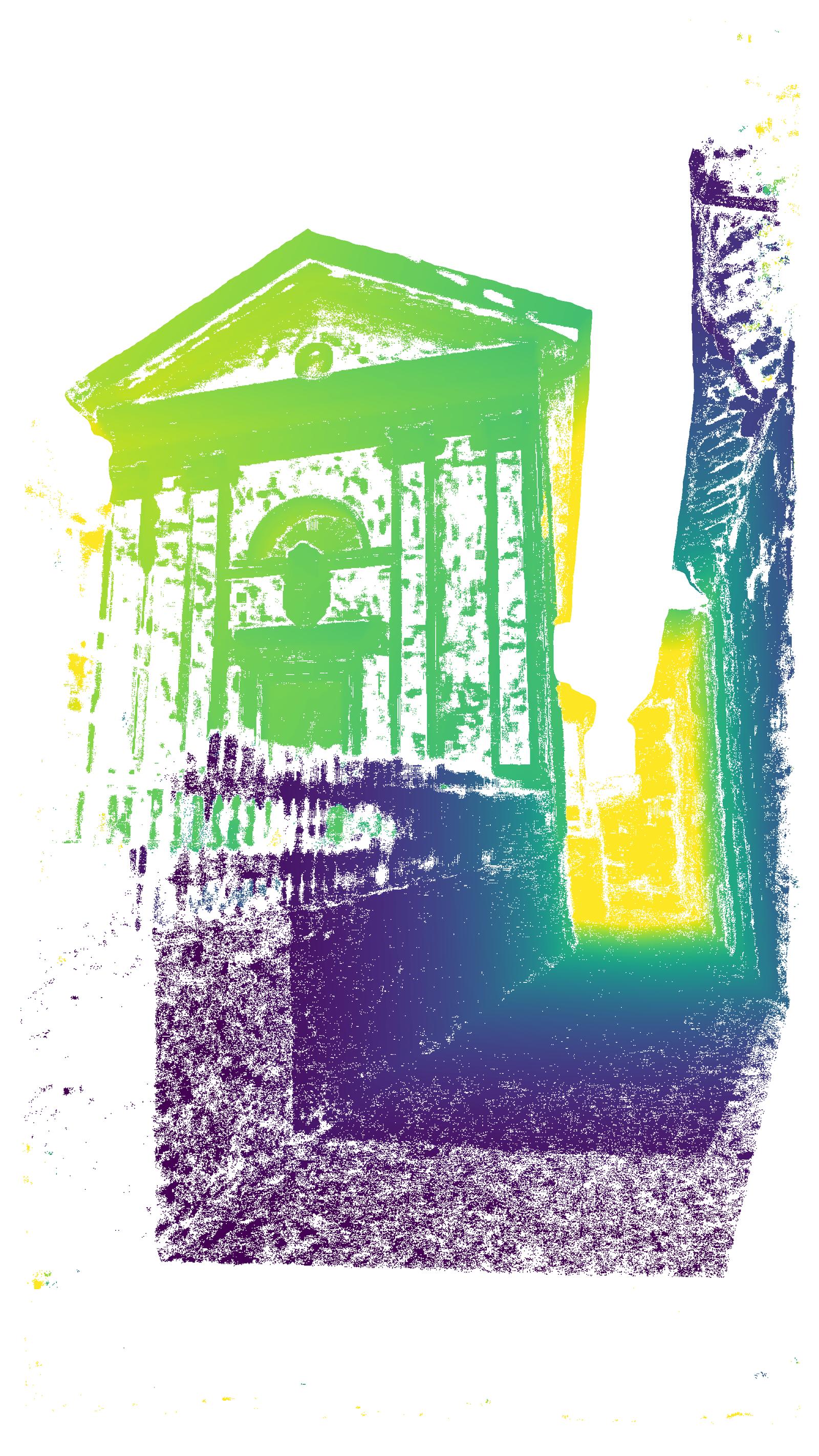}
    \end{minipage}
    \hspace{2pt} % <-- KEY CHANGE: Adds equal horizontal space
    \begin{minipage}[t]{0.375\textwidth}
        \vspace{0pt}
        \includegraphics[width=\textwidth]{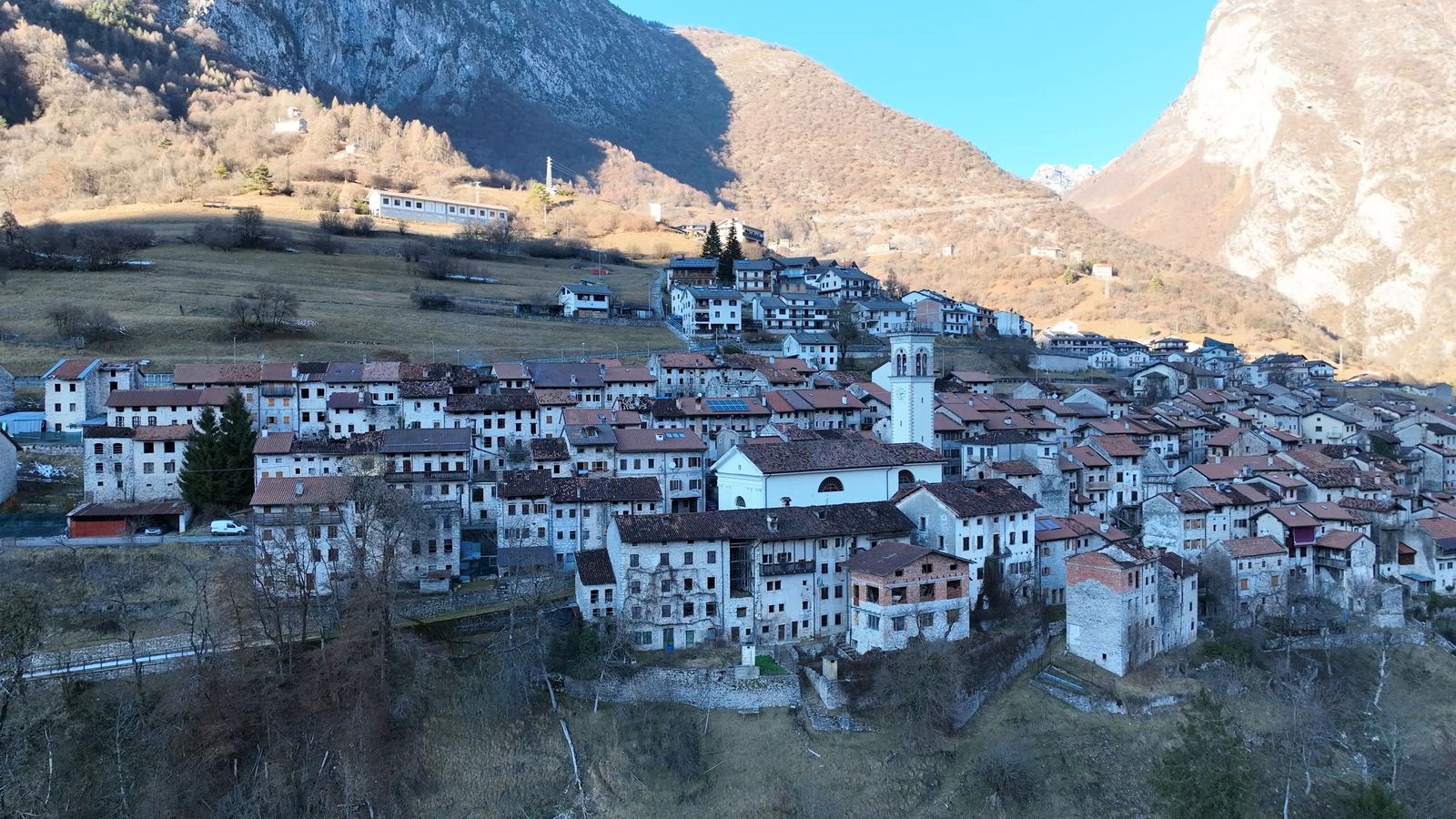}
        \vspace{2pt} % Adds space between the stacked images in this column
        \includegraphics[width=\textwidth]{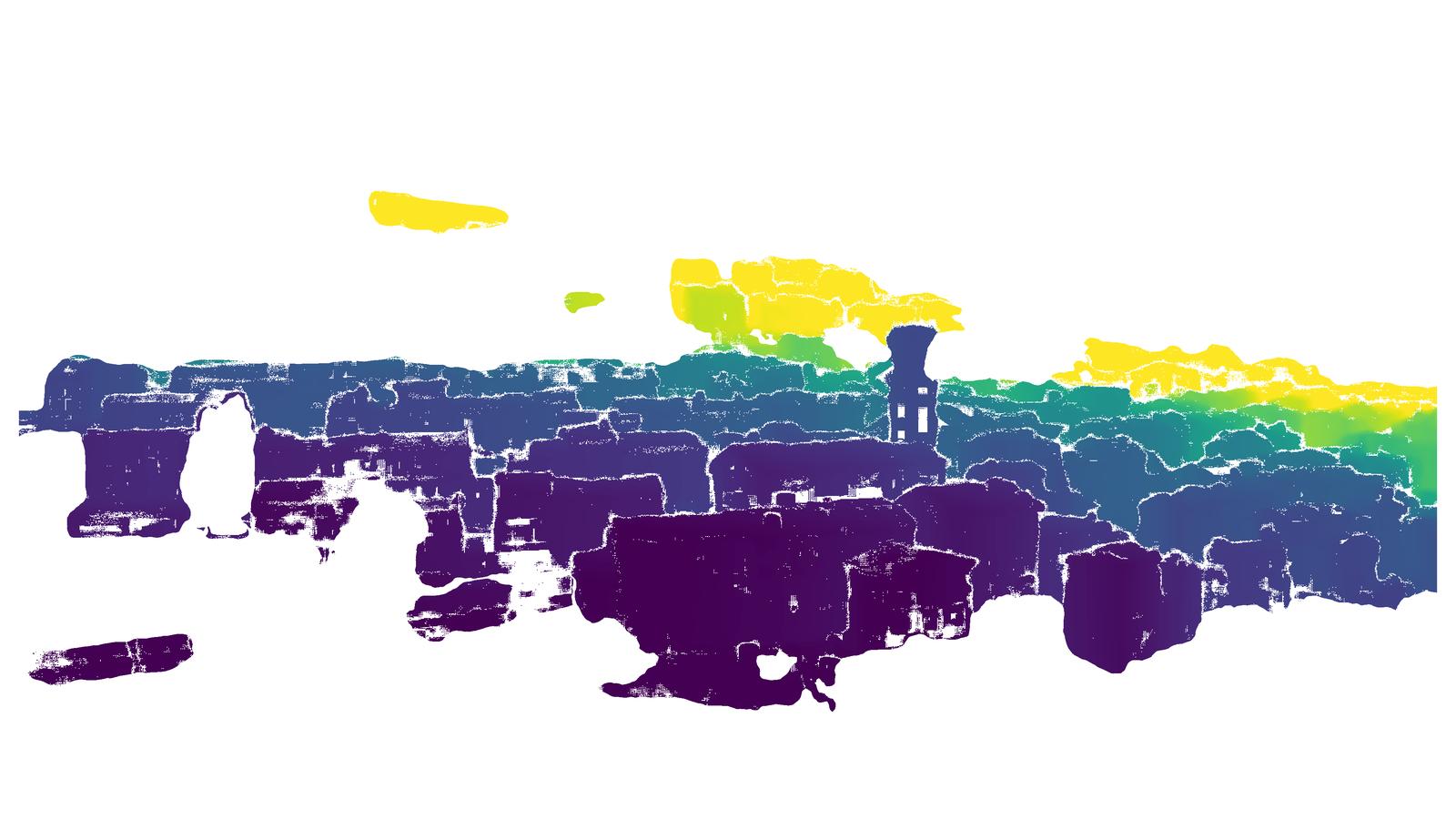}
    \end{minipage}
    
    \vspace{8pt} % <-- KEY CHANGE: This single command creates all the vertical space between Row 1 and Row 2
    
    % --- Row 2 ---
    \begin{minipage}[t]{0.24\textwidth}
        \vspace{0pt}
        \includegraphics[width=\textwidth]{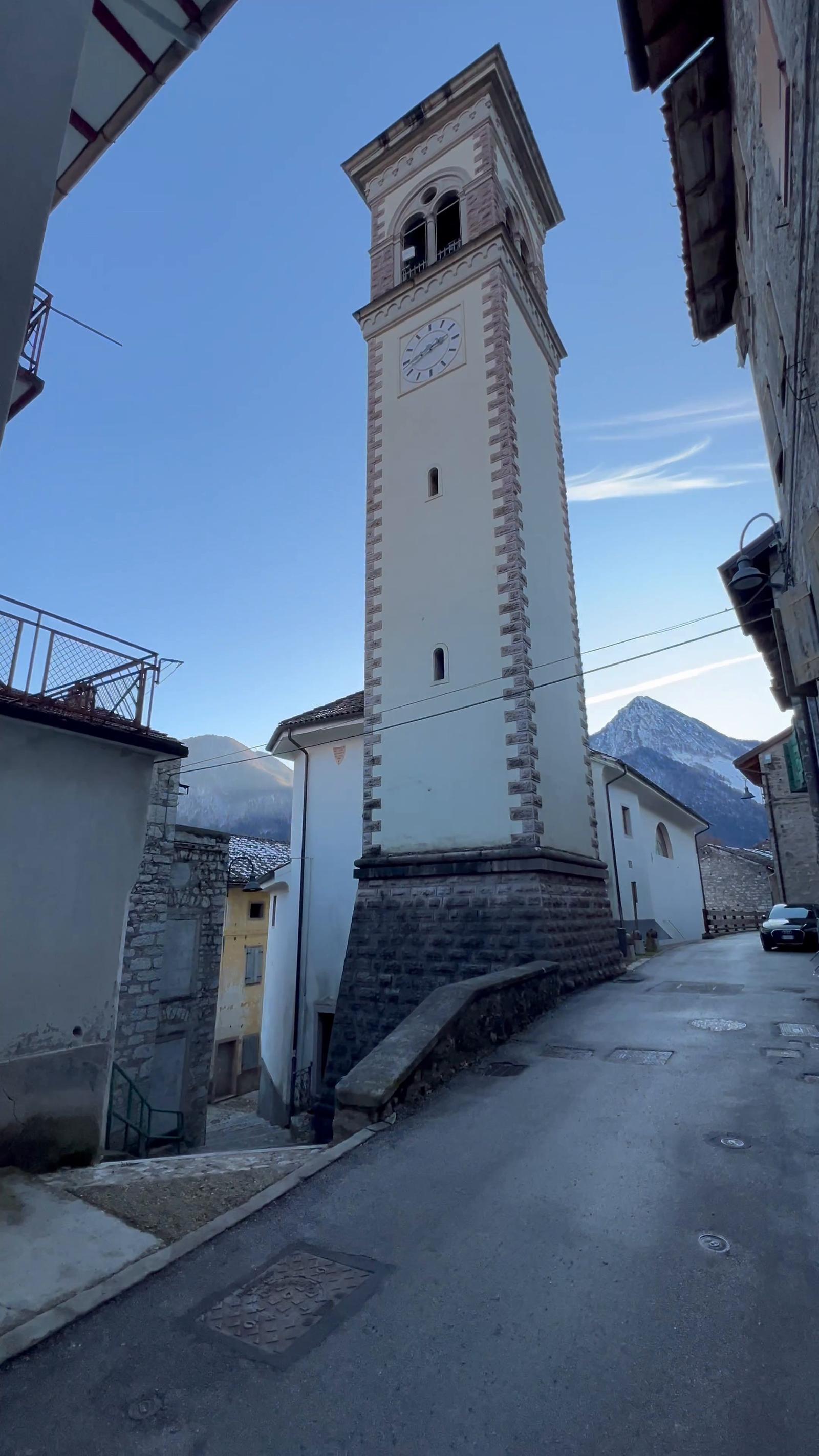}
    \end{minipage}
    \hspace{2pt} % <-- KEY CHANGE: Adds equal horizontal space
    \begin{minipage}[t]{0.24\textwidth}
        \vspace{0pt}
        \includegraphics[width=\textwidth]{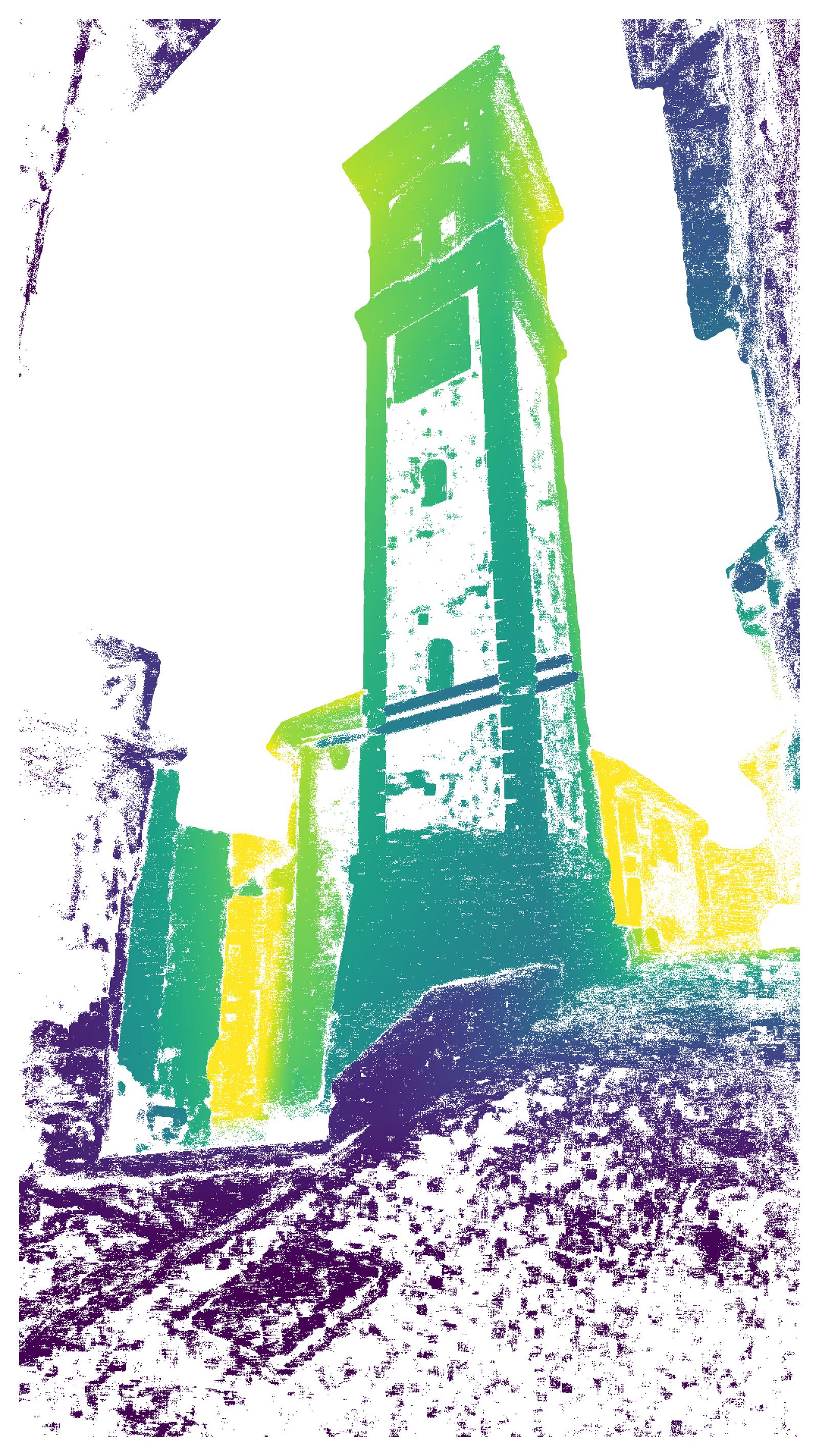}
    \end{minipage}
    \hspace{2pt} % <-- KEY CHANGE: Adds equal horizontal space
    \begin{minipage}[t]{0.375\textwidth}
        \vspace{0pt}
        \includegraphics[width=\textwidth]{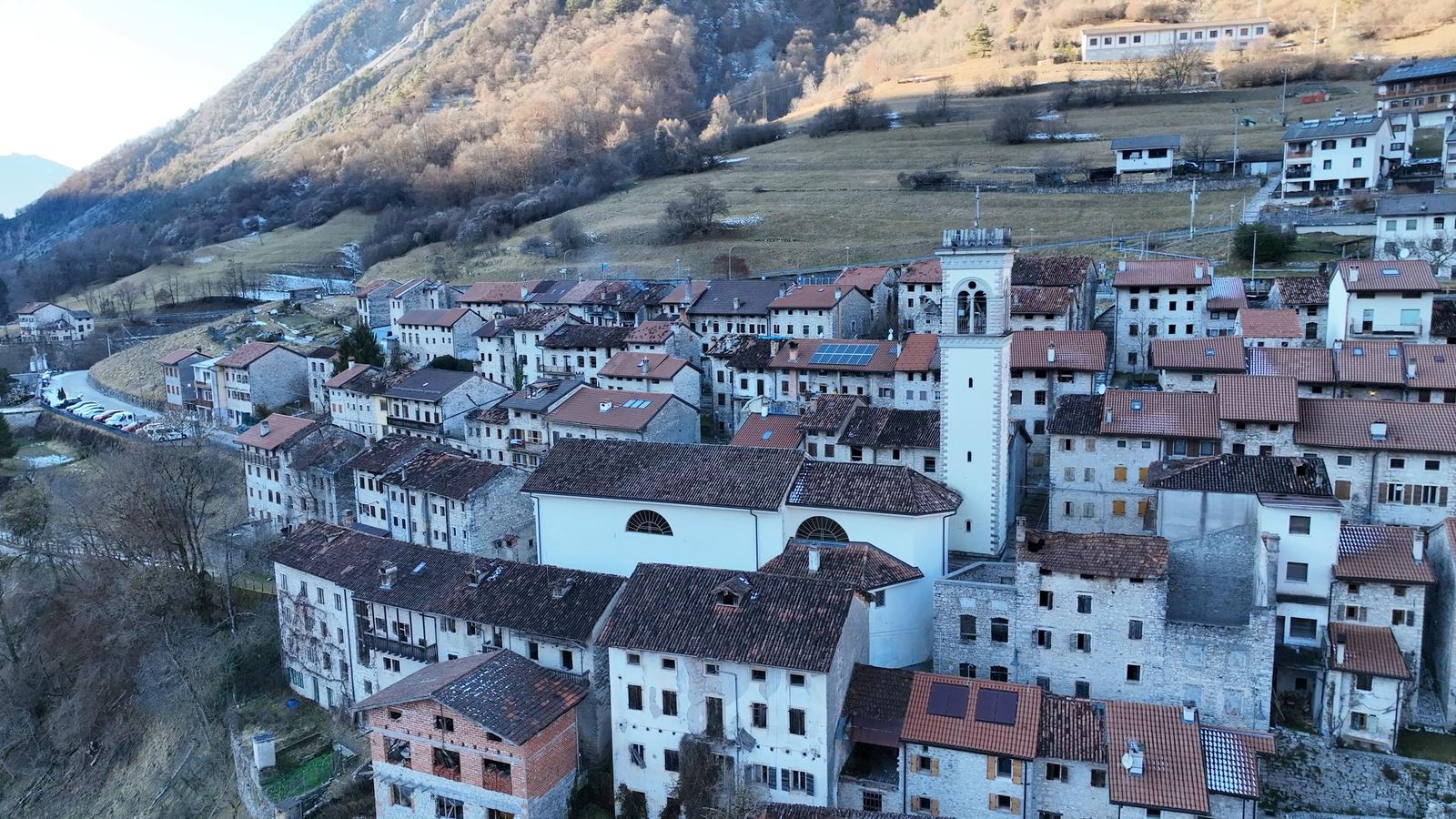}
        \vspace{2pt} % Adds space between the stacked images in this column
        \includegraphics[width=\textwidth]{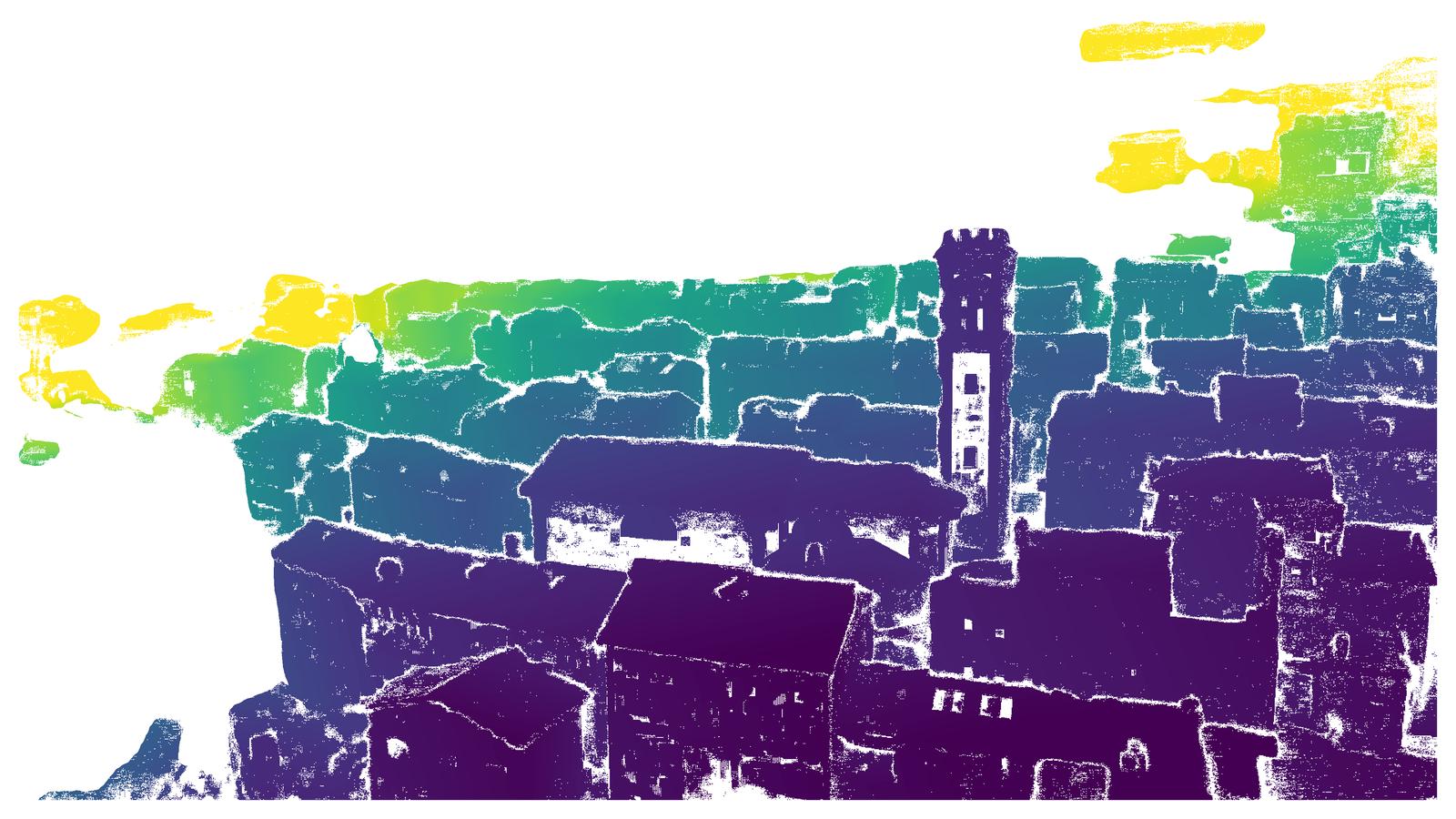}
    \end{minipage}

    \end{minipage} 
    \caption{\textbf{Example Scene from TerraSky3D.} Left: Sparse reconstruction of Erto e Casso, Pordenone, Italy. Right: Representative images collected from aerial and ground perspectives, shown with their corresponding semantically filtered depth maps.} 
    \label{fig:erto}
\end{figure*}

\subsection{Bidirectional Geometric Consistency}

To evaluate our depth maps, we perform a bidirectional reprojection consistency check. We utilize the COLMAP viewgraphs to select image pairs $(I_i, I_j)$ with at least 100 matches.
For each pair, we verify consistency by computing the cyclic reprojection error. Let $\mathbf{p}_i$ be a pixel coordinate in image $I_i$. We first unproject $\mathbf{p}_i$ to 3D space using its estimated depth $D_i(\mathbf{p}_i)$ and project it into the view of $I_j$ to obtain the coordinate $\mathbf{p}_{i \to j}$. We then sample the depth $D_j$ at this new location to project back into the original frame $I_i$, resulting in the coordinate $\mathbf{p}_{i \to j \to i}$. The cyclic error is defined as the Euclidean distance:
\begin{equation}
    E_{cyc}(\mathbf{p}_i) = ||\mathbf{p}_i - \mathbf{p}_{i \to j \to i}||_2
\end{equation}
We run the consistency checks in both directions, i.e., $I_i\rightarrow I_j$ and $I_j \rightarrow I_i$. To include all mixed aerial and ground pairs we perform this check across all possible image pairs. %rather than restricting the evaluation to the COLMAP viewgraph.
The evaluation considers only \textit{valid} pixels. A pixel $\mathbf{p}_i$ is said \textit{valid} if its projection falls within the image bounds. 
%\cref{tab:geo_consistency} reports the percentage of valid pixels at cumulative thresholds. 

We compare TerraSky3D with MegaDepth in terms of both scale and scope. To ensure a fair comparison, we downscale our 4K resolution images to a resolution of 1,600 pixels on the longest edge. We evaluate both datasets on their respective test sets, as these consist of scenes processed via the same pipeline as the training data, thereby accurately reflecting overall dataset characteristics. Specifically, for the MegaDepth~\cite{li2018megadepth} comparison, we focus on the IMC Phototourism~\cite{jin2021image} test set, as it constitutes a representative subset of the larger MegaDepth collection.

\begin{table}[t]
    \centering
    \caption{\textbf{Bidirectional Geometric Consistency on TerraSky3D.} We report performance in terms of cumulative inlier percentage under varying pixel error thresholds.} %Scene types are G (Ground), A (Aerial), and Mixed (Mixed Air and Ground).}
     \label{tab:geo_consistency}
    \resizebox{\columnwidth}{!}{% Adjust table size to column width
    \begin{tabular}{llcr cccc} 
        \toprule
        % Main header for metrics
        \multicolumn{4}{c}{} & \multicolumn{4}{c}{\textbf{Inliers (\%)}} \\ 
        \cmidrule(lr){5-8} % Rule over the metric columns
        & \multicolumn{1}{l}{\textbf{Scene}} & \textbf{Type} & \textbf{Pairs} &  \textbf{1px}  & \textbf{3px} & \textbf{5px} & \textbf{10px} \\ 
        \cmidrule(lr){1-2} \cmidrule(lr){3-3} \cmidrule(lr){4-4} \cmidrule(lr){5-8}
        
        \multirow{13}{*}{\rotatebox{90}{\textbf{MegaDepth}}} & 
        British Museum & G & 4104 & 39.0 & 51.3 & 53.4 & 54.9 \\
        & Florence Cathedral Side & G & 4134 & 47.8 & 62.3 & 65.3 & 67.7 \\
        & Lincoln Memorial Statue & G & 3195 & 36.6 & 56.4 & 62.7 & 67.9 \\
        & London Bridge & G & 2200 & 42.1 & 56.1 & 60.5 & 65.0 \\
        & Milan Cathedral & G & 4279 & 50.6 & 64.5 & 67.9 & 70.8 \\
        & Mount Rushmore & G & 4018 & 71.7 & 77.3 & 78.0 & 78.5 \\
        & Piazza San Marco & G & 2188 & 53.1 & 69.6 & 73.7 & 77.0 \\
        & Reichstag & G & 2116 & 33.8 & 44.0 & 46.7 & 48.9 \\
        & Sacre Coeur & G & 3060 & 24.3 & 32.8 & 35.5 & 38.4 \\
        & Sagrada Familia & G & 3550 & 39.8 & 58.5 & 63.5 & 67.7 \\
        & St Pauls Cathedral & G & 3267 & 40.9 & 58.5 & 63.5 & 67.8 \\
        & St Peters Square & G & 2915 & 36.7 & 50.1 & 53.7 & 56.9 \\
        & \textbf{Mean} & \textbf{G} & \textbf{3252} & \textbf{43.0} & \textbf{56.8} & \textbf{60.4} & \textbf{63.4} \\

        \cmidrule(lr){1-2} \cmidrule(lr){3-3} \cmidrule(lr){4-4} \cmidrule(lr){5-8}

        % Test set
        \multirow{13}{*}{\rotatebox{90}{\textbf{TerraSky3D}}} & 
        Graz Church & G & 12736 & 35.3 & 56.9 & 66.1 & 76.0 \\
        & Graz Townhall & G & 9556 & 40.0 & 59.4 & 67.7 & 77.0 \\
        & Graz University & G & 9792 & 59.6 & 74.5 & 78.7 & 82.4 \\
        & Munich Frauenkirche & G & 2446 & 63.4 & 81.1 & 85.6 & 89.3 \\
        & Munich Marienplatz & G & 5821 & 48.6 & 61.9 & 66.3 & 70.9 \\
        & Munich Theatine Church & G & 2617 & 60.8 & 77.1 & 81.3 & 85.0 \\
        & Salzburg Andrakirche & G & 3085 & 41.7 & 57.2 & 61.8 & 66.0 \\
        & Salzburg Rechte Altstadt & G & 1039 & 62.4 & 74.6 & 77.7 & 80.5 \\
        & Udine Devils Bridge & M & 9569 & 43.7 & 57.2 & 61.3 & 65.4 \\
        & Udine Fagagna Church & M & 3848 & 56.1 & 67.6 & 70.3 & 72.9 \\
        & Udine Villalta Castle & M & 9023 & 51.4 & 64.5 & 68.5 & 72.3 \\
        & Vienna State Opera & G & 2310 & 54.0 & 67.9 & 71.8 & 75.5 \\
        & \textbf{Mean} & \textbf{M} & \textbf{5986} & \textbf{51.4} & \textbf{66.7} & \textbf{71.4} & \textbf{76.1} \\
        
        %\cmidrule{2-8}
        %& All Training set (mean)& - \\ 
        \bottomrule
    \end{tabular}
    }
    
\end{table}

\cref{tab:geo_consistency} reports the percentage of \textit{valid} pixels at cumulative thresholds demonstrating that our data exhibits higher geometric consistency across all error thresholds. It is important to note that even a small percentage-wise improvement represents thousands of additional valid points that can be used in training or evaluation tasks. Furthermore, our dataset includes aerial and ground image pairs, a challenging cross-modal scenario absent in MegaDepth.

\begin{figure*}[t]
    \centering

    % Left image (top aligned)
    \begin{minipage}[t]{0.45\textwidth}
        \vspace{0pt}
        \centering
        \includegraphics[width=\textwidth]{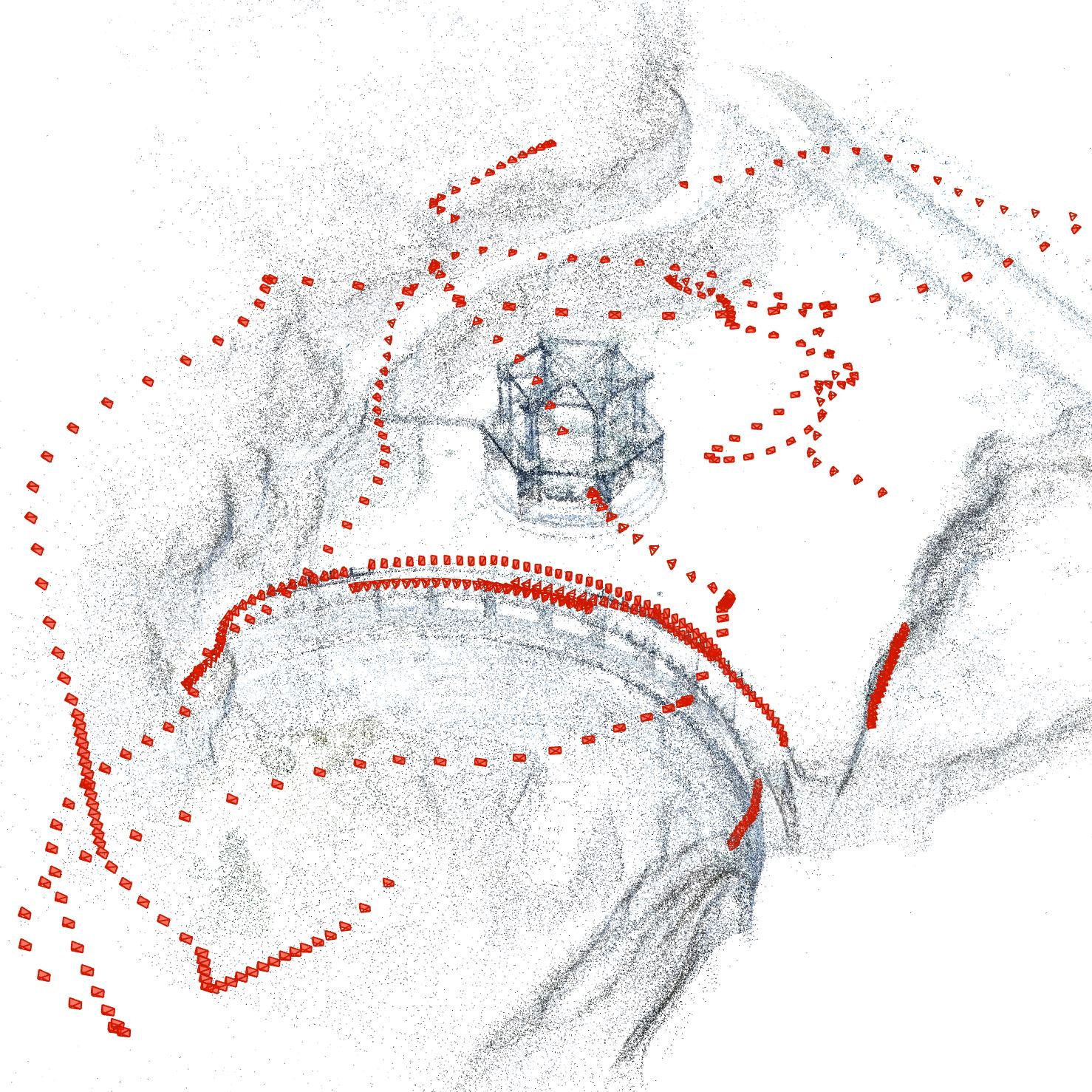}
    \end{minipage}
    \hfill
    % Right 2x4 grid (top aligned) - THIS IS THE MODIFIED PART
    \begin{minipage}[t]{0.53\textwidth}
        \vspace{0pt}
        \centering

    % --- Row 1 ---
    \begin{minipage}[t]{0.24\textwidth}
        \vspace{0pt} % Ensures minipage aligns at the very top
        \includegraphics[width=\textwidth]{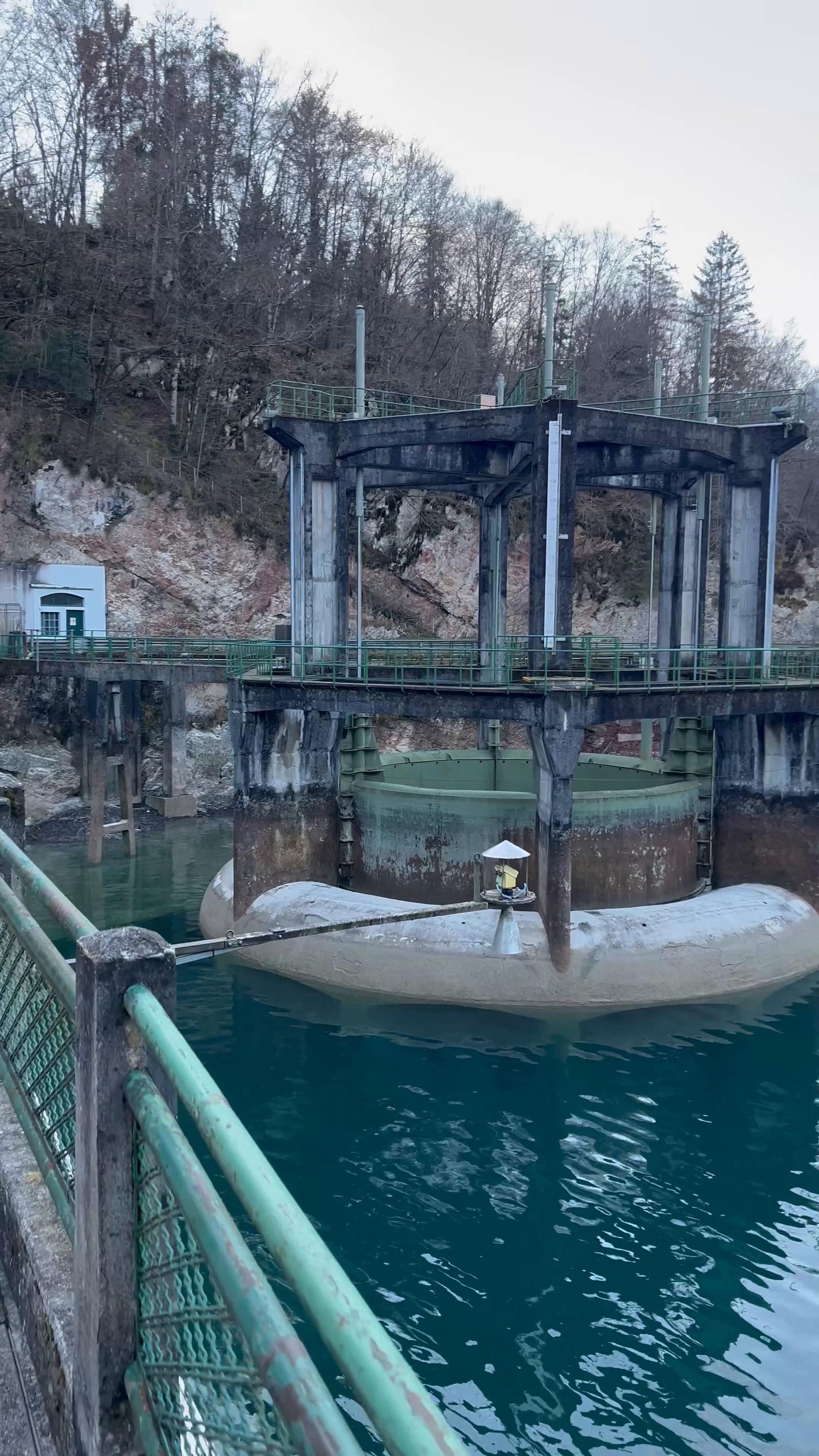}
    \end{minipage}
    \hspace{2pt} % <-- KEY CHANGE: Adds equal horizontal space
    \begin{minipage}[t]{0.24\textwidth}
        \vspace{0pt}
        \includegraphics[width=\textwidth]{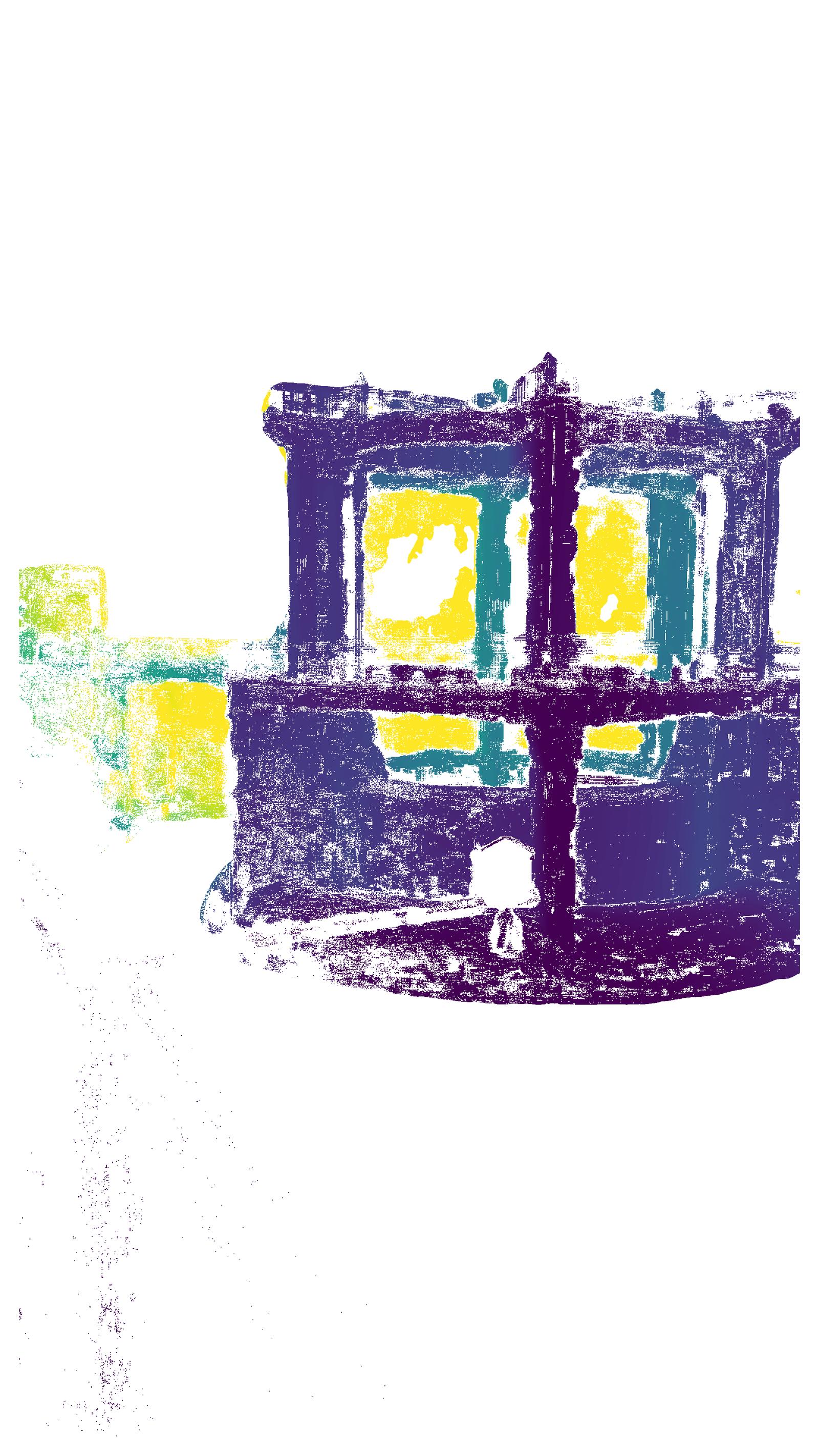}
    \end{minipage}
    \hspace{2pt} % <-- KEY CHANGE: Adds equal horizontal space
    \begin{minipage}[t]{0.375\textwidth}
        \vspace{0pt}
        \includegraphics[width=\textwidth]{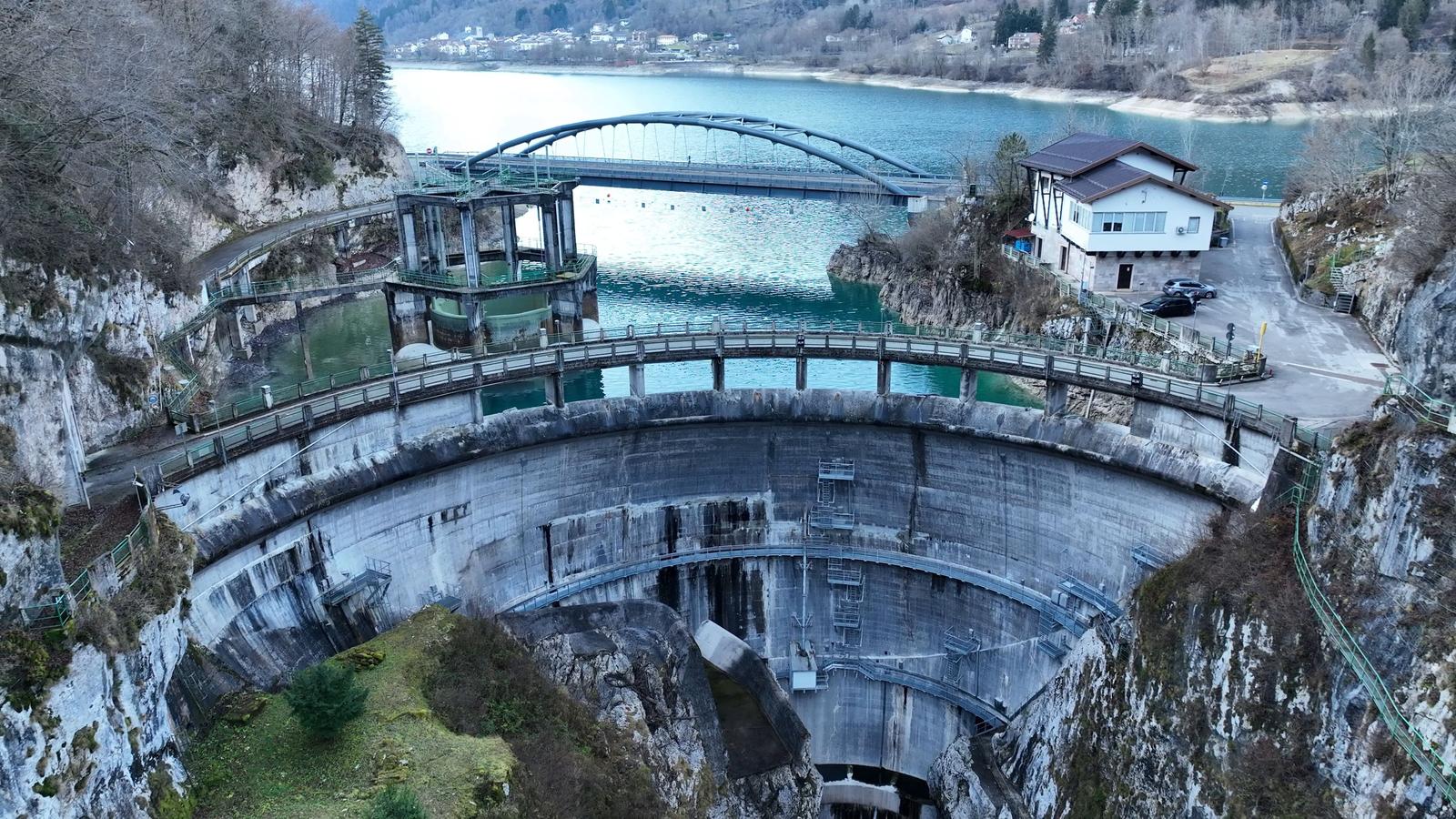}
        \vspace{2pt} % Adds space between the stacked images in this column
        \includegraphics[width=\textwidth]{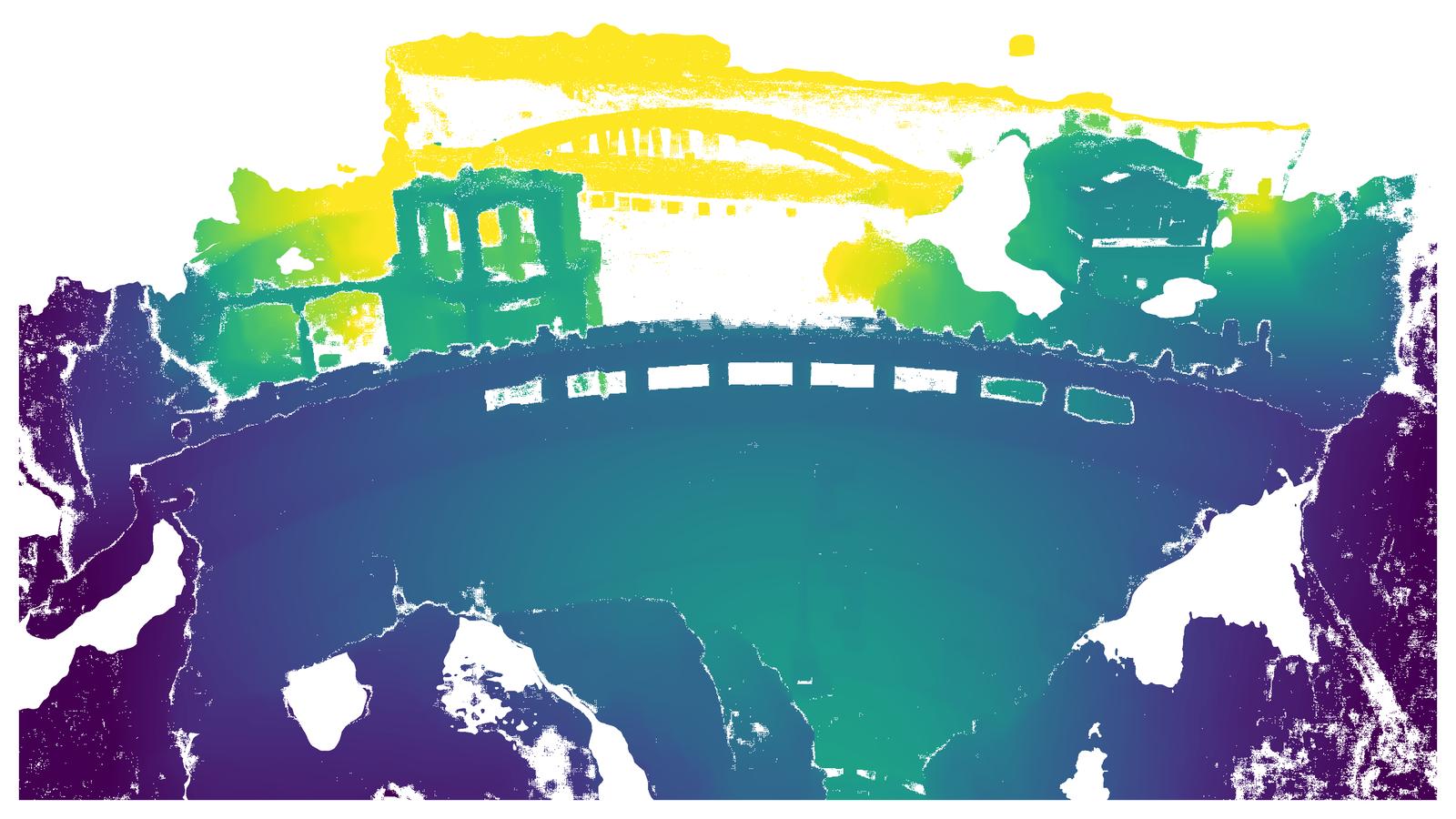}
    \end{minipage}
    
    \vspace{8pt} % <-- KEY CHANGE: This single command creates all the vertical space between Row 1 and Row 2
    
    % --- Row 2 ---
    \begin{minipage}[t]{0.24\textwidth}
        \vspace{0pt}
        \includegraphics[width=\textwidth]{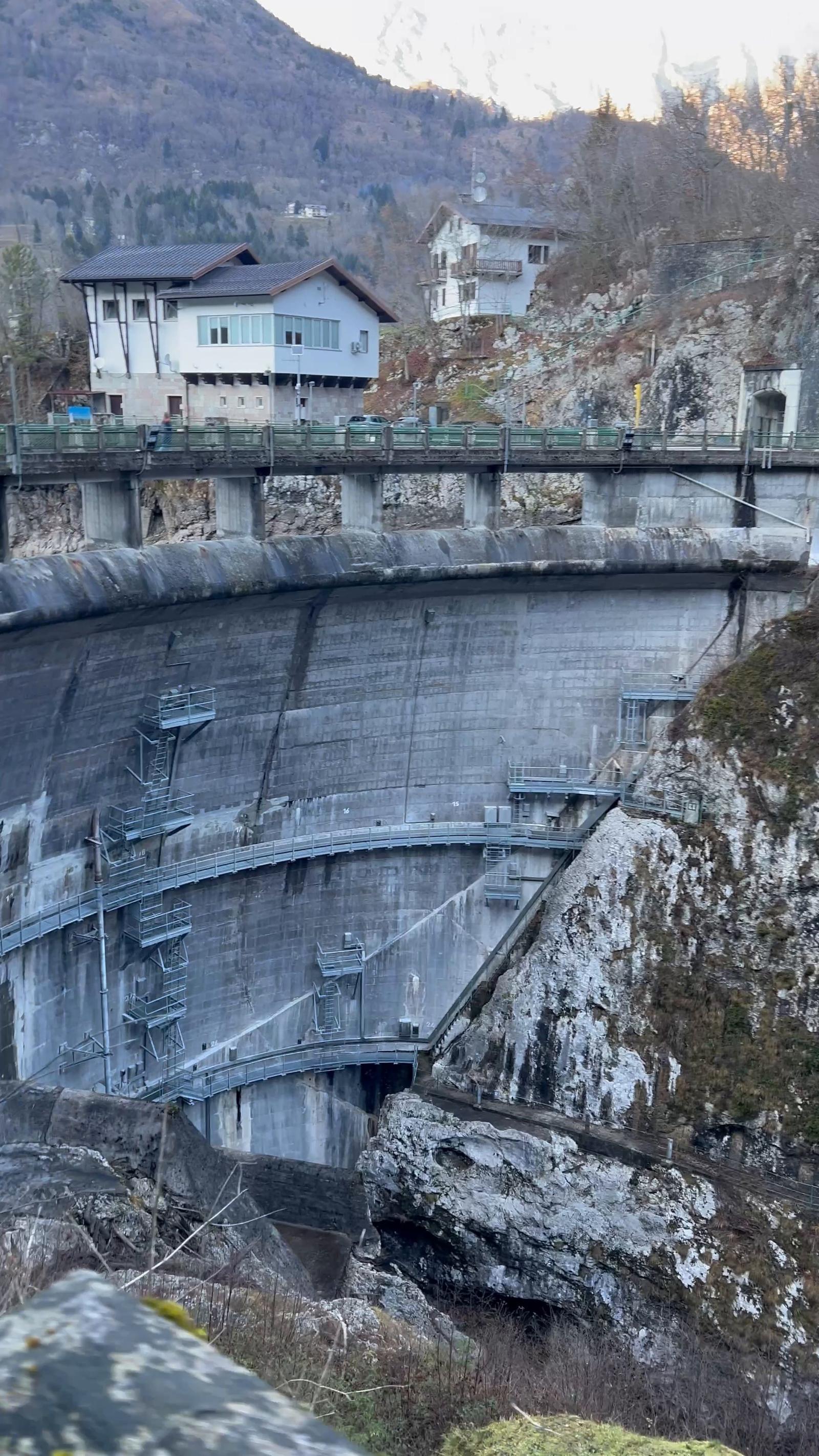}
    \end{minipage}
    \hspace{2pt} % <-- KEY CHANGE: Adds equal horizontal space
    \begin{minipage}[t]{0.24\textwidth}
        \vspace{0pt}
        \includegraphics[width=\textwidth]{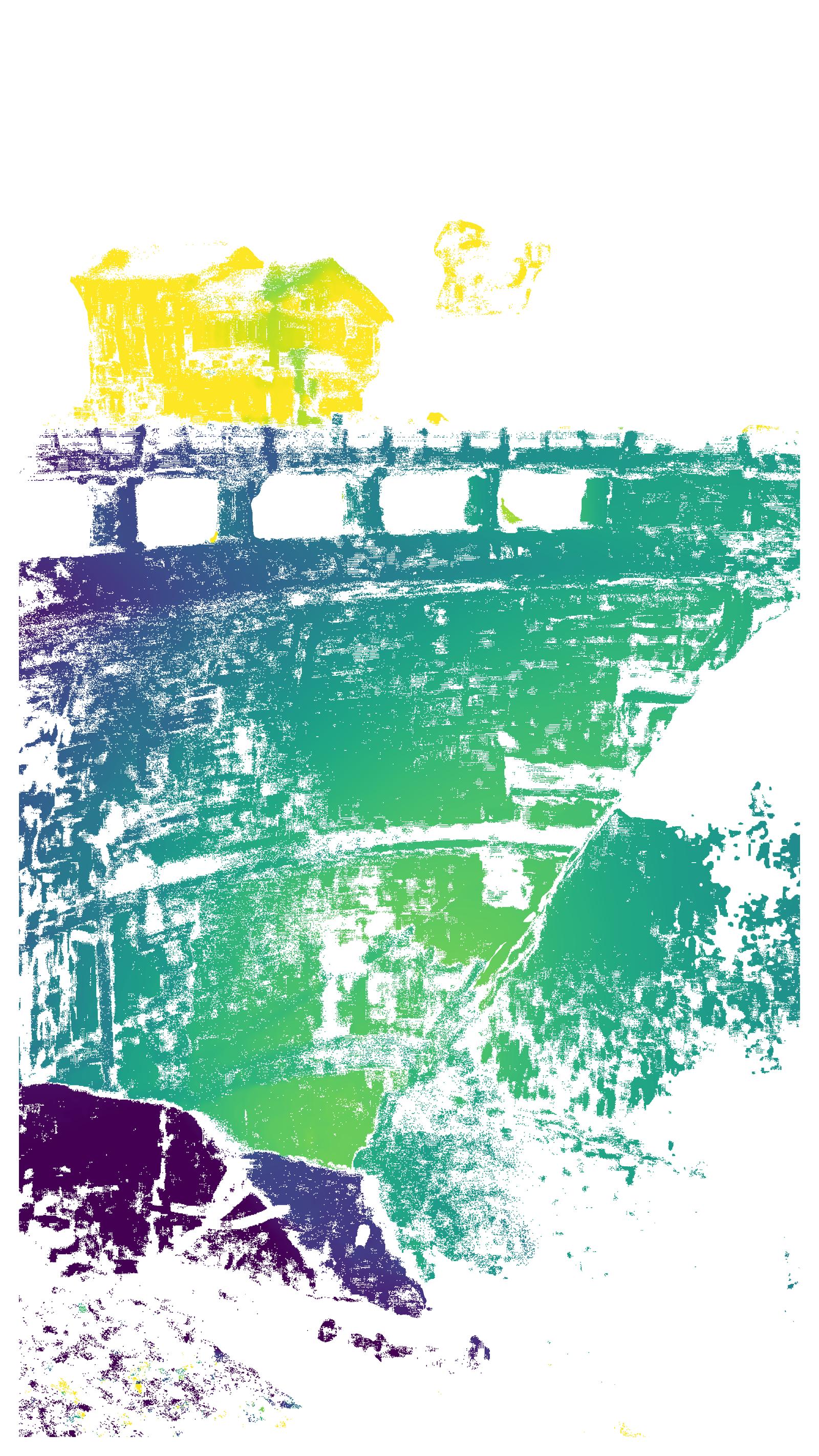}
    \end{minipage}
    \hspace{2pt} % <-- KEY CHANGE: Adds equal horizontal space
    \begin{minipage}[t]{0.375\textwidth}
        \vspace{0pt}
        \includegraphics[width=\textwidth]{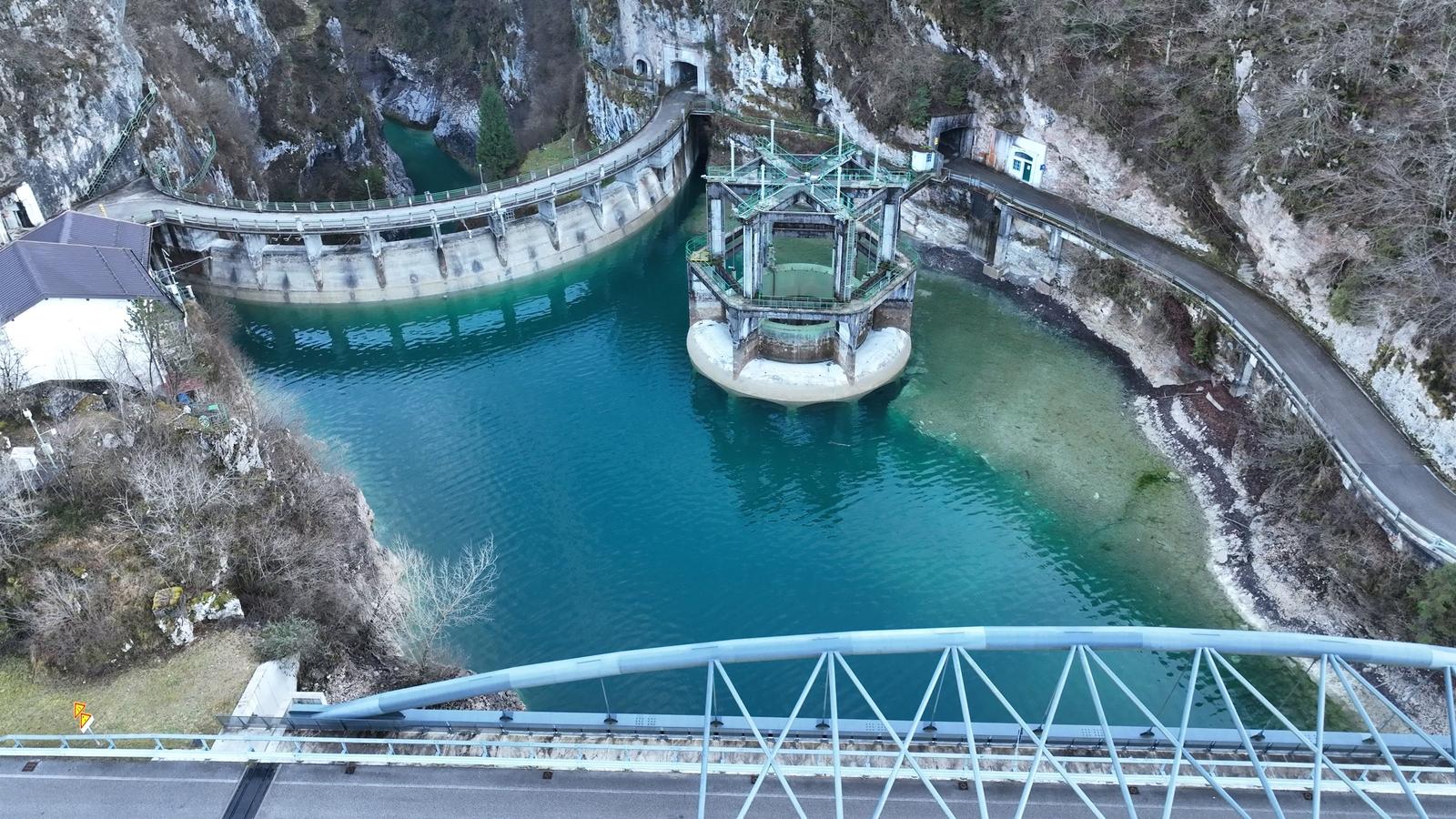}
        \vspace{2pt} % Adds space between the stacked images in this column
        \includegraphics[width=\textwidth]{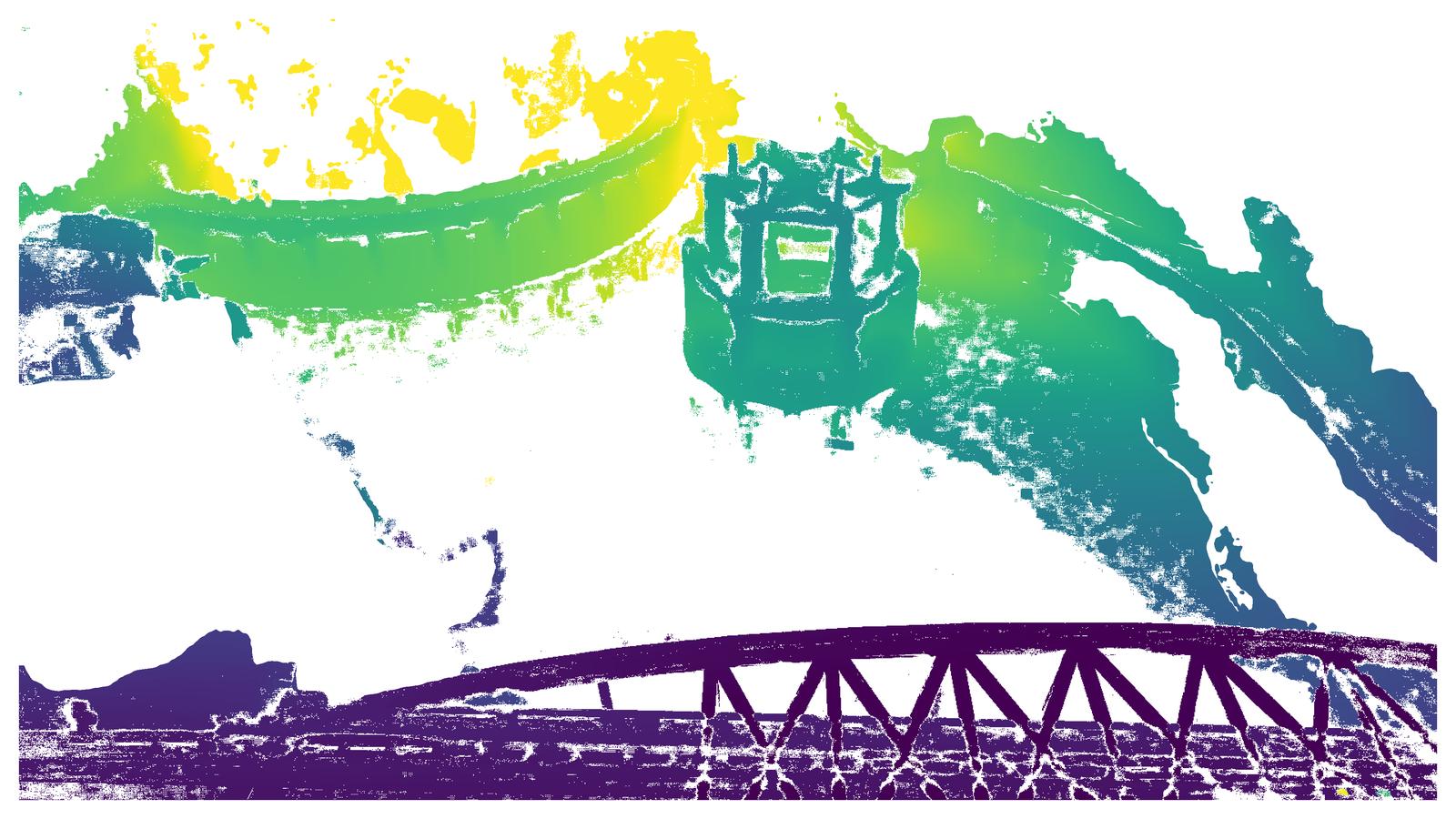}
    \end{minipage}

    \end{minipage} 
    \caption{\textbf{Example Scene from TerraSky3D.} Left: Sparse reconstruction of the Barcis Dam, Pordenone, Italy. Right: Representative images collected from aerial and ground perspectives, shown with their corresponding semantically filtered depth maps.} 
    \label{fig:barcis}
\end{figure*}

\section{Limitations}
\label{sec:limitations}

While TerraSky3D significantly expands the availability of public data by introducing complex aerial and ground scenes, it is nonetheless subject to specific technical limitations. Unlike datasets that rely on rendered synthetic frames, our dataset is grounded in raw photographs extracted from high-resolution video sequences. While this ensures real-world authenticity, it introduces unique challenges inherent to physical capture environments.

First, although the source sequences are high-definition, certain frames suffer from motion blur. This phenomenon is particularly pronounced during rapid camera movements or in low-light conditions. 

Furthermore, while we implement a robust automated filtering pipeline to uphold data integrity, occasionally the resulting depth maps and associated masks might be imprecise. In complex scenes with intricate geometry, automated processes may struggle to precisely delineate fine-grained boundaries, thin structures, or transparent surfaces. 

Finally, depth estimation quality is tied to angular variety and viewpoint density. To mitigate noise and structural incompleteness in regions with sparse multi-view coverage, we provide APD-MVS confidence masks. This allows users to perform further filtering to maintain higher structural fidelity and visual clarity.

\section{Conclusion}
\label{sec:conclusion}

We introduce \textbf{TerraSky3D}, a novel high-quality dataset comprising approximately 50,000 4K images, along with their corresponding sparse 3D reconstructions and depth maps. We address a critical gap in the literature by providing several scenes integrating aerial and ground-level perspectives, a modality essential for advancing the state of the art in cross-view localization and reconstruction. 

In particular, we show, that most current state-of-the-art models struggle in aerial and mixed scenarios, thus limiting their effectiveness in practical applications. Nevertheless, a SANDesc model retrained on our data outperforms the original model trained on MegaDepth. Ultimately, TerraSky3D provides a robust foundation for future research in large-scale 3D computer vision.

\paragraph{Acknowledgements} We thank Davide Casano, Luca Danelutti, Federico Fattori, Alessandro Menafra,  Florian Thaler, Runze Yuan,  and Stefano Zorzi for their contribution in data collection. 

This work has been supported by the FFG under Contract No. 881844 within the project ``Pro\textsuperscript{2}Future''.

{\small
\bibliographystyle{ieeenat_fullname}
\bibliography{11_references}
}

\ifarxiv \clearpage \appendix \section{Appendix Section}
\label{sec:appendix_section}
Supplementary material goes here.
 \fi

\end{document}